\documentclass[final]{fctr}
\usepackage[section]{placeins}   %
\usepackage{pifont}
\usepackage{pgfplots}
\pgfplotsset{compat=1.18}
\usetikzlibrary{arrows.meta,calc,positioning}
\definecolor{ciOrange}{HTML}{F26419} \definecolor{ciBlack}{HTML}{202833} \definecolor{ciStone}{HTML}{BFC3C7}
\definecolor{ciGray}{HTML}{9B9B95} \definecolor{ciBeige}{HTML}{EDDABB} \definecolor{ciGreen}{HTML}{4C8C6A}
\definecolor{ciBlue}{HTML}{3B719F} \definecolor{ciRed}{HTML}{C65A5A} \definecolor{ciAmber}{HTML}{D89A3D}
\definecolor{dsGreen}{HTML}{6B8E6B} \definecolor{dsBrown}{HTML}{B26B3D} \definecolor{dsGray}{HTML}{8A8278} \definecolor{dsInk}{HTML}{2A2520}

\newcommand{\benchmark}{NitiBench\textsuperscript{+}}
\newcommand{\nitibench}{NitiBench}

\title{\benchmark{}: A Temporal Thai Legal\\ Question Answering Benchmark Across 17 Domains}
\hypersetup{pdftitle={NitiBench+: A Temporal Thai Legal Question Answering Benchmark Across 17 Domains}, pdfauthor={Pawitsapak Akarajaradwong, Sarana Nutanong, Chompakorn Chaksangchaichot}}

\author{%
  Pawitsapak Akarajaradwong$^{1}$\quad Sarana Nutanong$^{2}$\quad Chompakorn Chaksangchaichot$^{1}$\\[3pt]
  \affil{$^{1}$FourCorners Team, VISAI AI, Thailand\quad $^{2}$VISTEC, Thailand}\\[1pt]
  \email{\{pawitsapaka\_visai, sarana.n, chompakornc\_pro\}@vistec.ac.th}\\[3pt]}

\resources{%
  \resourcerow{Dataset}{\href{https://huggingface.co/datasets/VISAI-AI/nitibench-plus}{\texttt{VISAI-AI/nitibench-plus}}}
  \resourcerow{Leaderboard}{\href{https://benchmark.fourcorners.law/nitibench-plus/}{\texttt{benchmark.fourcorners.law/nitibench-plus}}}}
\disclaimer{This report describes the evaluation of language models. Nothing in it is legal advice,
  and no model output quoted here should be relied on for a legal decision.}

\begin{document}
\maketitle

\begin{abstract}
Law changes over time. A court decides a case under the law in force on that day, and legal QA benchmarks grade answers against today's law.
We checked 156{,}650 decisions of four Thai courts. Up to 67\% of them cite law that has since been amended or repealed.
We introduce \benchmark{}, the sequel of our \nitibench{}, a Thai legal QA benchmark with 1{,}670 items across 17 domains, built from decisions of three courts.
For each cited section it holds both the text in force on the decision date and the current text, and its retrieval corpus holds every dated version of every Thai statute section.
Answers are graded on the Thai Bar Council rubric, the Bar-standard score, and on the overlap between the cited sections and the court's.
The report has two parts.
Part I measures the cost of the wrong text on 13 open-weight models. Reading the current text of amended law drops the Bar-standard score by 0.25 and missing repealed law drops it by 0.79. The penalty comes from rule changes, 0.45 on the 428 rule-changed items and none on the 331 wording-only items, and Citation F1 does not see it. The text in force is the fair setting.
Part II benchmarks today's models on that setting. 
A public leaderboard ranks 33 models on a held-out test split of 500 items stratified by domain, with a validation split of the same shape released for public use, from memory alone, with the court's sections provided, and as a retrieval agent over the law in force, with the cost per item and the hallucinated citations. 
The best model scores 7.78 out of 10 and the best open-weight model 6.39. Without a search tool only five open-weight models recall any Thai law, and with it 27 of the 33 models gain and six lose.
Leaderboard will take new models as they are released.
Legal benchmarks must track dates, and on the right text today's models still leave a wide gap on Thai law.
\end{abstract}

\section{Introduction}
\label{sec:intro}

Language models are already asked questions of Thai law.
Lawyers, students and the public ask which section applies to a set of facts, what the consequence is, and how a court would decide.
A benchmark that measures how well a model does this needs four things.
(1) It needs questions from real disputes, with the verdict and the reasoning a court gave.
(2) It needs the grading a Thai lawyer would apply, which scores the reasoning and not only the outcome.
(3) It needs the full statute book, so that a model can be tested as a retrieval agent and not only from memory.
And (4) it needs the version of each section that was in force when the case was decided, because the law changes and a court applies the law of its day.

No Thai legal benchmark has had the four together.
Our earlier benchmark, \nitibench{}~\citep{nitibench}, asks questions written for the Civil and Commercial Code and the Revenue Code against a 2024 snapshot of the text, and two years later 2.3\% of its citations are already outdated (\S\ref{sec:related}).
Benchmarks for English and Chinese law treat the law as a fixed text as well~\citep{guha2023legalbench,fei2024lawbench,lexam2026}.

\begin{figure}[t]
\centering
\input{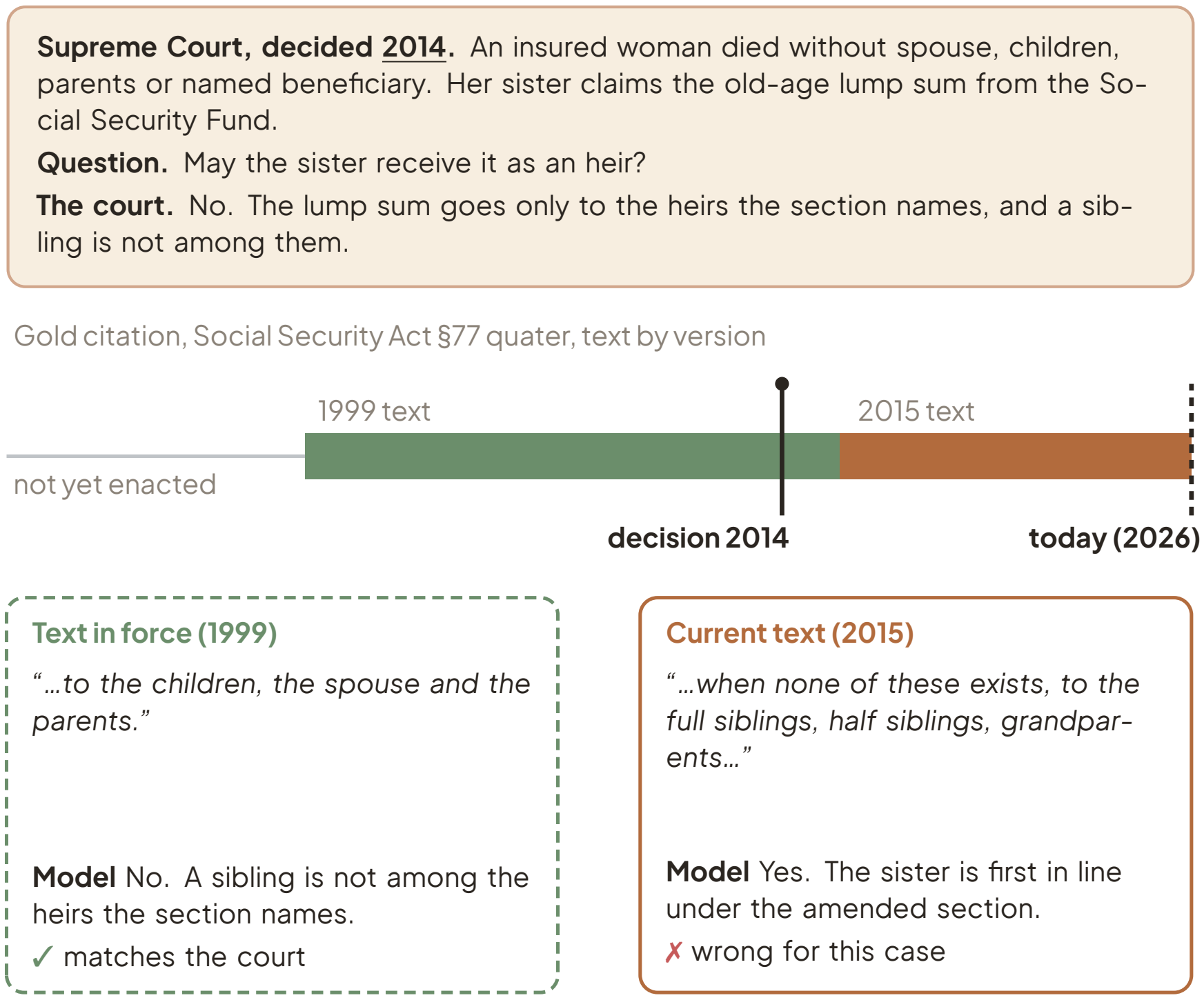}
\resizebox{0.72\textwidth}{!}{\begin{tikzpicture}\overviewCase\end{tikzpicture}}
\caption{An item of \benchmark{}. The cited section was rewritten after the decision. The same model reads the same case with the text in force on that day or with the current text.}
\label{fig:hero-case}
\end{figure}

\benchmark{}, its sequel, has all of them.
The benchmark is temporal in one sense only. Every question carries the date of its case, and the answer is graded against the law in force on that date.
It holds 1{,}670 items and 4{,}071 sub-questions from decisions of three Thai courts, in 17 domains of law from family and succession to administrative contracts (\S\ref{sec:benchmark}).
Each item gives the fact pattern and asks for the verdict, the reasoning and the sections applied, and the gold answer is copied from the decision.
Answers are graded on the Thai Bar Council rubric, from 0 to 10, which we call the Bar-standard score, and on the overlap between the sections the model cites and the sections the court relied on (\S\ref{sec:experiments}).
Every gold citation is resolved to the version in force on the decision date and paired with the current text (Figure~\ref{fig:hero-case}).
The retrieval corpus holds every version of every Thai statute section, 394{,}943 dated texts of 106{,}041 sections.

We built the benchmark for two purposes.

The first is to show that the version of the text matters, and to set up the task so that it does not distort the score.
Law is not a fixed text. Statutes are amended and repealed.
A question taken from a court decision has its answer in the text in force on that day, not in the current text.
Across 156{,}650 decisions of four Thai courts, the cited law has changed since the decision in up to 67\% of the cases by court, and of the 1{,}031 Acts Parliament has passed since 2002, 55\% amend or repeal an earlier one (\S\ref{sec:population-drift}).
Of 53 legal benchmarks from 19 jurisdictions, none grades the version of the law.
A benchmark that checks the section number but not the version rewards an answer drawn from the wrong text.
We ask what happens when a model reads the current text of a correct section, and whether searching the law in force on the decision date fixes the problem.
We ran two experiments on 13 open-weight models over the full benchmark.
First, we gave the model the correct citation and swapped the version, the text in force or the current text.
Second, we placed the model in a retrieval agent and changed its index from the current text to the text in force on the decision date.
Reading the current text of an amended section lowers the Bar-standard score by 0.25 points.
For questions on repealed sections, the current text contains no provision for the model to cite, and the score falls by 0.79 points.
Searching the law in force recovers part of these points for the models that search, and none for the models that do not (Part I, \S\ref{sec:drift}).
The text in force is the fair setting for a question taken from a dated decision, and every result in Part II uses it.

The second is to measure today's models on that setting.
Lawyers, students and the public already put legal questions to language models, and no measurement says how far the answers can be trusted.
We run a public leaderboard on a held-out test split of 500 items stratified by domain, 29 or 30 per domain, so that a new model can be added at a bounded cost, and we release a validation split of the same shape.
It ranks 33 models in three settings, from memory alone, with the court's sections provided, and as a retrieval agent over the law in force, and it reports the cost per item, the hallucinated citations and the score by domain.
The leaderboard then asks what limits the models, the knowledge of the statutes, the use of a search tool, or the application of the law to the facts, and what Thai and legal post-training adds to a base.
The leading model scores 7.78 out of 10 as a retrieval agent and the best open-weight model 6.39 (Part II, \S\ref{sec:platform}).

This report makes three contributions.
\begin{enumerate}[leftmargin=*,itemsep=2pt]
    \item \textbf{A temporal benchmark.} \benchmark{}, 1{,}670 items from three courts in 17 domains, with every gold citation resolved to the version in force on the decision date and paired with the current text, and the full dated statute corpus to retrieve from.
    \item \textbf{The cost of the version.} An audit of 156{,}650 Thai decisions, and the score 13 open-weight models lose when they read the current text instead of the text in force, by drift band and by the kind of change. The text in force is the fair setting.
    \item \textbf{A leaderboard and an analysis of today's models.} A public ranking of 33 models on a held-out 500-item test split in three settings, with citation precision and recall, hallucinated citations and the cost per item, an analysis of what limits the models, in statute knowledge, tool use and the application of the law, and a comparison with a general intelligence index, which sorts the tiers and not the frontier.
\end{enumerate}

\section{Related Work}
\label{sec:related}

\paragraph{Legal QA benchmarks.}
Legal QA benchmarks exist for English (LegalBench~\citep{guha2023legalbench}, LexGLUE~\citep{chalkidis2022lexglue}, CaseHOLD~\citep{zheng2021casehold}), Chinese (LawBench~\citep{fei2024lawbench}, LexEval~\citep{lexeval2024}), law-school examinations (LEXam~\citep{lexam2026}, KCL~\citep{kcl2026}), retrieval (LegalBench-RAG~\citep{pipitone2024legalbenchrag}), tool-using agents (LegalAgentBench~\citep{li2025legalagentbench}), and Thai law (\nitibench{}~\citep{nitibench}).
Most of them score a multiple-choice answer or a retrieved passage, and none of them grades the version of the law (\S\ref{sec:population-drift}).
\benchmark{} scores a free-form verdict and its reasoning on the rubric Thai bar examiners use, and the cited sections against the court's own.

\paragraph{Thai legal NLP.}
\benchmark{} is the sequel of \nitibench{}~\citep{nitibench}, our earlier Thai legal QA benchmark, and it is built to measure a language model on Thai law in every aspect a Thai lawyer would ask of it.
\nitibench{} has a CCL split of 3{,}729 questions on the Civil and Commercial Code and a Tax split of 50 questions from Revenue Department rulings, each written against a 2024 snapshot of the statute text.
Two years later, 2.3\% of its CCL citations no longer match the current text, and 25\% of its cited sections have been amended at some point (\S\ref{sec:population-drift}).
\benchmark{} widens every part of it. The questions come from decided cases of three courts across 17 domains instead of two codes. The answer is a free-form verdict with its reasoning, graded on the rubric Thai bar examiners use, with the cited sections checked against the court's own. Every citation is resolved to the version in force on the decision date and paired with the current text, and the retrieval corpus holds every dated version of every section, so a model can be tested from memory, with the court's sections, and as a retrieval agent. A public leaderboard keeps the measurement current as models are released.
On the model side, several Thai post-trainings of open bases have been released, among them Typhoon~\citep{typhoons}, Pathumma~\citep{pathumma}, THaLLE~\citep{thalle}, OpenThaiGPT~\citep{openthaigpt-thaillm-8b-v7p2-research-preview}, SEA-LION~\citep{sealionv48} and OpenThai 2.0 Legal~\citep{openthai2legal}.
Section~\ref{subsec:adapt} tests what they add to their bases on Thai law.

\paragraph{Versioned legal text.}
Several jurisdictions publish point-in-time statutes, including the UK, Australia, and Canada, and citation graphs exist across multiple legal systems~\citep{lawnetviz2022,lynx2021}.
No single resource combines dated statute versions, dated court rulings with resolved citations, and an evaluation benchmark.
FourCorners~\citep{akarajaradwong2026fourcorners} unifies all three for Thai law and supplies our corpus of dated statutes and rulings.
Recent studies of time in legal QA add dates to German statutory questions~\citep{prior2026oldfriend}, track versions of the French tax code~\citep{cymbler2026fiscalqa}, and compare Chinese judgments before and after the 2021 Civil Code~\citep{huang2026wronglaw}.
Each relies on a task-specific dataset.
\benchmark{} covers a national case-law corpus with the complete statute book in every version.

\paragraph{LLM judges for legal QA.}
LLM evaluators suffer from length and position bias and from variance between models~\citep{llmsjudges2024survey}.
LeMAJ~\citep{lemaj2025} breaks legal answers into atomic components.
\benchmark{} uses the official Thai Bar Council grading rubric, applied by models calibrated against human bar examination graders~\citep{akarajaradwong2026barexam}.

\section{How Much Law Drifts}
\label{sec:population-drift}

Using FourCorners~\citep{akarajaradwong2026fourcorners}, we examine drift from three angles, how often court citations become outdated, what existing benchmarks record about time, and how often Parliament amends existing statutes rather than passing new ones.
To measure court drift, we compare the statute text in force on the decision date with the text current on 1 September 2026. A cited section has \emph{drifted} if its text was amended or repealed between those two dates (Figure~\ref{fig:hero-case}). We evaluated every decision with resolved citations across four courts: the Supreme Court (51{,}566 decisions since 1935), the Constitutional Court (2{,}025 rulings since 1998), the Administrative Courts (101{,}315 judgments since 2001), and the Court of Appeal for Specialised Cases (1{,}744 decisions since 2016).

\begin{figure}[t]
    \centering
    \includegraphics[width=0.65\textwidth]{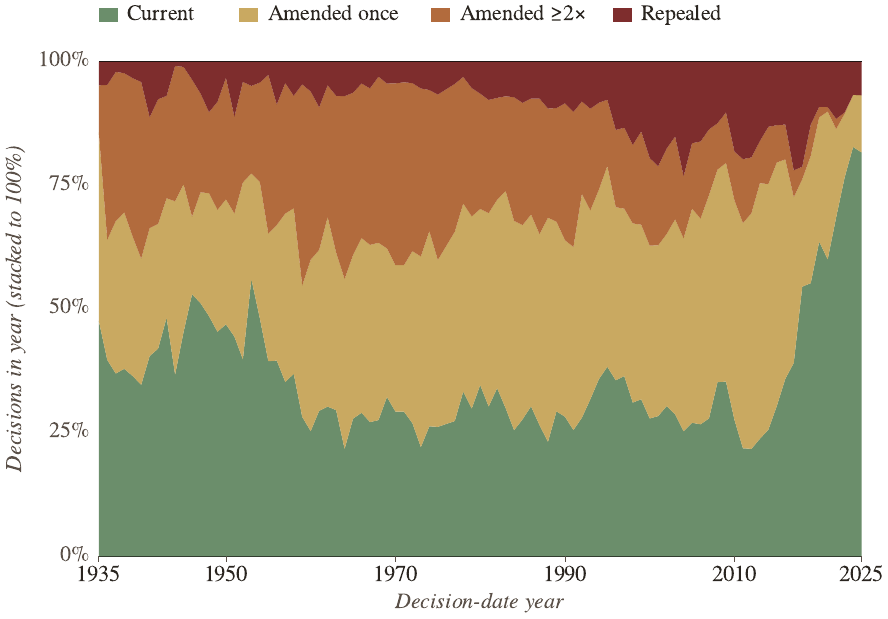}
    \caption{Drift by decision year, 1935--2025, stacked to 100\%. Each decision takes the most severe state among its citations. Totals in Table~\ref{tab:drift-by-court}.}
    \label{fig:population-drift}
\end{figure}

\begin{table}[t]
\centering
\small
\setlength{\tabcolsep}{4pt}
\begin{tabular}{l r rrr r}
\toprule
\textbf{Court} & \textbf{Decisions} & \textbf{Current} & \textbf{Amended} & \textbf{Repealed} & \textbf{Drifted} \\
\midrule
Supreme Court & 51{,}566 & 32.9 & 56.1 & 11.0 & \textbf{67.1} \\
Constitutional Court & 2{,}025 & 37.6 & 2.7 & 59.8 & \textbf{62.4} \\
Administrative Courts & 101{,}315 & 57.4 & 26.2 & 16.4 & \textbf{42.6} \\
Court of Appeal for Specialised Cases & 1{,}744 & 76.7 & 19.3 & 3.9 & \textbf{23.3} \\
\bottomrule
\end{tabular}
\caption{Decisions whose cited law has changed since the decision, by court, in percent. A decision takes the most severe state among its citations; Drifted is Amended plus Repealed.}
\label{tab:drift-by-court}
\end{table}

\paragraph{Most decisions cite law that has since changed.}
Table~\ref{tab:drift-by-court} categorizes each decision by its most severely drifted citation. Across all four courts, 79,394 of 156,650 decisions (50.7\%) cite law that has changed. \textbf{Specifically, cited law changed in 67.1\% of Supreme Court decisions, 62.4\% of Constitutional Court rulings, 42.6\% of Administrative Court judgments, and 23.3\% of specialised appeal court decisions.}
Each court drifts for different reasons
The Supreme Court relies on legal codes amended incrementally over decades.
The Constitutional Court cites constitutions replaced in 2007 and 2017 along with their organic acts.
The Administrative Courts cite general codes alongside repealed administrative laws (one in six judgments cites a repealed act, such as the 1992 Civil Service Act or the 1987 Expropriation Act).
The specialised appeal court cites commercial, labor, and family codes amended over the past decade.
This drift is not limited to older cases. As shown in Figure~\ref{fig:population-drift}, only Supreme Court decisions from the 2020s maintain over 80\% alignment with current text. For cases from the 2010s, half had a cited section rewritten within five years.

\paragraph{Existing benchmarks cannot see it.}
\begin{table}[!htbp]
\centering
\small
\setlength{\tabcolsep}{3.5pt}
\begin{tabular}{l r r r r r}
\toprule
\textbf{Released} & \textbf{Bench.} & \textbf{Cites statute} & \textbf{Item date} & \textbf{Version} & \textbf{Snapshot} \\
\midrule
before 2024 & 20 & 5 (+11) & 2 (+6) & 1 & 2 (+2) \\
2024--2025 & 19 & 9 (+5) & 1 (+5) & 3 & 3 (+2) \\
2026 & 14 & 6 (+8) & 7 (+4) & 5 (+1) & 5 (+1) \\
\midrule
All & 53 & 20 (+24) & 10 (+15) & 9 (+1) & 10 (+5) \\
\bottomrule
\end{tabular}
\caption{Can a legal benchmark test anything but current law? 53 legal NLP benchmarks from 19 jurisdictions, audited from their dataset cards and papers (Table~\ref{tab:benchmark-audit-full} lists every benchmark). Testing law in force needs the version in the gold and in the corpus. \emph{Cites statute}: gold answers cite statute sections. \emph{Item date}: items carry the date of the underlying decision, examination or question. \emph{Version}: gold answers identify which version of the statute they rely on. \emph{Snapshot}: the paper states the date of the statute text it used. Counts in parentheses are partial cases.}
\label{tab:benchmark-audit}
\end{table}

We checked 53 legal benchmarks across 19 jurisdictions for what they record about time (Table~\ref{tab:benchmark-audit}).
Of the 53, 44 cite statutes in their gold answers, 10 give the date of every item, 9 carry a statute version, and none grades a model on the version. Every reported score therefore reflects the current law, even for a benchmark built from past court decisions.
Table~\ref{tab:benchmark-audit-full} lists the 53 benchmarks. Each cell was read from the dataset's schema, its data files or its paper on 5 October 2026. The audit file released with the code holds, for every cell, the field names or the quoted sentence it rests on and a confidence grade. Two rows are graded low confidence because only the paper was available. In total, 44 benchmarks cite statutes in their gold answers (20 fully), 10 give full item dates and 15 partial ones, 9 carry a statute version, 5 of them the concurrent temporal studies, and 10 state a snapshot date.

\begin{table}[p]
\centering
\scriptsize
\setlength{\tabcolsep}{2.5pt}
\renewcommand{\arraystretch}{0.95}
\begin{tabular}{l r l c c c c}
\toprule
\textbf{Benchmark} & \textbf{Year} & \textbf{Jurisdiction} & \textbf{Cites} & \textbf{Date} & \textbf{Ver.} & \textbf{Snap.} \\
\midrule
\multicolumn{7}{l}{\textit{before 2024}} \\
CAIL2018~\citep{xiao2018cail} & 2018 & China & \checkmark & -- & -- & -- \\
JEC-QA~\citep{zhong2020jecqa} & 2020 & China & -- & -- & -- & -- \\
SARA~\citep{holzenberger2020sara} & 2020 & US & \checkmark & \checkmark & -- & -- \\
CaseHOLD~\citep{zheng2021casehold} & 2021 & US & ($\checkmark$) & -- & -- & -- \\
COLIEE statute law tasks~\citep{rabelo2022coliee} & 2021 & Japan & \checkmark & ($\checkmark$) & -- & \checkmark \\
CUAD~\citep{hendrycks2021cuad} & 2021 & US & -- & ($\checkmark$) & -- & -- \\
ECtHR datasets~\citep{chalkidis2021ecthr} & 2021 & Council of Europe & ($\checkmark$) & -- & -- & -- \\
ILDC~\citep{malik2021ildc} & 2021 & India & -- & ($\checkmark$) & -- & -- \\
Zalo AI Challenge 2021 Legal Text ~\citep{zalo2021legal} & 2021 & Vietnam & \checkmark & -- & -- & -- \\
BSARD~\citep{louis2022bsard} & 2022 & Belgium & \checkmark & -- & -- & \checkmark \\
LBOX OPEN~\citep{hwang2022lboxopen} & 2022 & Korea & ($\checkmark$) & -- & -- & -- \\
LexGLUE~\citep{chalkidis2022lexglue} & 2022 & Council of Europe, EU, & ($\checkmark$) & -- & -- & -- \\
NCBE MBE practice exam set used in~\citep{katz2024gpt4bar} & 2022 & US & -- & -- & -- & ($\checkmark$) \\
DISC-Law-Eval~\citep{yue2023disclawllm} & 2023 & China & ($\checkmark$) & ($\checkmark$) & -- & -- \\
LAiW~\citep{laiw2025} & 2023 & China & ($\checkmark$) & -- & -- & -- \\
LawBench~\citep{fei2024lawbench} & 2023 & China & ($\checkmark$) & -- & -- & -- \\
LegalBench~\citep{guha2023legalbench} & 2023 & US & ($\checkmark$) & ($\checkmark$) & -- & -- \\
LeXFiles / LegalLAMA~\citep{chalkidis2023lexfiles} & 2023 & Canada, Council of Eur & ($\checkmark$) & -- & -- & -- \\
Open Australian Legal QA~\citep{butler2023openausqa} & 2023 & Australia & ($\checkmark$) & ($\checkmark$) & \checkmark & ($\checkmark$) \\
SCALE / Swiss legal benchmark suit~\citep{niklaus2024scale} & 2023 & Switzerland & ($\checkmark$) & \checkmark & -- & -- \\
\multicolumn{7}{l}{\textit{2024--2025}} \\
ArabLegalEval~\citep{arablegaleval2024} & 2024 & Saudi Arabia & ($\checkmark$) & -- & -- & -- \\
IL-TUR~\citep{joshi2024iltur} & 2024 & India & \checkmark & ($\checkmark$) & -- & -- \\
KBL: Korean Benchmark for Legal la~\citep{kbl2024} & 2024 & Korea & ($\checkmark$) & ($\checkmark$) & -- & \checkmark \\
LegalBench-RAG~\citep{pipitone2024legalbenchrag} & 2024 & US & -- & -- & -- & -- \\
LeKUBE~\citep{wang2024lekube} & 2024 & China & \checkmark & -- & \checkmark & -- \\
LexEval~\citep{lexeval2024} & 2024 & China & ($\checkmark$) & -- & \checkmark & -- \\
NyayaAnumana~\citep{nigam2025nyayaanumana} & 2024 & India & -- & ($\checkmark$) & -- & -- \\
PredEx~\citep{nigam2024predex} & 2024 & India & -- & -- & -- & -- \\
STARD~\citep{su2024stard} & 2024 & China & \checkmark & -- & -- & -- \\
ALQAC~\citep{alqac2025} & 2025 & Vietnam & \checkmark & -- & -- & -- \\
Bar Exam QA~\citep{zheng2025reasoning} & 2025 & US & ($\checkmark$) & -- & -- & ($\checkmark$) \\
CLaw~\citep{claw2025} & 2025 & China & \checkmark & ($\checkmark$) & \checkmark & ($\checkmark$) \\
Housing Statute QA~\citep{zheng2025reasoning} & 2025 & US & \checkmark & ($\checkmark$) & -- & \checkmark \\
JBE-QA~\citep{jbeqa2025} & 2025 & Japan & -- & \checkmark & -- & -- \\
JuDGE~\citep{su2025judge} & 2025 & China & \checkmark & -- & -- & -- \\
LaborBench~\citep{hariri2025laborbench} & 2025 & US & -- & -- & -- & \checkmark \\
LegalAgentBench~\citep{li2025legalagentbench} & 2025 & China & ($\checkmark$) & -- & -- & -- \\
LexRAG~\citep{li2025lexrag} & 2025 & China & \checkmark & -- & -- & -- \\
NitiBench~\citep{nitibench} & 2025 & Thailand & \checkmark & -- & -- & -- \\
\multicolumn{7}{l}{\textit{2026}} \\
BenGER~\citep{nagl2026benger} & 2026 & Germany & ($\checkmark$) & ($\checkmark$) & -- & -- \\
CALRK-Bench~\citep{jung2026calrk} & 2026 & Korea & ($\checkmark$) & ($\checkmark$) & -- & -- \\
ClaimRAG-LAW~\citep{das2026claimrag} & 2026 & EU, Luxembourg & ($\checkmark$) & -- & -- & ($\checkmark$) \\
FiscalQA Pro~\citep{cymbler2026fiscalqa} & 2026 & France & \checkmark & \checkmark & \checkmark & \checkmark \\
GreekBarRetrieval~\citep{beta2026greekbarretrieval} & 2026 & Greece & \checkmark & \checkmark & -- & \checkmark \\
Legal RAG Bench~\citep{butler2026legalragbench} & 2026 & Australia & ($\checkmark$) & -- & -- & -- \\
Legal Research Bench~\citep{drozdov2026legalresearchbench} & 2026 & US & ($\checkmark$) & ($\checkmark$) & -- & -- \\
LegalSearch-R1 benchmark~\citep{fan2026legalsearchr1} & 2026 & China & \checkmark & \checkmark & \checkmark & \checkmark \\
LEXam~\citep{lexam2026} & 2026 & Switzerland & ($\checkmark$) & \checkmark & -- & -- \\
LexKairos~\citep{li2026lexkairos} & 2026 & China & ($\checkmark$) & \checkmark & \checkmark & -- \\
Prior et al. (German statutory QA)~\citep{prior2026oldfriend} & 2026 & Germany & \checkmark & ($\checkmark$) & ($\checkmark$) & -- \\
SearchFireSafety~\citep{chae2026beyondcaselaw} & 2026 & Korea & \checkmark & -- & -- & \checkmark \\
TALD (applicable-law determination~\citep{huang2026wronglaw} & 2026 & China & \checkmark & \checkmark & \checkmark & -- \\
TIDE~\citep{sobhani2026tide} & 2026 & Bangladesh & ($\checkmark$) & \checkmark & \checkmark & \checkmark \\
\bottomrule
\end{tabular}
\caption{The 53 legal NLP benchmarks in the audit, by release period. \checkmark{} yes, ($\checkmark$) in part, -- no. Cites, gold cites statute sections. Date, items carry a date. Ver., gold names the statute version. Snap., the paper states when its statute text was taken.}
\label{tab:benchmark-audit-full}
\end{table}

\paragraph{The law itself keeps moving.}
Legislative records confirm this pattern. Since 2002, Parliament published 1,031 Acts in the Royal Gazette. Of these, 567 (55\%) amended or repealed existing statutes rather than introducing new ones. This proportion rose from 53\% in the 2000s to 65\% in the 2020s. Over the same period, the executive issued 3,299 royal decrees and 4,440 ministerial regulations. Because most legislative activity updates existing law, static benchmarks based on snapshot statutes quickly fall out of sync.

\paragraph{\nitibench{} has already drifted.}
\begin{table}[!ht]
\centering
\small
\setlength{\tabcolsep}{4pt}
\begin{tabular}{l r r r}
\toprule
\textbf{Benchmark} & \textbf{Questions} & \textbf{Amended ever} & \textbf{Stale} \\
\midrule
NitiBench-CCL & 3,729 & 25\% & 2.3\% \\
NitiBench-Tax & 50 & 72\% & 0.0\% \\
\bottomrule
\end{tabular}
\caption{Temporal drift in NitiBench two years after release. Amended ever: percentage of cited sections amended at least once. Stale: percentage of citations whose embedded text is no longer current.}
\label{tab:external-audit}
\end{table}

Without version tracking, benchmarks cannot detect when their gold answers expire (\S\ref{sec:related}). We evaluated our own earlier benchmark, NitiBench~\citep{nitibench}, whose questions include the statutory text used during annotation (Table~\ref{tab:external-audit}):
NitiBench-CCL was built on 2024 statutes. Two years later, 2.3\% of its citations are outdated, including Section 1435 of the Civil and Commercial Code (marriage age, amended by the 2025 marriage equality act). Over 25\% of its cited sections have been amended previously, and errors will accumulate undetected.
To avoid drift, NitiBench-Tax sampled only post-2021 Revenue Department rulings that match the latest Revenue Code. While this preserved temporal accuracy, it shrank the benchmark to just 50 questions.
\benchmark{} avoids this trade-off by preserving decision dates and indexing both historical and current texts.

\section{The \benchmark{} Benchmark}
\label{sec:benchmark}

\benchmark{} consists of 1{,}670 items and 4{,}071 sub-questions spanning 17 Thai legal domains, sourced from three courts (Table~\ref{tab:domain-inventory}).
The Supreme Court provides private, criminal, and regulatory domains (1{,}303 items).
The Constitutional Court provides public constitutional law (50 items).
The Supreme Administrative Court provides administrative domains covering official acts, state contracts, public liability, and personnel (317 items).

Domains reflect the core areas of each court's reasoning. We preserved major fields, split standalone statutory areas (such as narcotics and traffic), and merged smaller categories. Each domain contains 56 to 146 items. Benchmark summary scores are equal-weighted macro-averages across domains.
Because judicial decisions do not follow question-answer formats, an extraction pipeline converts each decision into a fact pattern, sub-questions, and statutory provisions (\S\ref{sec:case-pipeline}). Gold verdicts and reasoning are preserved as verbatim excerpts from the decisions.

\begin{table}[!ht]
\centering
\small
\setlength{\tabcolsep}{4pt}
\resizebox{\textwidth}{!}{\begin{tabular}{l r r r r r r r r r r}
\toprule
& & & \multicolumn{3}{c}{\textbf{Drift band (items)}} & \multicolumn{5}{c}{\textbf{Sub-questions by type (court's own + four added kinds)}} \\
\cmidrule(lr){4-6}\cmidrule(lr){7-11}
\textbf{Domain} & \textbf{Items} & \textbf{Sub-q.} & \textbf{No drift} & \textbf{Amended} & \textbf{Repealed} & \textbf{Court's own} & \textbf{Governs} & \textbf{Conseq.} & \textbf{Combine} & \textbf{Lower ct.} \\
\midrule
Family \& Succession & 146 & 326 & 60 & 85 & 1 & 153 & 15 & 61 & 57 & 40 \\
Criminal & 146 & 352 & 13 & 106 & 27 & 157 & 39 & 52 & 43 & 61 \\
Civil \& Commercial & 144 & 351 & 63 & 60 & 21 & 153 & 31 & 63 & 56 & 48 \\
Land \& Property & 117 & 284 & 43 & 42 & 32 & 123 & 22 & 56 & 50 & 33 \\
Labor & 116 & 239 & 41 & 63 & 12 & 117 & 12 & 37 & 35 & 38 \\
Tax \& Revenue & 113 & 290 & 18 & 47 & 48 & 128 & 25 & 53 & 52 & 32 \\
Intellectual Property & 109 & 255 & 36 & 48 & 25 & 116 & 23 & 44 & 41 & 31 \\
Public Law & 98 & 263 & 27 & 26 & 45 & 119 & 26 & 47 & 55 & 16 \\
Civil Procedure & 87 & 180 & 13 & 61 & 13 & 91 & 11 & 26 & 17 & 35 \\
Natural Resources \& Licensing & 84 & 173 & 15 & 36 & 33 & 87 & 11 & 31 & 28 & 16 \\
Administrative Acts & 82 & 233 & 36 & 34 & 12 & 99 & 32 & 29 & 63 & 10 \\
Public Personnel & 81 & 244 & 32 & 21 & 28 & 98 & 31 & 36 & 55 & 24 \\
State Liability & 80 & 241 & 36 & 21 & 23 & 96 & 20 & 43 & 53 & 29 \\
Narcotics & 74 & 159 & 3 & 2 & 69 & 78 & 18 & 20 & 13 & 30 \\
Administrative Contracts & 74 & 227 & 28 & 31 & 15 & 82 & 20 & 49 & 41 & 35 \\
Traffic \& Transport & 63 & 117 & 8 & 48 & 7 & 64 & 8 & 19 & 11 & 15 \\
Securities \& Finance & 56 & 137 & 15 & 28 & 13 & 59 & 10 & 25 & 25 & 18 \\
\midrule
\textbf{Total} & \textbf{1{,}670} & \textbf{4{,}071} & 487 & 759 & 424 & 1{,}820 & 354 & 691 & 695 & 511 \\
\bottomrule
\end{tabular}}
\caption{Items and sub-questions per domain, by drift band and question type. Drift band is the most severe state among the citations an item relies on. Court's own are the questions the court answers. The four other kinds are defined in \S\ref{sec:case-pipeline}.}
\label{tab:domain-inventory}
\end{table}

\paragraph{Citations and sub-questions.}
Each item contains one or more sub-questions with its own verdict, reasoning, and citations. Dependent sub-questions are flagged and answered sequentially. Gold citations are resolved via FourCorners~\citep{akarajaradwong2026fourcorners} to the version active on the ruling date. Citations mentioned only for background context are kept in the record but ignored during scoring. Figure~\ref{fig:hero-case} illustrates an item with both texts for a drifted citation. Appendix~\ref{sec:examples} contains the full schema and a worked item.

\subsection{Construction pipeline}
\label{sec:case-pipeline}

The pipeline extracts facts, reasoning, and citations directly from the court's decision.

\paragraph{Selection.}
Candidate decisions are stratified by decade and drift status. Claude Opus 5 screens candidates, keeping challenging cases where the court applies at least three substantive statutory provisions. Fact-only disputes and procedural rulings are excluded. Duplicate holdings are grouped so no question appears more than twice, maximizing question diversity.

\paragraph{Extraction.}
Claude Fable 5 generates structured records from selected decisions. Each record contains the fact pattern, decision date, sub-questions with corresponding verdicts and reasoning, and exact source spans. Citations are classified as either primary or background. In addition to direct judicial rulings, we create four targeted question types:
(1) Identifying governing provisions.
(2) Determining legal consequences.
(3) Interpreting interactions between multiple provisions.
(4) Identifying departures from lower court decisions.
All gold answers remain verbatim text spans. Table~\ref{tab:domain-inventory} provides full item counts.

\paragraph{Verification.}
Deterministic checks confirm that text spans match the source records and citations are well-formed. Citations are resolved to the version active on the ruling date; items with unresolvable primary citations are removed.
Citations to external materials like ministerial notifications remain unscored, and joined cases are consolidated.
Finally, human annotators review every item against the original ruling to verify legal accuracy.

\subsection{Drift bands}
\label{sec:temporal-design}

Decisions span from 1949 to 2026. Items are assigned to one of three drift bands based on their most heavily altered citation:
\textbf{No-drift (Control):} 487 items (29\%) where cited statutes remain unchanged.
\textbf{Amended:} 759 items (46\%) where statutory text has been modified.
\textbf{Repealed:} 424 items (25\%) where the cited statute was subsequently repealed, or was already repealed at the time of the ruling (Appendix~\ref{app:drift-band-oracle}).
Because a single changed citation shifts an entire item out of the control band, 71\% of items exhibit drift even though 65\% of individual citations remain current (Table~\ref{tab:domain-inventory}).

\subsection{Current and In-force corpora}
\label{sec:two-corpora}

The retrieval corpus indexes 394{,}943 texts across 106{,}041 sections in 5{,}912 laws. Each section includes effective dates and operational status as of 1 September 2026. Across all benchmark items, 9{,}223 citations reference 3{,}006 sections across 298 laws. Each citation includes both its in-force and modern text:
5{,}869 citations are unchanged between versions.
1{,}934 citations are amended text.
1{,}263 citations are repealed laws with no modern counterpart.

We evaluate models across two statutory corpora.
The \textbf{Current corpus} reflects statutory text on 1 September 2026, matching traditional benchmark setups.
The \textbf{In-force corpus} reflects the statutory text active on each item's decision date.
These corpora differ only for amended and repealed sections. Experiments in \S\ref{sec:settings} supply models with cited text from one corpus or grant access to a retrieval index built over it.

\subsection{Release}
\label{sec:release}
The benchmark is released in three parts, defined in \S\ref{sec:split}.
The validation split, 498 cases, is public. For each case it gives the facts, the questions, the court's verdict and reasoning, and the gold citations with the version in force on the decision date and the current version, together with the text of those sections in both versions and the grading instruments, the rubric, the judge prompt and the citation-matching rules.
The test split, 500 cases, is held out and is never published, neither its ids nor its text. The leaderboard runs on it, and a model joins the leaderboard when we run it. We hold it out so that no model can be trained or tuned on the questions that rank it. The Artificial Analysis Intelligence Index v5 takes the same approach, with 60\% of its index resting on private test sets or held-out answers~\citep{artificialanalysis2026v5}.
The remaining 672 cases are kept in reserve.
The statute corpus, its index, the retrieval service and the harness are not released. A user of the validation split can run the Parametric and Gold-cited In-force settings and not the retrieval settings.
Every released file carries a canary string so that it can be filtered from training corpora.

\section{Experimental Setup}
\label{sec:experiments}

\subsection{Metrics}
\label{sec:metrics}

We report one score and one diagnostic.
The Bar-standard score is the main metric. It grades the whole answer as a Thai bar examiner would, the verdict, the reasoning and the sections applied.
Citation F1 is a diagnostic of one part of that answer, the statute references. It takes the sections the answer cites and compares them with the sections the court relied on.
The judge reads the answer with its citations, and the citations are therefore already graded inside the Bar-standard score. Citation F1 measures that one part on its own. We do not add the two into one number, because that would count the citations twice.
We show Citation F1 next to the score because the two can move apart, and the gap is informative. The current text of a changed law lowers the score and leaves Citation F1 almost unchanged (\S\ref{subsec:analysis}). A model can cite more sections than the court did and still score in the good tier (\S\ref{subsec:leaderboard}). Within a model, the citations account for 13 percent of the differences between answers, and the rest is the reasoning (\S\ref{par:cite}).

\begin{table}[!htb]
    \centering
    
    \small
    \setlength{\tabcolsep}{4pt}
    \begin{tabular}{@{}l c c c c c@{}}
    \toprule
    \headrow & \multicolumn{5}{c}{\textbf{Reasoning quality}} \\
    \headrow \textbf{Final answer} & Unusable & Some & Acceptable & Good & Very good \\
    \midrule
    \emph{matches gold}    & T4: $0$--$1$  & T5: $2$--$5$  & T6: $5$--$6$  & T7: $7$--$8$ & T8: $9$--$10$ \\
    \emph{differs}         & T1: $0$       & T2: $1$--$2$  &               & T3: $2$--$4$ &     \\
    \bottomrule
    \end{tabular}
\caption{The Thai Bar Council eight-tier rubric (used since 2014).}
    \label{tab:rubric}
\end{table}

\paragraph{Bar-standard score.}
The Bar-standard score is the grade of a free-form answer on the official Thai Bar Council grading rubric (Table~\ref{tab:rubric}, Appendix~\ref{app:rubric}). The rubric is the instrument and the score is its result. The rubric combines verdict accuracy with five tiers of reasoning quality.
A matching verdict (T4 to T8) scores from 0 to 10 points.
An incorrect verdict (T1 to T3) is capped at 4 points.
We call the resulting 0 to 10 grade the Bar-standard score.
Following \citet{akarajaradwong2026barexam}, we use a panel of three LLM judges (Claude 4.6 Opus, Gemini 3.1 Pro Preview, and GPT-5.4 Pro) evaluated against human bar examination standards. Each judge receives the verbatim rubric instructions (Appendix~\ref{app:judge-prompt}), and we report the median score across the panel.

\paragraph{Citation F1.}
Gold citations include all statutory sections required for a sub-question's verdict. Model citations are resolved to the statute corpus, and background citations mentioned in the ruling are excluded from the predicted set before scoring. We compute Citation F1 per sub-question as the harmonic mean of precision and recall (0 if no sections are cited). Precision and recall breakdowns appear in Table~\ref{tab:prf}.

\paragraph{Aggregation.}
Scores are computed hierarchically. Sub-questions are averaged within each item, items are averaged within each domain, and the overall benchmark score is the macro-average across all 17 domains.

\subsection{Settings}
\label{sec:settings}

We test five experimental configurations across all items:
\begin{enumerate}[leftmargin=*,itemsep=2pt]
    \item \textbf{Parametric.} The model receives only the question and decision date, testing pretraining recall.

    \item \textbf{Gold-cited Current.} The prompt includes the question, decision date, and verbatim text of gold-cited sections on current version.

    \item \textbf{Gold-cited In-force.} Identical to Gold-cited Current, but instead uses statutory text active on the decision date, including repealed provisions (serving as the upper bound for perfect retrieval).

    \item \textbf{RAG Current.} An agent retrieves sections using a dense index of current statutes via Qwen3-Embedding-8B~\citep{qwen3embed2025}, chosen for Thai following SEA-BED~\citep{seabed2025}. The tool returns the top 10 matches, and the model controls the tool freely.

    \item \textbf{RAG In-force.} Identical to RAG Current, but the evaluation harness filters retrieval candidates by the decision date. The model never selects dates or versions directly, isolating the impact of the retrieved text.
\end{enumerate}
The public leaderboard labels Parametric, Gold-cited In-force and RAG In-force as Closed book, Handout and Open book. This report keeps the names above.

\subsection{Models}
\label{sec:models}

\paragraph{Open-weight models.}
We evaluate 20 open-weight models.
Thirteen are served locally with vLLM~\citep{vllm} on four A100 80GB GPUs (Table~\ref{tab:model-configs} lists checkpoints and serving configuration, Table~\ref{tab:cost} the compute), in three groups.
General instruction models are Gemma 4 31B and Gemma 4 26B-A4B~\citep{gemma4}, Nemotron 3.5 Lightning 30B-A3B~\citep{nemotron3}, and Muse Glimmer 30B~\citep{museglimmer}.
The Nemotron 3 post-training family holds Nano 30B-A3B and Super 120B-A12B~\citep{nemotron3}, their SEA-LION v4.8 derivatives~\citep{sealionv48}, and OpenThai 2.0 Legal~\citep{openthai2legal}, which is based on Nemotron 3 Nano Omni 30B-A3B Reasoning~\citep{nvidia2026nemotron3nanoomni}.
The Thai specialized models are the four ThaiLLM 8B models, Typhoon-S~\citep{typhoons}, Pathumma~\citep{pathumma}, THaLLE~\citep{thalle} and OpenThaiGPT~\citep{openthaigpt-thaillm-8b-v7p2-research-preview}. Each is a post-training of ThaiLLM-8B~\citep{thaillm8b}, a Thai continued pretraining of Qwen3-8B-Base~\citep{qwen3technicalreport} that has no chat template and cannot answer the benchmark itself.
The other seven are Qwen3.8 27B~\citep{qwen38}, OpenThai 2.0 27B~\citep{openthai2_27b}, DeepSeek V4.1 Flash~\citep{deepseekv41}, DeepSeek V4 Flash~\citep{deepseekai2026deepseekv4}, GLM 5.3 Flash~\citep{glm5team2026glm5vibecodingagentic,zai2026glm53flash}, Qwen3-8B~\citep{qwen3technicalreport}, Alibaba's own post-training of Qwen3-8B-Base, the base the ThaiLLM 8B models descend from, which stands in for their unservable base in the comparisons of \S\ref{subsec:limits}, and Nemotron 3 Nano Omni 30B-A3B~\citep{nvidia2026nemotron3nanoomni}, NVIDIA's multimodal model built on the Nemotron 3 Nano backbone and the base of OpenThai 2.0 Legal.

\paragraph{Proprietary models.}
Thirteen models are available only through their providers' APIs. Claude Haiku 5.5~\citep{anthropic2026haiku55}, Claude Sonnet 5.5~\citep{anthropic2026sonnet55}, Claude Opus 5.5~\citep{anthropic2026opus55}, Claude Fable 5.1~\citep{anthropic2026fable51} and Claude Fable 5.2~\citep{anthropic2026fable52}, Gemini 4 Argon~\citep{google2026gemini4argon}, Gemini 4 Carbon~\citep{google2026gemini4carbon} and Gemini 4 Hydrogen~\citep{google2026gemini4hydrogen}, GPT-6.1 Sol~\citep{openai2026gpt61sol} and GPT-6.1 Terra~\citep{openai2026gpt61terra}, Muse Spark 1.3~\citep{meta2026musespark13} and Muse Ember~\citep{meta2026museember}, and GLM 5.3 FlashX Legal~\citep{zai2026glm53flashxlegal}, Z.ai's legal post-training of GLM 5.3 Flash.

\paragraph{Serving and sampling.}
Every model receives the same prompts and the same retrieval tool.
A locally served model runs with the defaults shipped in its repository (\texttt{generation\_config.json}). Where that file sets none, vLLM's own defaults apply (temperature 1.0, top-p 1.0). No model receives settings of its own. A model behind an API runs with its provider's defaults.
No prompt is tuned per model.

\subsection{Two protocols}
\label{sec:protocols}
Every table caption names the protocol it follows.
The benchmark protocol runs the 13 locally served open-weight models on all 1{,}670 items, in the five settings, with $k = 3$ samples per item, and reports \textbf{avg@3} (additional results in Appendix~\ref{app:results}). Part I follows it.
The leaderboard protocol runs all 33 models on the held-out 500-item test split (\S\ref{sec:split}), in Parametric, Gold-cited In-force and RAG In-force. The 13 models of the benchmark protocol keep their three samples there, and Qwen3-8B and Nemotron 3 Nano Omni have three as well. Every other row has one sample per item.

\section{Part I. The Cost of the Current Text}
\label{sec:drift}

Part I asks whether the version of the text matters.
It follows the benchmark protocol, 13 open-weight models, 1{,}670 items, three samples per item. It first reports the 13 models in the five settings, the base of every comparison that follows. It then swaps the version of the cited sections, the text in force against the current text, and measures what a model loses when it reads the current text of a law that changed after the decision (\S\ref{sec:population-drift}). It asks whether a search tool over the law in force gives the loss back, and whether the loss comes from the rule or from the wording. The answer sets the setting that Part II uses.

\begin{table}[t]
\centering
\small
\setlength{\tabcolsep}{3.5pt}
\resizebox{\textwidth}{!}{%
\begin{tabular}{l ccccc c ccccc}
\toprule
& \multicolumn{5}{c}{\textbf{Citation F1}} && \multicolumn{5}{c}{\textbf{Bar-standard (0--10)}} \\
\cmidrule(lr){2-6} \cmidrule(lr){8-12}
\textbf{Model} & \textbf{Param.} & \textbf{GC Cur.} & \textbf{GC InF.} & \textbf{RAG Cur.} & \textbf{RAG InF.} && \textbf{Param.} & \textbf{GC Cur.} & \textbf{GC InF.} & \textbf{RAG Cur.} & \textbf{RAG InF.} \\
\midrule
\multicolumn{12}{l}{\textit{General}} \\
Gemma 4 31B & \textbf{0.304} & 0.645 & 0.778 & \underline{0.432} & \underline{0.512} && \textbf{3.68} & \underline{4.60} & \underline{5.11} & \underline{4.20} & \underline{4.55} \\
Gemma 4 26B-A4B & \underline{0.272} & 0.667 & 0.791 & 0.228 & 0.262 && \underline{3.42} & 4.37 & 4.79 & 3.61 & 3.71 \\
Nemotron 3.5 Lightning 30B-A3B & 0.028 & 0.655 & 0.770 & 0.287 & 0.354 && 1.76 & 3.08 & 3.44 & 2.26 & 2.47 \\
Muse Glimmer 30B & 0.198 & \textbf{0.742} & \textbf{0.878} & \textbf{0.456} & \textbf{0.551} && 3.34 & \textbf{5.18} & \textbf{5.75} & \textbf{4.39} & \textbf{4.94} \\
\midrule
\multicolumn{12}{l}{\textit{Nemotron 3 bases and their adaptations}} \\
Nemotron 3 Nano 30B-A3B & 0.020 & 0.650 & 0.762 & 0.205 & 0.260 && 1.22 & 2.44 & 2.71 & 1.52 & 1.66 \\
\hspace{0.9em}$\hookrightarrow$\,SEA-LION v4.8 30B-A3B & 0.038 & 0.644 & 0.753 & 0.283 & 0.347 && 1.69 & 3.16 & 3.49 & 2.25 & 2.50 \\
OpenThai 2.0 Legal 30B-A3B & 0.073 & 0.427 & 0.474 & 0.004 & 0.005 && 1.16 & 1.84 & 1.95 & 0.88 & 0.88 \\
Nemotron 3 Super 120B-A12B & 0.099 & \underline{0.691} & \underline{0.814} & 0.358 & 0.444 && 2.08 & 3.79 & 4.27 & 2.92 & 3.31 \\
\hspace{0.9em}$\hookrightarrow$\,SEA-LION v4.8 120B-A12B & 0.092 & 0.677 & 0.801 & 0.305 & 0.362 && 2.05 & 3.78 & 4.26 & 2.71 & 2.94 \\
\midrule
\multicolumn{12}{l}{\textit{ThaiLLM 8B derivatives}} \\
Typhoon-S 8B & 0.096 & 0.604 & 0.687 & 0.013 & 0.036 && 1.73 & 2.62 & 2.79 & 1.61 & 1.56 \\
Pathumma 8B Think & 0.140 & 0.592 & 0.669 & 0.124 & 0.161 && 1.71 & 2.61 & 2.70 & 1.40 & 1.47 \\
THaLLE 0.2 8B & 0.022 & 0.611 & 0.713 & 0.001 & 0.002 && 1.21 & 2.34 & 2.61 & 1.11 & 1.11 \\
OpenThaiGPT 8B & 0.020 & 0.567 & 0.667 & 0.000 & 0.000 && 0.94 & 1.79 & 1.95 & 0.88 & 0.88 \\
\bottomrule
\end{tabular}
}
\caption{Citation F1 and Bar-standard score (0--10) by model and setting, avg@3 over 1{,}670 items. GC is Gold-cited. Cur.\ and InF.\ are the current text and the text in force. Best per column in bold, second best underlined. Spread across the three samples in Table~\ref{tab:sample-sd}.}
\label{tab:main-results}
\end{table}

\subsection{Open-weight models on the full benchmark}
\label{subsec:overall}
Table~\ref{tab:main-results} reports the Bar-standard score and Citation F1 for the 13 open-weight models in the five settings.
Gold-cited In-force provides the historically accurate text and is the best setting for every model.
Citation F1 ranges from 0.47 to 0.88, and Bar-standard scores range from 2.0 to 5.8 out of 10.
Without provided text (Parametric), only Gemma 4 and Muse Glimmer can recall some laws from memory, reaching a maximum Citation F1 of 0.30.
Even with the correct sections provided, no model exceeds 5.8 in Bar-standard score.
The sections supply the correct law. Applying that law to the facts remains the major challenge, and leaves substantial headroom on \benchmark{}.

RAG In-force is the setting closest to a deployed assistant, a model with a search tool over the law in force.
Scores there run from 0.88 to 4.94, with a mean of 2.45 over the 13 models.
Muse Glimmer 30B leads with 4.94, Gemma 4 31B follows with 4.55, then Gemma 4 26B-A4B with 3.71 and Nemotron 3 Super with 3.31.
The four ThaiLLM 8B models and OpenThai 2.0 Legal score below 1.6.
Citation F1 follows the same order, 0.551 for Muse Glimmer and 0.512 for Gemma 4 31B, and no 8B model reaches 0.17.
The spread across the three samples is 0.00 to 0.09 points on this setting (Table~\ref{tab:sample-sd}), so the ranking is stable.
The best of the three samples per item is in Table~\ref{tab:best3} (Appendix~\ref{app:bestk}).

\begin{table}[!htbp]
\centering
\small
\setlength{\tabcolsep}{4pt}
\begin{tabular}{l cc c cc}
\toprule
& \multicolumn{2}{c}{\textbf{Gold-cited}} && \multicolumn{2}{c}{\textbf{RAG}} \\
\cmidrule(lr){2-3} \cmidrule(lr){5-6}
\textbf{Model} & \textbf{Amend.} & \textbf{Repeal.} && \textbf{Amend.} & \textbf{Repeal.} \\
\midrule
\multicolumn{6}{l}{\textit{General}} \\
Gemma 4 31B & $+0.45^{*}$ & $+0.98^{*}$ && $+0.29^{*}$ & $+0.87^{*}$ \\
Gemma 4 26B-A4B & $+0.31^{*}$ & $+0.97^{*}$ && $+0.03$ & $+0.35^{*}$ \\
Nemotron 3.5 Lightning 30B-A3B & $+0.31^{*}$ & $+0.86^{*}$ && $+0.14^{*}$ & $+0.60^{*}$ \\
Muse Glimmer 30B & $+0.45^{*}$ & $+1.35^{*}$ && $+0.32^{*}$ & $+1.51^{*}$ \\
\midrule
\multicolumn{6}{l}{\textit{Nemotron 3 bases and their adaptations}} \\
Nemotron 3 Nano 30B-A3B & $+0.22^{*}$ & $+0.58^{*}$ && $+0.10^{*}$ & $+0.38^{*}$ \\
\hspace{0.9em}$\hookrightarrow$\,SEA-LION v4.8 30B-A3B & $+0.29^{*}$ & $+0.88^{*}$ && $+0.18^{*}$ & $+0.63^{*}$ \\
OpenThai 2.0 Legal 30B-A3B & $+0.12^{*}$ & $+0.27^{*}$ && $+0.03$ & $-0.00$ \\
Nemotron 3 Super 120B-A12B & $+0.39^{*}$ & $+1.10^{*}$ && $+0.29^{*}$ & $+0.92^{*}$ \\
\hspace{0.9em}$\hookrightarrow$\,SEA-LION v4.8 120B-A12B & $+0.42^{*}$ & $+1.13^{*}$ && $+0.14^{*}$ & $+0.70^{*}$ \\
\midrule
\multicolumn{6}{l}{\textit{ThaiLLM 8B derivatives}} \\
Typhoon-S 8B & $+0.09^{*}$ & $+0.48^{*}$ && $-0.05$ & $-0.05$ \\
Pathumma 8B Think & $-0.01$ & $+0.45^{*}$ && $+0.01$ & $+0.22^{*}$ \\
THaLLE 0.2 8B & $+0.19^{*}$ & $+0.70^{*}$ && $+0.01$ & $-0.00$ \\
OpenThaiGPT 8B & $+0.09^{*}$ & $+0.48^{*}$ && $+0.00$ & $-0.01$ \\
\midrule
All models & $+0.25^{*}$ & $+0.79^{*}$ && $+0.11^{*}$ & $+0.47^{*}$ \\
\quad No-drift control & \multicolumn{2}{c}{$+0.01$} && \multicolumn{2}{c}{$-0.02$} \\
\bottomrule
\end{tabular}
\caption{Gain in Bar-standard score from the text in force over the current text, avg@3. A star marks a 95\% interval that excludes zero. The No-drift control is the same contrast on items whose text is identical in both prompts. Intervals in Table~\ref{tab:contrast-rubric-full}.}
\label{tab:contrast-rubric}
\end{table}

\begin{table}[!htbp]
\centering
\small
\setlength{\tabcolsep}{4pt}
\begin{tabular}{l c cc}
\toprule
\textbf{Amendment} & \textbf{Median $d$} & \textbf{$\Delta$ Bar-standard} & \textbf{$\Delta$ Citation F1} \\
\midrule
Rule changed (428 items) & 0.34 & +0.45 {\scriptsize[+0.38, +0.53]} & +0.04 {\scriptsize[+0.03, +0.06]} \\
Wording only (331 items) & 0.10 & -0.00 {\scriptsize[-0.03, +0.03]} & -0.00 {\scriptsize[-0.01, +0.00]} \\
\bottomrule
\end{tabular}
\caption{Cost of the current text by what the amendment changed. The judge labeled each changed gold section of the 759 Amended items as changing the rule the question turns on or only the wording. An item is rule-changed when any of its sections is. Median $d$ is the median size of the change. $\Delta$ is Gold-cited In-force minus Current, pooled over 13 judged models, with 95\% intervals from a bootstrap over item clusters.}
\label{tab:rule-change}
\end{table}

\subsection{The cost of the current text}
\label{subsec:temporal_cost}
Table~\ref{tab:contrast-rubric} compares the Bar-standard scores of historically accurate text (In-force) against modern text (Current).
The No-drift control group shifts by only 0.01.
In contrast, using the correct historical text improves scores by 0.25 on Amended items and 0.79 on Repealed items.
Every model improves on Repealed items, and every model but Pathumma on Amended items.
Stronger models see the largest gains, improving by up to 0.45 on Amended items and 1.35 on Repealed items.
Because the rubric evaluates verdicts and reasoning together, Table~\ref{tab:verdict-split} reports court agreement alongside scores for agreeing answers.
For drifted items, in-force text yields modest agreement gains (+1.6\,pp on amended, +4.5\,pp on repealed) but larger reasoning gains among agreeing answers (+0.29 and +0.95 points).
The right text occasionally corrects a verdict. Its main benefit is a stronger reasoning behind it.

\begin{table}[!ht]
\centering
\small
\setlength{\tabcolsep}{3pt}
\begin{tabular}{l cc cc}
\toprule
& \multicolumn{2}{c}{\textbf{Agrees with court (\%)}} & \multicolumn{2}{c}{\textbf{Score if agreeing}} \\
\cmidrule(lr){2-3} \cmidrule(lr){4-5}
\textbf{Band} & \textbf{Current} & \textbf{In force} & \textbf{Current} & \textbf{In force} \\
\midrule
No drift & 60.9 & 61.2 & 5.07 & 5.06 \\
Amended & 57.6 & 59.2 & 4.83 & 5.12 \\
Repealed & 53.9 & 58.4 & 3.88 & 4.83 \\
\bottomrule
\end{tabular}
\caption{Verdict and reasoning in the two Gold-cited settings, pooled over 13 judged models, avg@3. Agrees with court is the share of answers in tiers T4 to T8. Score if agreeing is the mean score among those answers.}
\label{tab:verdict-split}
\end{table}

\subsection{Recovery through historical retrieval}
\label{subsec:fix}
The RAG columns of Table~\ref{tab:contrast-rubric} test these effects inside a retrieval agent.
Searching on a historical index recovers a Bar-standard score of 0.11 on Amended items and 0.47 on Repealed items.
Recovery is limited by the search tool. The modern index finds the gold sections 40\% of the time, the historical index 48\% of the time. Repealed laws do not appear in a modern index.
Models that search often, such as Muse Glimmer, recover more. Weaker models that rarely search recover nothing.

\subsection{Changes of the rule and changes of the wording}
\label{subsec:analysis}
\begin{figure}[!htbp]\centering
\input{figures/amendment_examples.tex}
\resizebox{0.78\textwidth}{!}{\begin{tikzpicture}\amendmentExamples\end{tikzpicture}}
\caption{Two Amended items. Green is the text in force at the decision, brown the current text, bold the words that changed. The first amendment changed only the words. The second moved one cross-reference and changed the rule.}
\label{fig:amendment-examples}\end{figure}

The cost comes from changes of the rule, not of the wording.
The Amended band holds 759 items whose cited sections changed but still exist (\S\ref{sec:temporal-design}).
We manually labeled each of them by whether the amendment changed the rule the question turns on or only the phrasing. Figure~\ref{fig:amendment-examples} shows one of each.
For the 428 rule-changed items, the current text lowers the score by 0.45. For the 331 wording-only items, it costs 0.00, the same as the No-drift control (Table~\ref{tab:rule-change}).
The overall cost of 0.25 on Amended items is the average of the two groups.

\begin{figure}[!htbp]\centering
\input{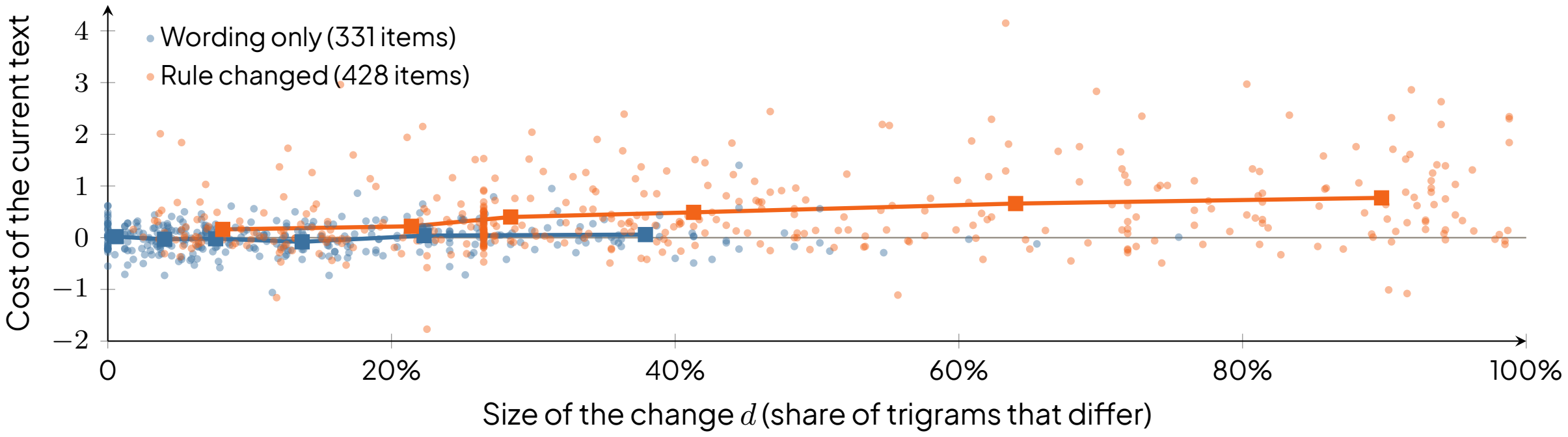}
\caption{Cost of the current text on Amended items against the size of the text change $d$. Each dot is one item. Squares are the mean cost in six bins of equal size.}
\label{fig:amendment-rule}\end{figure}

Figure~\ref{fig:amendment-rule} sets this label against the size of the text change, measured with character trigrams.
If $T(x)$ is the set of three-character sequences of a text $x$, the distance between the text in force $a$ and the current text $b$ is
\begin{equation}
d(a,b) = 1 - \frac{|T(a) \cap T(b)|}{|T(a) \cup T(b)|}.
\end{equation}
The distance is 0 when the texts are identical and 1 when they share no trigram.
Wording-only edits cost nothing at any size, $-$0.02 below the median distance ($d < 0.24$) and 0.05 above it.
Rule changes cost at every size, 0.21 points below the median and 0.54 above it.
The size of an edit does not predict its cost. Whether the amendment changes the rule does.
A second measure of the size, the normalized Levenshtein distance between the two texts, agrees. Table~\ref{tab:amendment} correlates both measures, averaged over the amended gold sections of an item, with the per-item change from Gold-cited Current to Gold-cited In-force. Larger amendments cost more, on the Bar-standard score and on Citation F1, and the correlations are small. The size of an amendment is one of several factors behind the per-item cost.

\begin{table}[!ht]
\centering
\small
\setlength{\tabcolsep}{3.5pt}
\begin{tabular}{l cc}
\toprule
\textbf{Amendment size} & \textbf{$\Delta$ Bar-standard} & \textbf{$\Delta$ Citation F1} \\
\midrule
1 $-$ character-trigram Jaccard & +0.35 {\scriptsize[+0.28, +0.41]} & +0.29 {\scriptsize[+0.22, +0.37]} \\
Normalized Levenshtein distance & +0.34 {\scriptsize[+0.27, +0.41]} & +0.28 {\scriptsize[+0.20, +0.35]} \\
\bottomrule
\end{tabular}
\caption{Spearman correlation between the size of the amendment and the per-item change from Gold-cited Current to Gold-cited In-force, on the 759 Amended items, pooled over the judged models. Brackets give 95\% intervals from a bootstrap over item clusters.}
\label{tab:amendment}
\end{table}

The models still cite the right sections under both texts.
On rule-changed items Citation F1 drops by 0.04 while the Bar-standard score drops by 0.45.

\section{Part II. The Leaderboard and the Limits of the Open-Weight Models}
\label{sec:platform}
\label{sec:capabilities}
\label{sec:results}

Part II measures today's models on the setting Part I established, the text in force. It follows the leaderboard protocol (\S\ref{sec:protocols}). Every model runs on the held-out test split of 500 items in three settings, and the public leaderboard page reports the same runs. The page is live and takes new models as we run them. The numbers in this report are the leaderboard as of 13 November 2026.
The page names the settings for a general reader, Closed book for Parametric, Open book for RAG In-force and Handout for Gold-cited In-force. This report keeps the names of \S\ref{sec:settings}. In short, Parametric is the model alone, RAG In-force is the model with a tool that searches the law in force, and Gold-cited In-force is the model with the court's sections handed to it.
Part II ranks the models (\S\ref{subsec:leaderboard}), sets the score against its cost (\S\ref{subsec:cost}), reads the score by domain (\S\ref{subsec:platform-domains}) and counts invented citations (\S\ref{subsec:halluc}). It ends with what limits the models (\S\ref{subsec:limits}) and sets a general intelligence index against the leaderboard (\S\ref{subsec:general-index}).

\subsection{The test split and the validation split}
\label{sec:split}

\begin{figure}[!htbp]\centering
\input{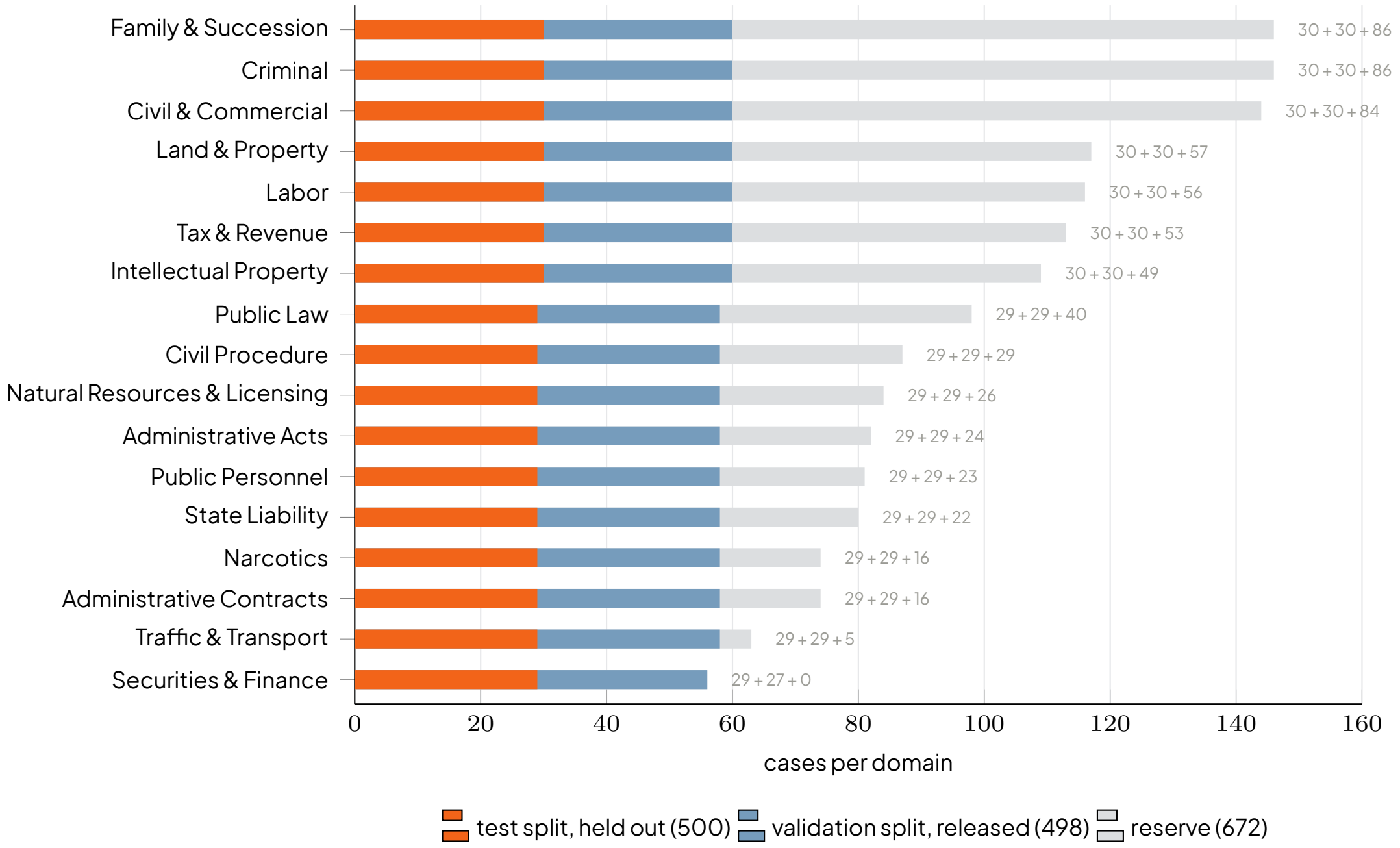}
\caption{The three parts of the benchmark. For each domain, the cases in the held-out test split (orange), the released validation split (blue) and the reserve (gray). The label gives the three counts.}
\label{fig:split}\end{figure}

A leaderboard has to take in new models as they are released, and a run over all 1{,}670 items at three samples is too costly to repeat for each one (\S\ref{sec:protocols}).
A leaderboard also has to stay honest once its questions are public. A model can be trained on a published test set, and its public score then stops meaning anything. GSM1k rebuilt a public arithmetic benchmark in private and found several model families up to 13 percent weaker on the private twin~\citep{zhang2024gsm1k}, and an audit of Chatbot Arena found providers testing dozens of private variants and publishing the best~\citep{singh2025leaderboard}.
We therefore cut the benchmark into three parts (Figure~\ref{fig:split}). The test split holds 500 items, 29 or 30 from each of the 17 domains, with the drift bands of each domain in the domain's own proportion and one item per shared-question cluster. The leaderboard runs on it, and it is never published. The validation split is drawn by the same recipe from the remaining items, 498 in all, with no item that shares a question with a test item, and it is released with the text of its gold sections (\S\ref{sec:release}). The 672 items left over are the reserve, which replaces the test split if it ever leaks.
The two splits are drawn to be alike, and the full-benchmark runs of Part I show that they are.

Table~\ref{tab:split-agreement} gives the 13 models of the benchmark protocol on the full benchmark, the test split and the validation split, from the same judged rows. A model's test score and validation score differ by 0.05 points on average and by 0.16 at most, and the two rank the models the same way, with a Spearman correlation of 0.97 to 0.99 across the three settings. The test split also tracks the full benchmark within 0.07 points for every model.
Scores on a split are macro-averaged over the 17 domains, as on the full benchmark.
The gap between a model's validation score and its test score is also the leaderboard's check against training on the released data. For the 13 models it stays within 0.16 points. A new model whose validation score runs far ahead of its test score has seen the validation split.

\begin{table}[!htb]
\centering
\small
\setlength{\tabcolsep}{3.2pt}
\begin{tabular}{l ccc ccc ccc}
\toprule
& \multicolumn{3}{c}{Parametric} & \multicolumn{3}{c}{Gold-cited In-force} & \multicolumn{3}{c}{RAG In-force} \\
\cmidrule(lr){2-4} \cmidrule(lr){5-7} \cmidrule(lr){8-10}
\textbf{Model} & Full & Test & Val. & Full & Test & Val. & Full & Test & Val. \\
\midrule
\multicolumn{10}{l}{\textit{General}} \\
Gemma 4 31B & 3.68 & 3.66 & 3.74 & 5.11 & 5.16 & 5.07 & 4.55 & 4.57 & 4.52 \\
Gemma 4 26B-A4B & 3.42 & 3.44 & 3.43 & 4.79 & 4.84 & 4.76 & 3.71 & 3.68 & 3.74 \\
Nemotron 3.5 Lightning 30B-A3B & 1.76 & 1.73 & 1.82 & 3.44 & 3.49 & 3.46 & 2.47 & 2.47 & 2.47 \\
Muse Glimmer 30B & 3.34 & 3.35 & 3.40 & 5.75 & 5.83 & 5.72 & 4.94 & 4.93 & 4.95 \\
\midrule
\multicolumn{10}{l}{\textit{Nemotron 3 bases and their adaptations}} \\
Nemotron 3 Nano 30B-A3B & 1.22 & 1.15 & 1.27 & 2.71 & 2.71 & 2.75 & 1.66 & 1.62 & 1.67 \\
\hspace{0.9em}$\hookrightarrow$\,SEA-LION v4.8 30B-A3B & 1.69 & 1.66 & 1.71 & 3.49 & 3.53 & 3.54 & 2.50 & 2.45 & 2.54 \\
OpenThai 2.0 Legal 30B-A3B & 1.16 & 1.18 & 1.18 & 1.95 & 1.92 & 1.95 & 0.88 & 0.87 & 0.89 \\
Nemotron 3 Super 120B-A12B & 2.08 & 2.05 & 2.15 & 4.27 & 4.29 & 4.37 & 3.31 & 3.23 & 3.36 \\
\hspace{0.9em}$\hookrightarrow$\,SEA-LION v4.8 120B-A12B & 2.05 & 2.04 & 2.07 & 4.26 & 4.29 & 4.28 & 2.94 & 2.88 & 3.04 \\
\midrule
\multicolumn{10}{l}{\textit{ThaiLLM 8B derivatives}} \\
Typhoon-S 8B & 1.73 & 1.74 & 1.77 & 2.79 & 2.77 & 2.82 & 1.56 & 1.54 & 1.56 \\
Pathumma 8B Think & 1.71 & 1.71 & 1.70 & 2.70 & 2.70 & 2.71 & 1.47 & 1.46 & 1.50 \\
THaLLE 0.2 8B & 1.21 & 1.18 & 1.25 & 2.61 & 2.62 & 2.64 & 1.11 & 1.10 & 1.11 \\
OpenThaiGPT 8B & 0.94 & 0.95 & 0.97 & 1.95 & 1.90 & 2.00 & 0.88 & 0.87 & 0.89 \\
\midrule
\multicolumn{10}{l}{\textbf{Agreement over the 13 models}} \\
Spearman, test against validation & \multicolumn{3}{c}{0.967} & \multicolumn{3}{c}{0.995} & \multicolumn{3}{c}{0.989} \\
Spearman, test against full & \multicolumn{3}{c}{0.973} & \multicolumn{3}{c}{1.000} & \multicolumn{3}{c}{0.995} \\
Mean absolute difference, test against validation & \multicolumn{3}{c}{0.05} & \multicolumn{3}{c}{0.05} & \multicolumn{3}{c}{0.05} \\
Largest absolute difference, test against validation & \multicolumn{3}{c}{0.12} & \multicolumn{3}{c}{0.11} & \multicolumn{3}{c}{0.16} \\
\bottomrule
\end{tabular}
\caption{The two splits are fair. Bar-standard score of the 13 models of the benchmark protocol on the full benchmark (1,670 items), the held-out test split (500) and the released validation split (498), three samples per item, the same judge rows. Val.\ is the validation split.}
\label{tab:split-agreement}
\end{table}

\subsection{The leaderboard}
\label{subsec:leaderboard}

\begin{figure}[!htb]\centering   %
\input{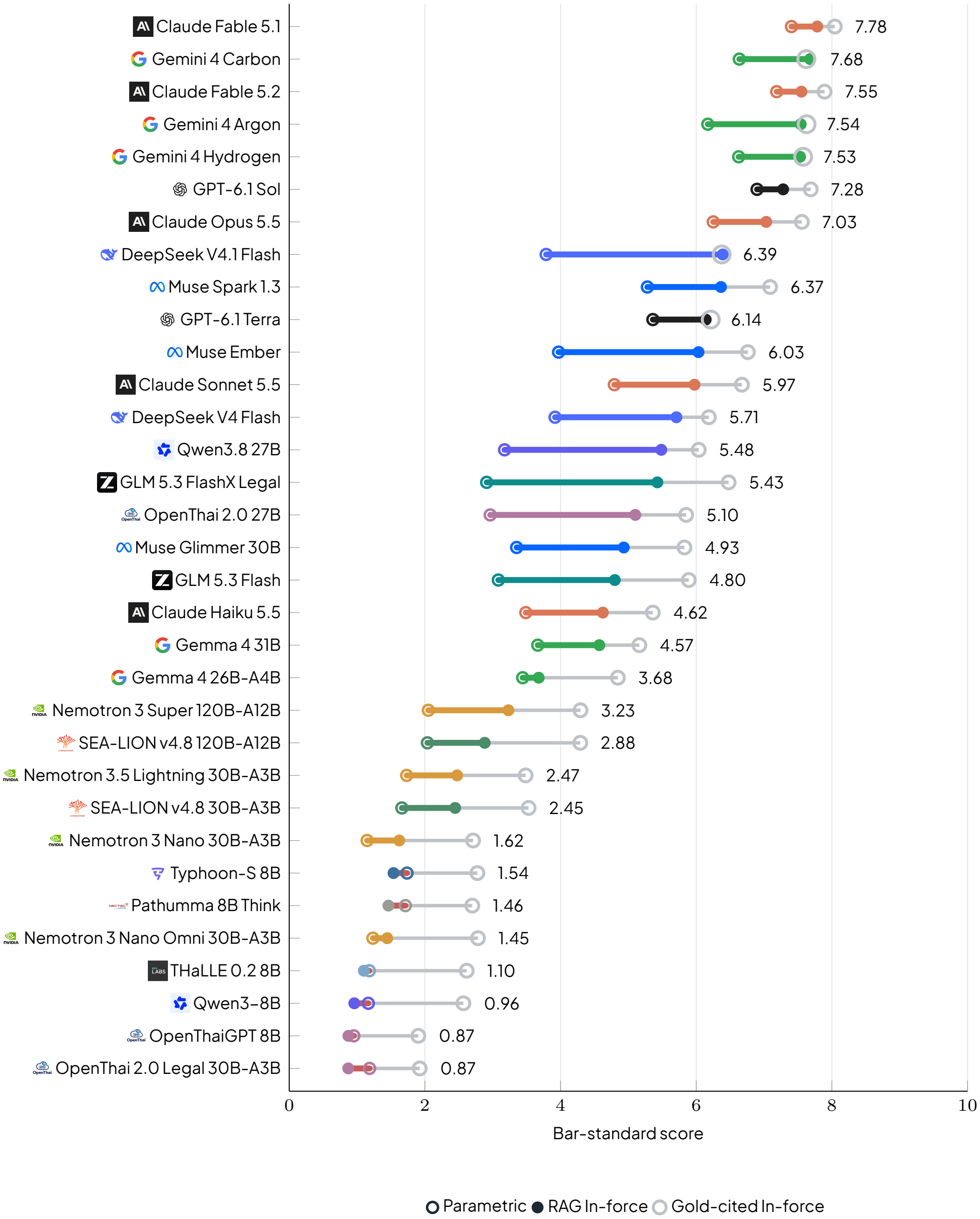}
\caption{The leaderboard as of 13 November 2026, 500 items. One line per row with the RAG In-force score printed. Hollow dot Parametric, solid dot RAG In-force, ring Gold-cited In-force. The colored segment is the gain from the search tool, red when the tool lowers the score.}
\label{fig:leaderboard}\end{figure}

Figure~\ref{fig:leaderboard} ranks the 33 models by their Bar-standard score in RAG In-force, the setting closest to a deployed assistant, where the model searches the law in force itself. Table~\ref{tab:leaderboard} in Appendix~\ref{app:leaderboard} gives every number.
Claude Fable 5.1 is the best model at 7.78. Gemini 4 Carbon follows with 7.68, Claude Fable 5.2 with 7.55, Gemini 4 Argon with 7.54 and Gemini 4 Hydrogen with 7.53. GPT-6.1 Sol scores 7.28 and Claude Opus 5.5 7.03.
The best open-weight model is DeepSeek V4.1 Flash at 6.39, and its predecessor DeepSeek V4 Flash scores 5.71. Qwen3.8 27B scores 5.48 and its Thai post-training OpenThai 2.0 27B 5.10. Muse Glimmer 30B, the best model of the benchmark protocol, is seventeenth with 4.93.
Per-domain scores of every row are in Figure~\ref{fig:domains} and Table~\ref{tab:leaderboard-domains}.

The Parametric column shows what a model knows with no text in front of it.
Claude Fable 5.1 scores 7.40 from memory alone, more than any open-weight model reaches with the search tool, Claude Fable 5.2 7.19 and GPT-6.1 Sol 6.90.
The proprietary models score 2.91 to 7.40 in this column, the open-weight models 0.95 to 3.92.

The Gold-cited In-force column is each row's ceiling, the score it reaches when the court's sections are handed to it.
The five best models reach 7.58 to 8.04 there, and their RAG In-force runs stay within 0.4 points of it. Finding the sections costs them little.
Lower in the table the distance grows, to 0.74 for Claude Haiku 5.5, 1.06 for Nemotron 3 Super and 1.41 for SEA-LION v4.8 120B.

The cited sections show two kinds of answer.
Every proprietary model has a recall above its precision, 0.69 to 0.88 against 0.36 to 0.61, and so do six open-weight models, among them the two DeepSeek models, 0.80 against 0.37 and 0.74 against 0.48. They cite more sections than the court did and catch most of the court's sections among them.
GPT-6.1 Sol has the best F1, 0.657.
Most open-weight models of the benchmark protocol do the reverse. They cite few sections and miss many of the court's. Gemma 4 31B has a precision of 0.642 and a recall of 0.460.

\subsection{What the score costs}
\label{subsec:cost}

\begin{figure}[!htbp]\centering
\input{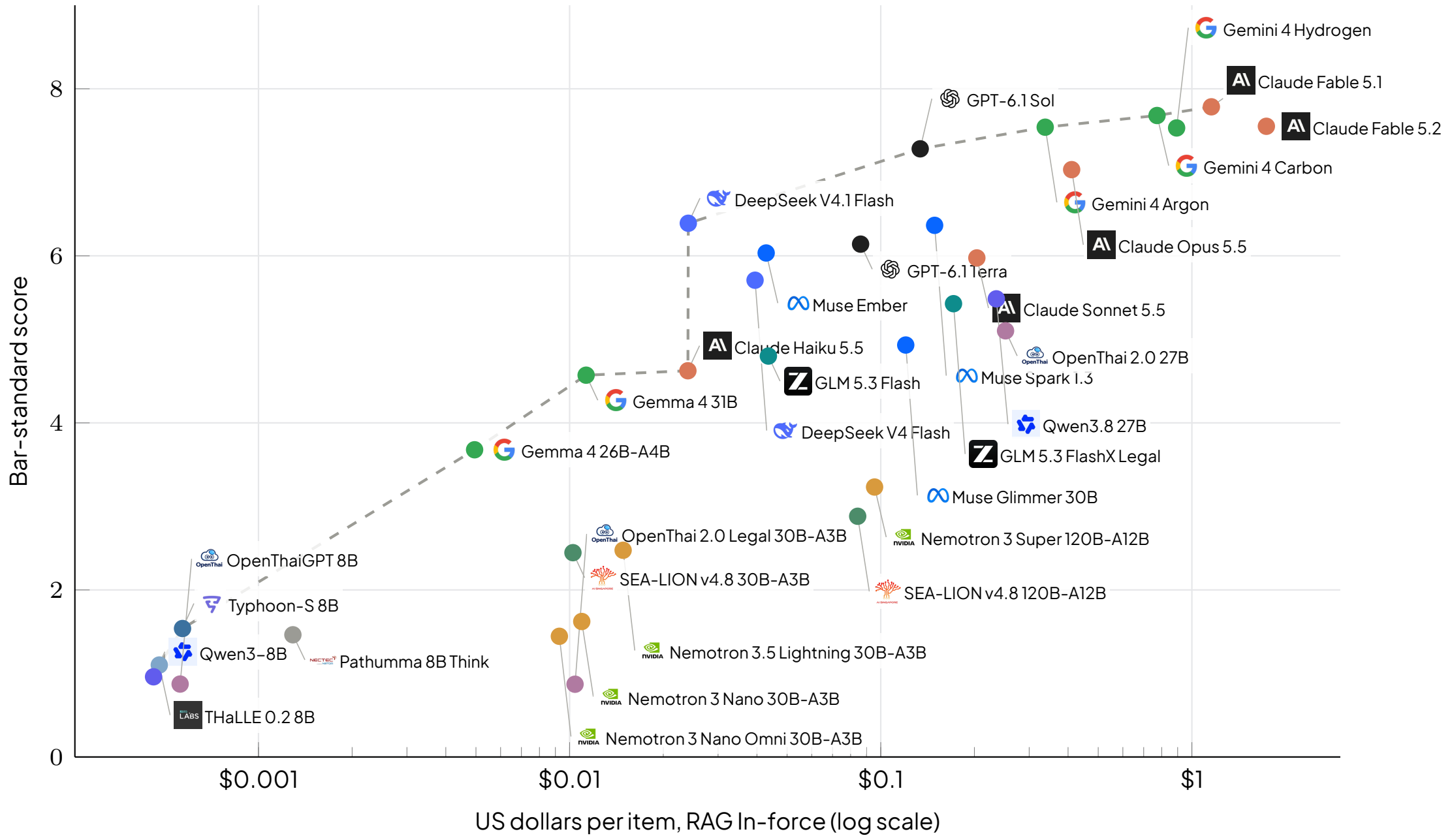}
\caption{Bar-standard score against US dollars per item in RAG In-force. The dashed line is the Pareto frontier.}
\label{fig:cost}\end{figure}

An item's cost is what the answer would have cost to buy. For a model behind an API it is the provider's list price on the day the model was added (26 October to 13 November 2026) applied to the measured input, cached and output tokens. For an open-weight model served locally it is the public API price of the same weights applied to the measured tokens, or the price of the base model when the model has no public price or no paid endpoint.
Figure~\ref{fig:cost} sets the score of each row against that cost in RAG In-force. Table~\ref{tab:leaderboard-cost} in Appendix~\ref{app:leaderboard} gives the cost, the output length, the searches and the hallucinated citations of each row.
An item costs from a tenth of a cent for the 8B models to 1.736 dollars for Claude Fable 5.2.
GPT-6.1 Sol at 0.134 dollars scores 7.28, at about one ninth of Claude Fable 5.1's price.
DeepSeek V4.1 Flash scores 6.39 at 0.024 dollars, the price of Claude Haiku 5.5, which scores 4.62.
Writing more does not buy score. DeepSeek V4.1 Flash writes 25.6 thousand tokens per item, DeepSeek V4 Flash 45.4 thousand and Qwen3.8 27B 52.3 thousand. GPT-6.1 Sol writes 4.1 thousand and scores above both.
Searching more does not buy score either. Among the models that search, the calls per item run from 1.2 for Gemma 4 26B-A4B to 45.5 for Gemini 4 Hydrogen, with no relation to rank.

\subsection{Performance by domain}
\label{subsec:platform-domains}

\begin{figure}[!htbp]\centering
\input{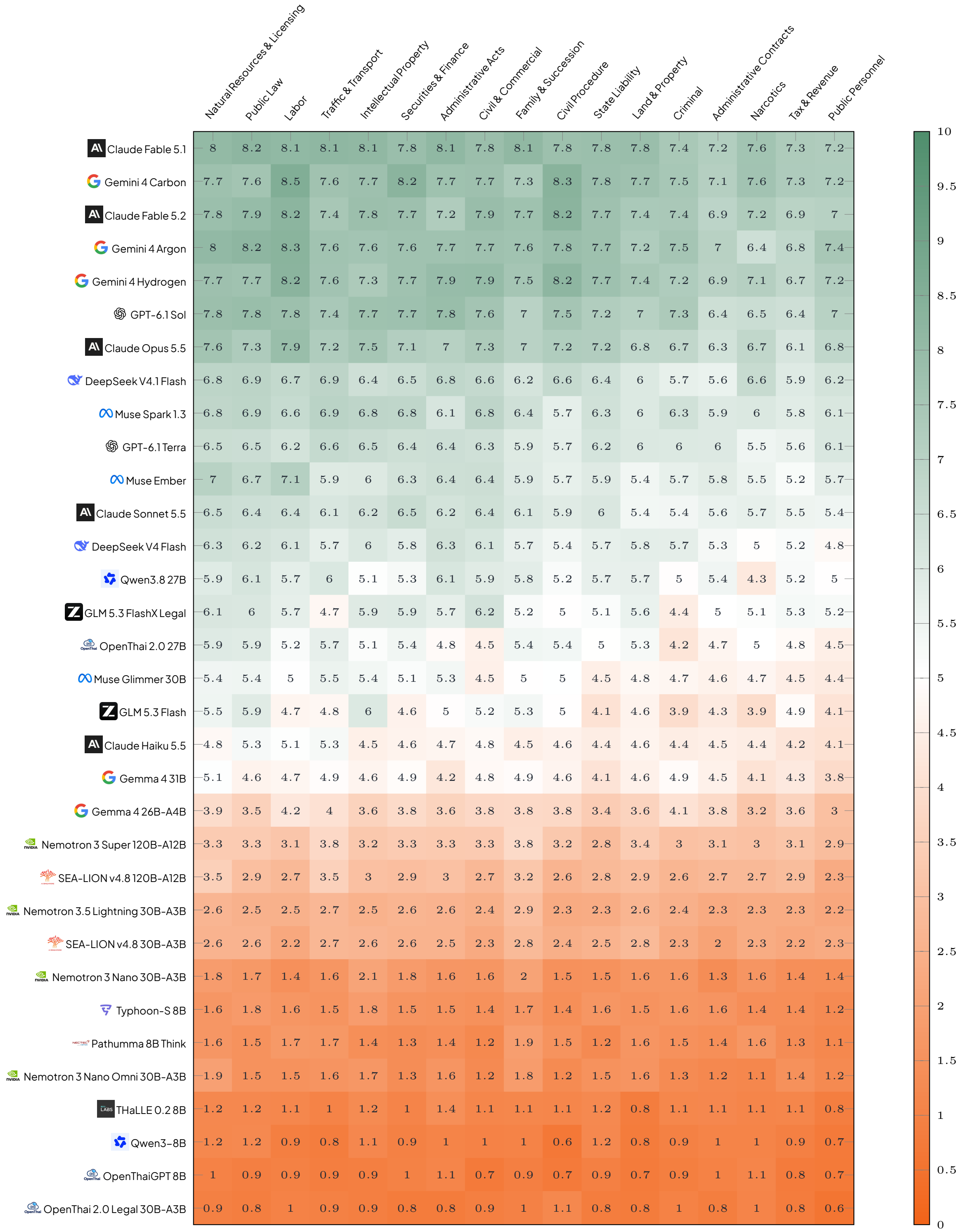}
\caption{Bar-standard score by leaderboard row and domain in RAG In-force. Columns run from the easiest domain to the hardest over the 33 models. Numbers in Table~\ref{tab:leaderboard-domains}.}
\label{fig:domains}\end{figure}

Figure~\ref{fig:domains} gives the RAG In-force score of every row in every domain.
Averaged over the 33 models, the easiest domain is Natural Resources \& Licensing at 4.86 and the hardest is Public Personnel at 4.11.
The three lowest domains are Narcotics, Tax \& Revenue and Public Personnel, and the three highest Natural Resources \& Licensing, Public Law and Labor.
The order of the rows holds from one domain to the next. Claude Fable 5.1 is the best model in 9 of the 17 domains, Gemini 4 Carbon in four, Gemini 4 Argon in three and Gemini 4 Hydrogen in one, and the scores of Claude Fable 5.1 run from 7.16 to 8.17.
A reader who needs one domain should read that column rather than the overall score.
The 13 open-weight models of the benchmark protocol give the same order on all 1{,}670 items at three samples per item, with Public Personnel again the lowest domain (Appendix~\ref{app:matrix}).
Narcotics is low for a reason of its own. In 69 of its 74 cases the court applied law that has since been repealed, and a model must answer from the text in force and not from what it learned about the current code.

\subsection{Hallucinated citations}
\label{subsec:halluc}

\begin{figure}[!htb]\centering   %
\input{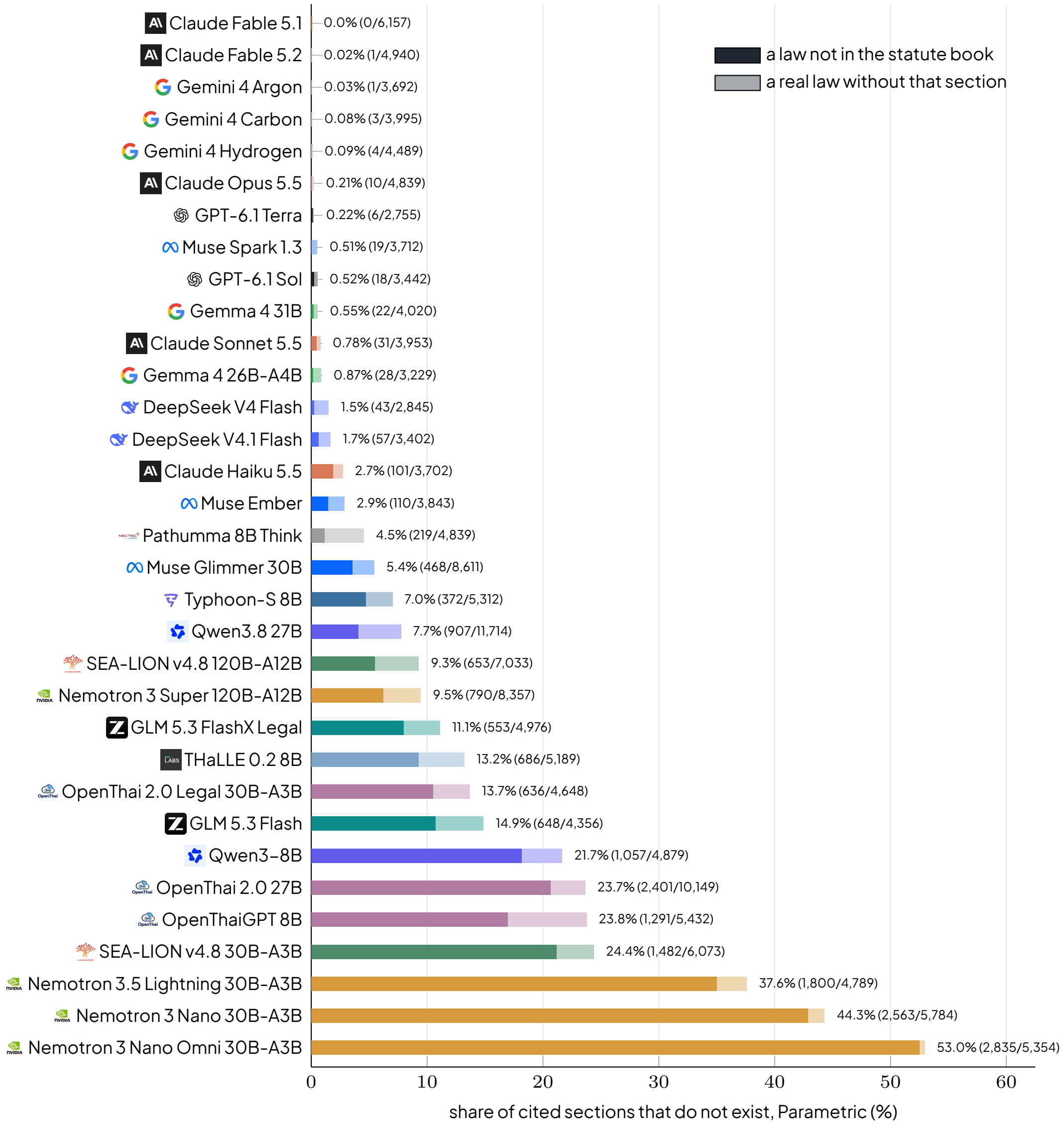}
\caption{Hallucinated citations in the Parametric setting. The share of cited sections that do not exist, split into laws that are not in the statute book and real laws cited with a section they do not have.}
\label{fig:halluc}\end{figure}

A model can cite a section that does not exist. We check every cited section against the statute corpus. A citation is hallucinated when the cited law is not in the corpus, or when the law exists and has no such section.
Citations the corpus cannot decide are left out of the count, and so are references without a section number. The corpus cannot decide an amending act or a ministerial regulation, a royal decree it does not hold, or a section of a repealed law whose text it does not hold. A real instrument the corpus lacks never counts as hallucinated.
The rate is the share of hallucinated citations among the checked citations of a setting.
Figure~\ref{fig:halluc} gives the share of such citations in the Parametric setting, where every citation comes from the model's memory.
The best models almost never do it. Claude Fable 5.1 invents none of the 6{,}157 sections it cites, Claude Fable 5.2 one of 4{,}940 and Gemini 4 Argon one of 3{,}692. Gemini 4 Carbon invents 3 of 3{,}995 and Gemini 4 Hydrogen 4 of 4{,}489. Claude Opus 5.5 and GPT-6.1 Terra invent 0.2 percent. GLM 5.3 FlashX Legal, a legal post-training, invents 11.1 percent, and its base GLM 5.3 Flash 14.9 percent.
At the other end, Nemotron 3 Nano Omni 30B-A3B invents 53.0 percent of the sections it cites, Nemotron 3 Nano 30B-A3B 44.3 percent, Nemotron 3.5 Lightning 30B-A3B 37.6 percent, SEA-LION v4.8 30B-A3B 24.4 percent and OpenThaiGPT 8B 23.8 percent.
Most invented citations are invented laws. Over the 33 models, 16{,}209 of the 19{,}815 hallucinated citations name a law that is not in the statute book, most of them from Nemotron 3 Nano Omni and Nemotron 3 Nano, and 3{,}606 name a real law with a section it does not have.
With the search tool (Table~\ref{tab:leaderboard-cost}) one thing decides the rate, whether the model searches.
The 13 proprietary models are at or near zero, and so are the two open-weight DeepSeek models.
The ThaiLLM 8B models invent 2.0 to 23.8 percent of the sections they cite, Qwen3-8B 19.2 percent and OpenThai 2.0 Legal 19.4 percent. These are the models that never or rarely call the tool (\S\ref{subsec:limits}).

\subsection{What limits the models}
\label{subsec:limits}

Three things can hold a model back on this benchmark, what it knows of the statutes, how it uses the search tool and how it applies the law to the facts. The three settings separate them. Parametric shows what the model knows from memory, RAG In-force what it does with the search tool, and Gold-cited In-force how it applies the court's sections once it has them. We take them in turn.

\begin{table}[!htb]
\centering
\small
\setlength{\tabcolsep}{5pt}
\begin{tabular}{l ccc ll r}
\toprule
& \multicolumn{3}{c}{\textbf{Bar-standard}} & & & \\
\cmidrule(lr){2-4}
\textbf{Model} & Param. & RAG InF. & GC InF. & \textbf{Gain}\up & \textbf{Gap}\down & \textbf{Searches} \\
\midrule
\multicolumn{7}{l}{\textit{Proprietary}} \\
Claude Fable 5.1 & \best{7.40} & \best{7.78} & \best{8.04} & \textcolor{fcGreen}{+0.38} & +0.26 & 8.6 \\
Gemini 4 Carbon & 6.64 & \second{7.68} & 7.62 & \textcolor{fcGreen}{+1.04} & \best{$-$0.06} & 19.1 \\
Claude Fable 5.2 & \second{7.19} & 7.55 & \second{7.89} & \textcolor{fcGreen}{+0.36} & +0.34 & 9.8 \\
Gemini 4 Argon & 6.17 & 7.54 & 7.63 & \textcolor{fcGreen}{+1.37} & +0.09 & 7.6 \\
Gemini 4 Hydrogen & 6.63 & 7.53 & 7.58 & \textcolor{fcGreen}{+0.90} & +0.05 & 45.5 \\
GPT-6.1 Sol & 6.90 & 7.28 & 7.69 & \textcolor{fcGreen}{+0.38} & +0.41 & 6.3 \\
Claude Opus 5.5 & 6.25 & 7.03 & 7.56 & \textcolor{fcGreen}{+0.78} & +0.53 & 5.3 \\
Muse Spark 1.3 & 5.28 & 6.37 & 7.09 & \textcolor{fcGreen}{+1.09} & +0.72 & 9.0 \\
GPT-6.1 Terra & 5.36 & 6.14 & 6.21 & \textcolor{fcGreen}{+0.78} & +0.07 & 9.2 \\
Muse Ember & 3.97 & 6.03 & 6.76 & \textcolor{fcGreen}{+2.06} & +0.73 & 26.8 \\
Claude Sonnet 5.5 & 4.79 & 5.97 & 6.67 & \textcolor{fcGreen}{+1.18} & +0.70 & 5.7 \\
GLM 5.3 FlashX Legal & 2.91 & 5.43 & 6.48 & \textcolor{fcGreen}{\second{+2.52}} & +1.05 & 16.1 \\
Claude Haiku 5.5 & 3.49 & 4.62 & 5.36 & \textcolor{fcGreen}{+1.14} & +0.74 & 7.4 \\
\midrule
\multicolumn{7}{l}{\textit{General}} \\
DeepSeek V4.1 Flash & 3.79 & 6.39 & 6.38 & \textcolor{fcGreen}{\best{+2.60}} & \second{$-$0.01} & 17.9 \\
DeepSeek V4 Flash & 3.92 & 5.71 & 6.19 & \textcolor{fcGreen}{+1.79} & +0.48 & 19.0 \\
Muse Glimmer 30B & 3.35 & 4.93 & 5.83 & \textcolor{fcGreen}{+1.58} & +0.89 & 15.8 \\
GLM 5.3 Flash & 3.08 & 4.80 & 5.89 & \textcolor{fcGreen}{+1.72} & +1.09 & 10.3 \\
Gemma 4 31B & 3.66 & 4.57 & 5.16 & \textcolor{fcGreen}{+0.91} & +0.59 & 3.4 \\
Gemma 4 26B-A4B & 3.44 & 3.68 & 4.84 & \textcolor{fcGreen}{+0.24} & +1.17 & 1.2 \\
Nemotron 3.5 Lightning 30B-A3B & 1.73 & 2.47 & 3.49 & \textcolor{fcGreen}{+0.74} & +1.01 & 5.3 \\
\midrule
\multicolumn{7}{l}{\textit{Qwen3.8 27B and its Thai adaptation}} \\
Qwen3.8 27B & 3.18 & 5.48 & 6.04 & \textcolor{fcGreen}{+2.31} & +0.55 & 17.6 \\
\hspace{0.9em}$\hookrightarrow$\,OpenThai 2.0 27B & 2.96 & 5.10 & 5.85 & \textcolor{fcGreen}{+2.14} & +0.75 & 16.6 \\
\midrule
\multicolumn{7}{l}{\textit{Nemotron 3 bases and their adaptations}} \\
Nemotron 3 Nano 30B-A3B & 1.15 & 1.62 & 2.71 & \textcolor{fcGreen}{+0.47} & +1.09 & 2.9 \\
\hspace{0.9em}$\hookrightarrow$\,SEA-LION v4.8 30B-A3B & 1.66 & 2.45 & 3.53 & \textcolor{fcGreen}{+0.78} & +1.08 & 4.7 \\
\hspace{0.9em}$\hookrightarrow$\,Nemotron 3 Nano Omni 30B-A3B & 1.23 & 1.45 & 2.78 & \textcolor{fcGreen}{+0.21} & +1.34 & 2.9 \\
\hspace{1.8em}$\hookrightarrow$\,OpenThai 2.0 Legal 30B-A3B & 1.18 & 0.87 & 1.92 & \textcolor{fcRed}{$-$0.31} & +1.05 & 0.0 \\
Nemotron 3 Super 120B-A12B & 2.05 & 3.23 & 4.29 & \textcolor{fcGreen}{+1.18} & +1.06 & 6.5 \\
\hspace{0.9em}$\hookrightarrow$\,SEA-LION v4.8 120B-A12B & 2.04 & 2.88 & 4.29 & \textcolor{fcGreen}{+0.84} & +1.41 & 6.2 \\
\midrule
\multicolumn{7}{l}{\textit{Qwen3-8B-Base descendants}} \\
Qwen3-8B & 1.17 & 0.96 & 2.57 & \textcolor{fcRed}{$-$0.21} & +1.61 & 0.0 \\
\hspace{0.9em}$\hookrightarrow$\,Typhoon-S 8B & 1.74 & 1.54 & 2.77 & \textcolor{fcRed}{$-$0.20} & +1.24 & 0.4 \\
\hspace{0.9em}$\hookrightarrow$\,Pathumma 8B Think & 1.71 & 1.46 & 2.70 & \textcolor{fcRed}{$-$0.25} & +1.24 & 2.8 \\
\hspace{0.9em}$\hookrightarrow$\,THaLLE 0.2 8B & 1.18 & 1.10 & 2.62 & \textcolor{fcRed}{$-$0.08} & +1.52 & 0.0 \\
\hspace{0.9em}$\hookrightarrow$\,OpenThaiGPT 8B & 0.95 & 0.87 & 1.90 & \textcolor{fcRed}{$-$0.08} & +1.03 & 0.0 \\
\bottomrule
\end{tabular}
\caption{What the search tool adds, leaderboard protocol, 500 items. Bar-standard score in Parametric (Param.), RAG In-force (RAG InF.) and Gold-cited In-force (GC InF.). Gain is RAG In-force minus Parametric. Gap is Gold-cited In-force minus RAG In-force, the part a perfect retriever would add. Searches are retrieval calls per item in RAG In-force. Indented rows are post-trainings of the base above them. OpenThai 2.0 Legal is indented under Nemotron 3 Nano Omni, its base. In the last group the four ThaiLLM 8B models are not post-trainings of Qwen3-8B. Both descend from Qwen3-8B-Base, the ThaiLLM models through ThaiLLM-8B, a Thai continued pretraining that cannot be served, and Qwen3-8B through Alibaba's own post-training. Best per column in bold, second best underlined; for Gap the smallest is best.}
\label{tab:tool-use}
\end{table}

\paragraph{Knowledge of the statutes.}
The Parametric setting asks how much Thai law a model knows from training alone. The model gets the question and nothing else, and every section it cites comes from memory.
The proprietary models of the large providers know a good deal. They score 3.49 to 7.40 and cite the court's sections with an F1 of 0.26 to 0.60 (Table~\ref{tab:leaderboard}). GLM 5.3 FlashX Legal is the exception among them, 2.91 and an F1 of 0.12, the level of its open-weight base.
The open-weight models know little. They score 0.95 to 3.92, and only five of them recall some Thai law, the two DeepSeek models with an F1 of 0.33 and 0.32, Gemma 4 31B 0.31, Gemma 4 26B-A4B 0.27 and Muse Glimmer 0.21.
No other passes 0.16. Pathumma and Qwen3.8 27B come closest, 0.15 and 0.14. The ThaiLLM 8B models cite the court's sections for 2 to 15 percent of sub-questions, and the Thai post-trainings know little more law than their reference models (Table~\ref{tab:adapt}).
An open-weight model that answers from memory is guessing at the law for most questions.

\paragraph{Use of the search tool.}
The RAG In-force setting gives the model a tool that searches the law in force.
Table~\ref{tab:tool-use} shows two numbers for each of the 33 models. The gain is what the tool adds, the RAG In-force score minus the Parametric score. The gap is how far the model still stays below its Gold-cited In-force score, where the court's sections are given, and it is the most a better search could still add.
For most models the tool helps. Twenty-seven gain, by 0.21 to 2.60 points. Six lose, by 0.08 to 0.31.
Among the open-weight models, those that search most gain most. DeepSeek V4 Flash makes 19.0 calls per item and gains 1.79 points, DeepSeek V4.1 Flash 17.9 calls and 2.60 points, Qwen3.8 27B 17.6 and 2.31, OpenThai 2.0 27B 16.6 and 2.14, and Muse Glimmer 15.8 and 1.58.
Eleven of the 13 proprietary models gain less, 0.36 to 1.37 points with 5.3 to 45.5 calls per item, because they knew more before they searched. The two that knew least from memory gain most, Muse Ember 2.06 points with 26.8 calls (3.97 from memory) and GLM 5.3 FlashX Legal 2.52 with 16.1 calls (2.91).
The six that lose are the four ThaiLLM 8B models, Qwen3-8B and OpenThai 2.0 Legal, and they lose because they do not use the tool. Qwen3-8B, THaLLE, OpenThaiGPT and OpenThai 2.0 Legal never call it. Typhoon-S calls it on one item in three. Pathumma calls it 2.8 times per item and still loses 0.25 points. It retrieves text it cannot use.
For these six models the tool only fills the context. It brings them no law.
The gap is small at the top. For the proprietary models it is at most 1.05 points, and Gemini 4 Carbon scores 0.06 above its Gold-cited In-force score. DeepSeek V4.1 Flash has closed it, 6.39 as an agent against 6.38 with the sections provided. Every other open-weight model stays 0.48 to 1.61 points below its Gold-cited In-force score.
Even the best agents find the right sections for only part of the questions. With the tool the proprietary models cite the court's sections with an F1 of 0.46 to 0.66. With the sections provided they reach 0.86 to 0.92.

\paragraph{Application of the law.}
The Gold-cited In-force setting gives the model the court's sections in the prompt. Nothing is left to find, and the score measures one thing, how well the model applies the law it was given to the facts of the case.
With the sections provided the proprietary models score 5.36 to 8.04 and the open-weight models 1.90 to 6.38. Three open-weight models pass 6, DeepSeek V4.1 Flash at 6.38, DeepSeek V4 Flash at 6.19 and Qwen3.8 27B at 6.04.
The share of answers that reach the court's verdict shows the same distance. The proprietary models reach it in 72 to 92 percent of their answers, the open-weight models in 39 to 79 percent.
Retrieval does not set this distance. The sections were in the prompt. What remains between the model and a good answer is the reasoning from the statute to the case, and no open-weight model closes it.

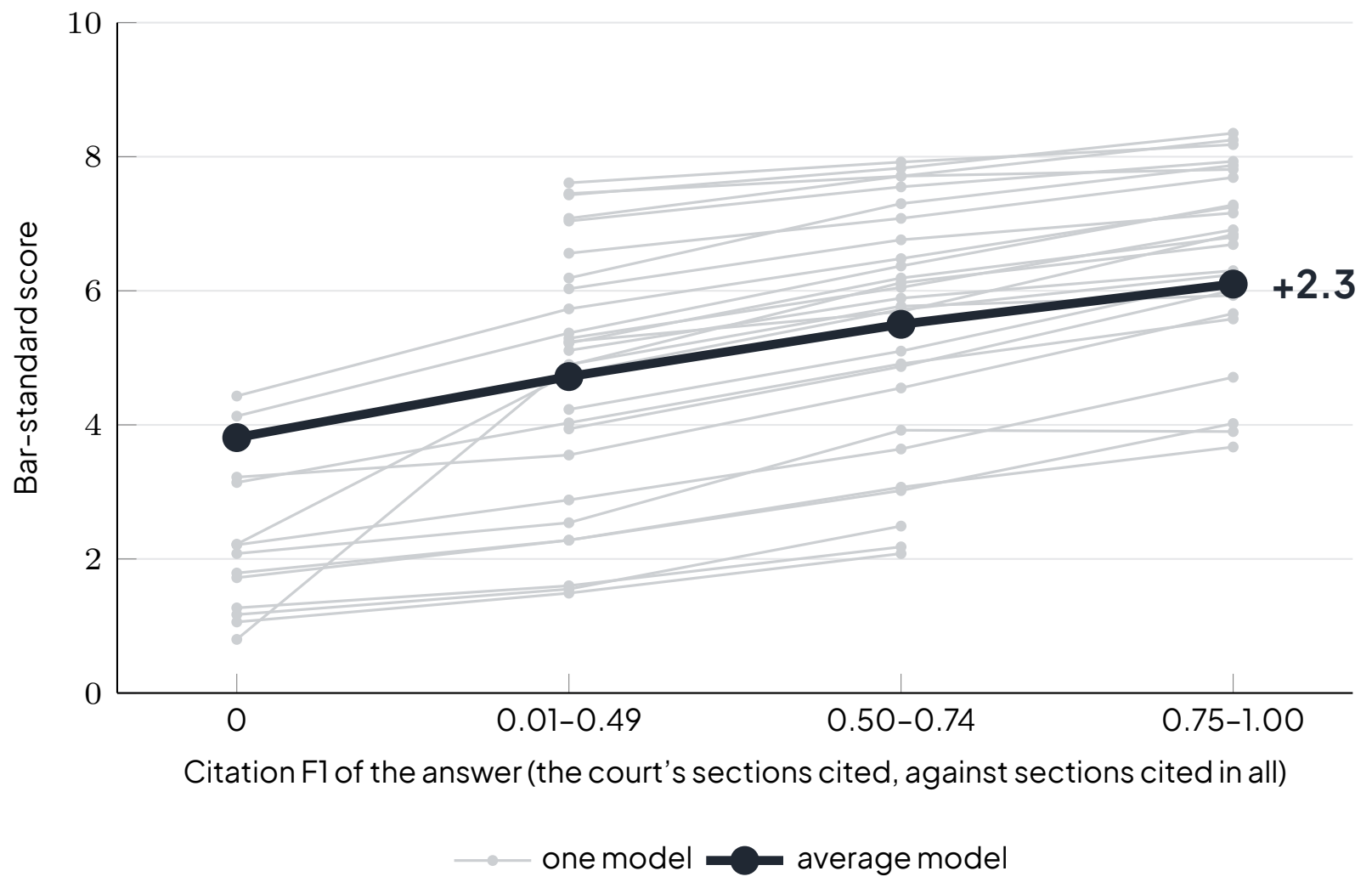
\begin{figure}[!htb]\centering   %
\begin{tikzpicture}
\begin{axis}[width=0.7\textwidth, height=6.8cm, ymin=0, ymax=10, ytick={0,2,...,10}, symbolic x coords={none,under half,half or more,all}, xtick={none,under half,half or more,all}, xticklabels={{0},{0.01--0.49},{0.50--0.74},{0.75--1.00}}, enlarge x limits=0.12,
  x tick label style={font=\scriptsize}, xlabel={Citation F1 of the answer (the court's sections cited, against sections cited in all)}, xlabel style={font=\scriptsize},
  ylabel={Bar-standard score}, ylabel style={font=\scriptsize}, tick label style={font=\scriptsize}, ymajorgrids, grid style={ciStone!40}, axis lines*=left, clip=false,
  legend style={at={(0.5,-0.2)}, anchor=north, legend columns=2, font=\scriptsize, draw=none, fill=none}, legend cell align=left]
\addplot[ciStone!80, semithick, mark=*, mark size=0.9pt, mark options={fill=ciStone!80}] coordinates {(none,3.14) (under half,4.03) (half or more,4.91) (all,5.58)};
\addplot[ciStone!80, semithick, mark=*, mark size=0.9pt, mark options={fill=ciStone!80}, forget plot] coordinates {(none,3.22) (under half,3.55) (half or more,4.55) (all,5.66)};
\addplot[ciStone!80, semithick, mark=*, mark size=0.9pt, mark options={fill=ciStone!80}, forget plot] coordinates {(none,1.79) (under half,2.28) (half or more,3.07) (all,3.67)};
\addplot[ciStone!80, semithick, mark=*, mark size=0.9pt, mark options={fill=ciStone!80}, forget plot] coordinates {(under half,4.23) (half or more,5.10) (all,6.17)};
\addplot[ciStone!80, semithick, mark=*, mark size=0.9pt, mark options={fill=ciStone!80}, forget plot] coordinates {(none,1.27) (under half,1.60) (half or more,2.18)};
\addplot[ciStone!80, semithick, mark=*, mark size=0.9pt, mark options={fill=ciStone!80}, forget plot] coordinates {(none,1.72) (under half,2.28) (half or more,3.02) (all,4.02)};
\addplot[ciStone!80, semithick, mark=*, mark size=0.9pt, mark options={fill=ciStone!80}, forget plot] coordinates {(none,2.21) (under half,2.88) (half or more,3.64) (all,4.71)};
\addplot[ciStone!80, semithick, mark=*, mark size=0.9pt, mark options={fill=ciStone!80}, forget plot] coordinates {(none,2.08) (under half,2.54) (half or more,3.92) (all,3.90)};
\addplot[ciStone!80, semithick, mark=*, mark size=0.9pt, mark options={fill=ciStone!80}, forget plot] coordinates {(none,1.17) (under half,1.55) (half or more,2.49)};
\addplot[ciStone!80, semithick, mark=*, mark size=0.9pt, mark options={fill=ciStone!80}, forget plot] coordinates {(under half,5.24) (half or more,5.69) (all,6.84)};
\addplot[ciStone!80, semithick, mark=*, mark size=0.9pt, mark options={fill=ciStone!80}, forget plot] coordinates {(none,0.80) (under half,4.90) (half or more,5.77) (all,5.93)};
\addplot[ciStone!80, semithick, mark=*, mark size=0.9pt, mark options={fill=ciStone!80}, forget plot] coordinates {(none,1.06) (under half,1.49) (half or more,2.08)};
\addplot[ciStone!80, semithick, mark=*, mark size=0.9pt, mark options={fill=ciStone!80}, forget plot] coordinates {(under half,3.94) (half or more,4.87) (all,6.01)};
\addplot[ciStone!80, semithick, mark=*, mark size=0.9pt, mark options={fill=ciStone!80}, forget plot] coordinates {(under half,5.29) (half or more,6.05) (all,6.91)};
\addplot[ciStone!80, semithick, mark=*, mark size=0.9pt, mark options={fill=ciStone!80}, forget plot] coordinates {(under half,6.56) (half or more,7.08) (all,7.69)};
\addplot[ciStone!80, semithick, mark=*, mark size=0.9pt, mark options={fill=ciStone!80}, forget plot] coordinates {(under half,7.61) (half or more,7.92) (all,8.18)};
\addplot[ciStone!80, semithick, mark=*, mark size=0.9pt, mark options={fill=ciStone!80}, forget plot] coordinates {(under half,7.04) (half or more,7.55) (all,7.93)};
\addplot[ciStone!80, semithick, mark=*, mark size=0.9pt, mark options={fill=ciStone!80}, forget plot] coordinates {(under half,6.19) (half or more,7.30) (all,7.87)};
\addplot[ciStone!80, semithick, mark=*, mark size=0.9pt, mark options={fill=ciStone!80}, forget plot] coordinates {(under half,6.03) (half or more,6.76) (all,7.16)};
\addplot[ciStone!80, semithick, mark=*, mark size=0.9pt, mark options={fill=ciStone!80}, forget plot] coordinates {(under half,4.89) (half or more,6.12) (all,6.69)};
\addplot[ciStone!80, semithick, mark=*, mark size=0.9pt, mark options={fill=ciStone!80}, forget plot] coordinates {(none,2.22) (under half,4.76) (half or more,5.72) (all,6.24)};
\addplot[ciStone!80, semithick, mark=*, mark size=0.9pt, mark options={fill=ciStone!80}, forget plot] coordinates {(under half,5.11) (half or more,5.89) (all,6.30)};
\addplot[ciStone!80, semithick, mark=*, mark size=0.9pt, mark options={fill=ciStone!80}, forget plot] coordinates {(none,4.43) (under half,5.73) (half or more,6.48) (all,7.25)};
\addplot[ciStone!80, semithick, mark=*, mark size=0.9pt, mark options={fill=ciStone!80}, forget plot] coordinates {(under half,5.22) (half or more,6.19) (all,6.80)};
\addplot[ciStone!80, semithick, mark=*, mark size=0.9pt, mark options={fill=ciStone!80}, forget plot] coordinates {(under half,7.08) (half or more,7.71) (all,7.81)};
\addplot[ciStone!80, semithick, mark=*, mark size=0.9pt, mark options={fill=ciStone!80}, forget plot] coordinates {(none,4.13) (under half,5.37) (half or more,6.37) (all,7.28)};
\addplot[ciStone!80, semithick, mark=*, mark size=0.9pt, mark options={fill=ciStone!80}, forget plot] coordinates {(under half,7.43) (half or more,7.83) (all,8.35)};
\addplot[ciStone!80, semithick, mark=*, mark size=0.9pt, mark options={fill=ciStone!80}, forget plot] coordinates {(under half,7.45) (half or more,7.71) (all,8.25)};
\addplot[fcInk, line width=2pt, mark=*, mark size=2.2pt] coordinates {(none,3.81) (under half,4.72) (half or more,5.50) (all,6.10)};
\node[anchor=west, font=\small\bfseries, text=fcInk, xshift=5pt] at (axis cs: all, 6.10) {+2.3};
\legend{one model, average model}
\end{axis}
\end{tikzpicture}
\caption{Bar-standard score by the Citation F1 of the answer, RAG In-force. Gray, one line per model that cites. Black, the average model, with each model centered on its own mean. The number is the rise of the black line.}
\label{fig:cite}\end{figure}

\paragraph{Does citing the right sections increase the Bar-standard score?}
\label{par:cite}
Figure~\ref{fig:cite} sorts each model's answers by their Citation F1 and plots the mean score of each group. Every line rises. On the average model, answers with an F1 of 0.75 or more score 2.3 points above answers that cited none of the court's sections.
F1 is the one citation number to track. Within a model it follows the score as closely as recall does, a median correlation of 0.34 against 0.33, and more closely than precision, 0.29. Of its two halves, recall carries more weight, 0.29 against 0.10 in a regression with both.
Citing related law beyond the court's sections costs little. Answers that cite all of the court's sections and more than as many again still score 0.45 above the model's mean, and 78 percent of their extra sections are in the same act as one of the court's.
The gray lines sit about six points apart. The model matters more than its citations.
Citations explain 13 percent of the differences between answers. The rest is the reasoning the rubric grades. With the court's sections handed to the model, in Gold-cited In-force, the share falls to 5 percent.
The judge sees the citations and the rubric rewards a right one, and part of the link is built in. Even so, the score is not citation matching in disguise. Better citations are worth about two points of ten. The reasoning is worth the rest.

\begin{table}[!htbp]
\centering
\footnotesize
\setlength{\tabcolsep}{3pt}
\begin{tabular}{l lll ll}
\toprule
& \multicolumn{3}{c}{\textbf{Bar-standard}} & \multicolumn{2}{c}{\textbf{In RAG In-force}} \\
\cmidrule(lr){2-4} \cmidrule(lr){5-6}
\textbf{Model} & Param. & GC InF. & RAG InF. & \textbf{Citation F1} & \textbf{Searches} \\
\midrule
Nemotron 3 Nano 30B-A3B & \makebox[2.8em][l]{1.15}\makebox[3.9em][l]{} & \makebox[2.8em][l]{2.71}\makebox[3.9em][l]{} & \makebox[2.8em][l]{1.62}\makebox[3.9em][l]{} & \makebox[3.6em][l]{0.251}\makebox[4.6em][l]{} & \makebox[2.8em][l]{2.9}\makebox[3.9em][l]{} \\
\hspace{0.9em}$\hookrightarrow$\,SEA-LION v4.8 30B-A3B & \makebox[2.8em][l]{1.66}\makebox[3.9em][l]{{\scriptsize\textcolor{fcGreen}{(+0.51)}}} & \makebox[2.8em][l]{3.53}\makebox[3.9em][l]{{\scriptsize\textcolor{fcGreen}{(+0.82)}}} & \makebox[2.8em][l]{2.45}\makebox[3.9em][l]{{\scriptsize\textcolor{fcGreen}{(+0.83)}}} & \makebox[3.6em][l]{0.335}\makebox[4.6em][l]{{\scriptsize\textcolor{fcGreen}{(+0.084)}}} & \makebox[2.8em][l]{4.7}\makebox[3.9em][l]{{\scriptsize\textcolor{fcGreen}{(+1.8)}}} \\
\hspace{0.9em}$\hookrightarrow$\,Nemotron 3 Nano Omni 30B-A3B & \makebox[2.8em][l]{1.23}\makebox[3.9em][l]{{\scriptsize\textcolor{fcGreen}{(+0.08)}}} & \makebox[2.8em][l]{2.78}\makebox[3.9em][l]{{\scriptsize\textcolor{fcGreen}{(+0.07)}}} & \makebox[2.8em][l]{1.45}\makebox[3.9em][l]{{\scriptsize\textcolor{fcRed}{($-$0.17)}}} & \makebox[3.6em][l]{0.194}\makebox[4.6em][l]{{\scriptsize\textcolor{fcRed}{($-$0.057)}}} & \makebox[2.8em][l]{2.9}\makebox[3.9em][l]{{\scriptsize\textcolor{fcMuted}{(+0.0)}}} \\
\hspace{1.8em}$\hookrightarrow$\,OpenThai 2.0 Legal 30B-A3B & \makebox[2.8em][l]{1.18}\makebox[3.9em][l]{{\scriptsize\textcolor{fcRed}{($-$0.05)}}} & \makebox[2.8em][l]{1.92}\makebox[3.9em][l]{{\scriptsize\textcolor{fcRed}{($-$0.86)}}} & \makebox[2.8em][l]{0.87}\makebox[3.9em][l]{{\scriptsize\textcolor{fcRed}{($-$0.58)}}} & \makebox[3.6em][l]{0.005}\makebox[4.6em][l]{{\scriptsize\textcolor{fcRed}{($-$0.189)}}} & \makebox[2.8em][l]{0.0}\makebox[3.9em][l]{{\scriptsize\textcolor{fcRed}{($-$2.9)}}} \\
Nemotron 3 Super 120B-A12B & \makebox[2.8em][l]{2.05}\makebox[3.9em][l]{} & \makebox[2.8em][l]{4.29}\makebox[3.9em][l]{} & \makebox[2.8em][l]{3.23}\makebox[3.9em][l]{} & \makebox[3.6em][l]{0.424}\makebox[4.6em][l]{} & \makebox[2.8em][l]{6.5}\makebox[3.9em][l]{} \\
\hspace{0.9em}$\hookrightarrow$\,SEA-LION v4.8 120B-A12B & \makebox[2.8em][l]{2.04}\makebox[3.9em][l]{{\scriptsize\textcolor{fcRed}{($-$0.01)}}} & \makebox[2.8em][l]{4.29}\makebox[3.9em][l]{{\scriptsize\textcolor{fcMuted}{(+0.00)}}} & \makebox[2.8em][l]{2.88}\makebox[3.9em][l]{{\scriptsize\textcolor{fcRed}{($-$0.35)}}} & \makebox[3.6em][l]{0.351}\makebox[4.6em][l]{{\scriptsize\textcolor{fcRed}{($-$0.073)}}} & \makebox[2.8em][l]{6.2}\makebox[3.9em][l]{{\scriptsize\textcolor{fcRed}{($-$0.3)}}} \\
Qwen3.8 27B & \makebox[2.8em][l]{\best{3.18}}\makebox[3.9em][l]{} & \makebox[2.8em][l]{\second{6.04}}\makebox[3.9em][l]{} & \makebox[2.8em][l]{\best{5.48}}\makebox[3.9em][l]{} & \makebox[3.6em][l]{\second{0.480}}\makebox[4.6em][l]{} & \makebox[2.8em][l]{17.6}\makebox[3.9em][l]{} \\
\hspace{0.9em}$\hookrightarrow$\,OpenThai 2.0 27B & \makebox[2.8em][l]{2.96}\makebox[3.9em][l]{{\scriptsize\textcolor{fcRed}{($-$0.22)}}} & \makebox[2.8em][l]{5.85}\makebox[3.9em][l]{{\scriptsize\textcolor{fcRed}{($-$0.19)}}} & \makebox[2.8em][l]{5.10}\makebox[3.9em][l]{{\scriptsize\textcolor{fcRed}{($-$0.38)}}} & \makebox[3.6em][l]{\best{0.509}}\makebox[4.6em][l]{{\scriptsize\textcolor{fcGreen}{(+0.029)}}} & \makebox[2.8em][l]{16.6}\makebox[3.9em][l]{{\scriptsize\textcolor{fcRed}{($-$1.0)}}} \\
Qwen3-8B & \makebox[2.8em][l]{1.17}\makebox[3.9em][l]{} & \makebox[2.8em][l]{2.57}\makebox[3.9em][l]{} & \makebox[2.8em][l]{0.96}\makebox[3.9em][l]{} & \makebox[3.6em][l]{0.000}\makebox[4.6em][l]{} & \makebox[2.8em][l]{0.0}\makebox[3.9em][l]{} \\
\hspace{0.9em}$\hookrightarrow$\,Typhoon-S 8B & \makebox[2.8em][l]{1.74}\makebox[3.9em][l]{{\scriptsize\textcolor{fcGreen}{(+0.57)}}} & \makebox[2.8em][l]{2.77}\makebox[3.9em][l]{{\scriptsize\textcolor{fcGreen}{(+0.20)}}} & \makebox[2.8em][l]{1.54}\makebox[3.9em][l]{{\scriptsize\textcolor{fcGreen}{(+0.58)}}} & \makebox[3.6em][l]{0.034}\makebox[4.6em][l]{{\scriptsize\textcolor{fcGreen}{(+0.034)}}} & \makebox[2.8em][l]{0.4}\makebox[3.9em][l]{{\scriptsize\textcolor{fcGreen}{(+0.4)}}} \\
\hspace{0.9em}$\hookrightarrow$\,Pathumma 8B Think & \makebox[2.8em][l]{1.71}\makebox[3.9em][l]{{\scriptsize\textcolor{fcGreen}{(+0.54)}}} & \makebox[2.8em][l]{2.70}\makebox[3.9em][l]{{\scriptsize\textcolor{fcGreen}{(+0.13)}}} & \makebox[2.8em][l]{1.46}\makebox[3.9em][l]{{\scriptsize\textcolor{fcGreen}{(+0.50)}}} & \makebox[3.6em][l]{0.155}\makebox[4.6em][l]{{\scriptsize\textcolor{fcGreen}{(+0.155)}}} & \makebox[2.8em][l]{2.8}\makebox[3.9em][l]{{\scriptsize\textcolor{fcGreen}{(+2.8)}}} \\
\hspace{0.9em}$\hookrightarrow$\,THaLLE 0.2 8B & \makebox[2.8em][l]{1.18}\makebox[3.9em][l]{{\scriptsize\textcolor{fcGreen}{(+0.01)}}} & \makebox[2.8em][l]{2.62}\makebox[3.9em][l]{{\scriptsize\textcolor{fcGreen}{(+0.05)}}} & \makebox[2.8em][l]{1.10}\makebox[3.9em][l]{{\scriptsize\textcolor{fcGreen}{(+0.14)}}} & \makebox[3.6em][l]{0.003}\makebox[4.6em][l]{{\scriptsize\textcolor{fcGreen}{(+0.003)}}} & \makebox[2.8em][l]{0.0}\makebox[3.9em][l]{{\scriptsize\textcolor{fcMuted}{(+0.0)}}} \\
\hspace{0.9em}$\hookrightarrow$\,OpenThaiGPT 8B & \makebox[2.8em][l]{0.95}\makebox[3.9em][l]{{\scriptsize\textcolor{fcRed}{($-$0.22)}}} & \makebox[2.8em][l]{1.90}\makebox[3.9em][l]{{\scriptsize\textcolor{fcRed}{($-$0.67)}}} & \makebox[2.8em][l]{0.87}\makebox[3.9em][l]{{\scriptsize\textcolor{fcRed}{($-$0.09)}}} & \makebox[3.6em][l]{0.000}\makebox[4.6em][l]{{\scriptsize\textcolor{fcMuted}{(+0.000)}}} & \makebox[2.8em][l]{0.0}\makebox[3.9em][l]{{\scriptsize\textcolor{fcMuted}{(+0.0)}}} \\
GLM 5.3 Flash & \makebox[2.8em][l]{\second{3.08}}\makebox[3.9em][l]{} & \makebox[2.8em][l]{5.89}\makebox[3.9em][l]{} & \makebox[2.8em][l]{4.80}\makebox[3.9em][l]{} & \makebox[3.6em][l]{0.414}\makebox[4.6em][l]{} & \makebox[2.8em][l]{10.3}\makebox[3.9em][l]{} \\
\hspace{0.9em}$\hookrightarrow$\,GLM 5.3 FlashX Legal & \makebox[2.8em][l]{2.91}\makebox[3.9em][l]{{\scriptsize\textcolor{fcRed}{($-$0.17)}}} & \makebox[2.8em][l]{\best{6.48}}\makebox[3.9em][l]{{\scriptsize\textcolor{fcGreen}{(+0.59)}}} & \makebox[2.8em][l]{\second{5.43}}\makebox[3.9em][l]{{\scriptsize\textcolor{fcGreen}{(+0.63)}}} & \makebox[3.6em][l]{0.459}\makebox[4.6em][l]{{\scriptsize\textcolor{fcGreen}{(+0.045)}}} & \makebox[2.8em][l]{16.1}\makebox[3.9em][l]{{\scriptsize\textcolor{fcGreen}{(+5.8)}}} \\
\bottomrule
\end{tabular}
\caption{Bases and their post-trainings on the leaderboard split. Bar-standard score in Parametric (Param.), Gold-cited In-force (GC InF.) and RAG In-force (RAG InF.), and Citation F1 and searches per item in RAG In-force. Indented rows are post-trainings of the base above them, with the change against the base in brackets. OpenThai 2.0 Legal is indented under Nemotron 3 Nano Omni, its base. In the last group the reference row is Qwen3-8B, a sibling of the ThaiLLM 8B models rather than their base. Both descend from Qwen3-8B-Base, the ThaiLLM models through ThaiLLM-8B, a Thai continued pretraining that cannot be served, and Qwen3-8B through Alibaba's own post-training. Best per column in bold, second best underlined.}
\label{tab:adapt}
\end{table}

\paragraph{Thai and legal post-training.}
\label{subsec:adapt}
Several groups take an open-weight model and train it further on Thai text or on legal text. The aim is a model that handles Thai law.
Nine such models are on the split, each with a reference model (Table~\ref{tab:adapt}). For five of them the reference is the base they were trained from. The four ThaiLLM 8B models are post-trainings of ThaiLLM-8B, a Thai continued pretraining of Qwen3-8B-Base that has no chat template and cannot answer the benchmark. Their reference is Qwen3-8B, Alibaba's own post-training of the same Qwen3-8B-Base, a sibling rather than a parent. That comparison measures the Thai route, continued pretraining on Thai text and then post-training, against Alibaba's post-training of the same base. OpenThai 2.0 Legal's base is Nemotron 3 Nano Omni 30B-A3B, NVIDIA's multimodal model on the Nemotron 3 Nano backbone, which scores within 0.08 of Nano in every setting. Table~\ref{tab:adapt} shows Omni under Nano and OpenThai 2.0 Legal under Omni.

The extra training taught the models little law.
In the Parametric setting, where a model answers from memory, each post-training scores close to its base.
SEA-LION v4.8 30B-A3B gains the most, 0.51 points over Nemotron 3 Nano, and still cites almost none of the court's sections from memory, a Citation F1 of 0.04 in that setting (Table~\ref{tab:leaderboard}).
OpenThai 2.0 Legal, trained on Thai legal text, scores within 0.05 of its base. GLM 5.3 FlashX Legal, Z.ai's legal post-training of GLM 5.3 Flash, scores 0.17 below its base. OpenThai 2.0 27B scores 0.22 below Qwen3.8 27B. At 8B the Thai route is worth about half a point for two of the four models. Typhoon-S and Pathumma score 0.57 and 0.54 above Qwen3-8B, THaLLE is level with it and OpenThaiGPT is 0.22 below.

What the extra training changed in the larger pairs is how the model behaves with the search tool, and in three of the five it changed it for the worse.
SEA-LION v4.8 30B-A3B searches more than its base, 4.7 calls per item against 2.9. It gains 0.83 points in RAG In-force and its Citation F1 rises from 0.25 to 0.34.
SEA-LION v4.8 120B-A12B searches as often as Nemotron 3 Super and scores 0.35 below it. OpenThai 2.0 27B scores 0.38 below Qwen3.8 27B.
OpenThai 2.0 Legal stopped calling the tool. Its base makes 2.9 calls per item and it makes none, and it loses 0.58 points.
GLM 5.3 FlashX Legal is the one post-training that gains with the tool. It searches more than its base, 16.1 calls per item against 10.3, scores 0.63 above it in RAG In-force and 0.59 above it with the court's sections provided, and its Citation F1 rises from 0.41 to 0.46. From memory it scores 0.17 below its base. The legal training taught it to search and to apply the law, not the law itself.
Searching the law in force is the one thing that lifts an open-weight model on this benchmark (Table~\ref{tab:tool-use}). Losing it costs more than the training added.

At 8B the base is the limit.
Qwen3-8B and the four ThaiLLM 8B models land within 0.9 points of each other in every setting. None of the five calls the search tool, and the half point that Typhoon-S and Pathumma gain in Parametric is the same half point they gain in RAG In-force.

Two things raise an open-weight model on this benchmark, a stronger base and a search tool the model uses.
Across the nine comparisons, training on Thai or legal text added tool use once, in GLM 5.3 FlashX Legal, and removed it once. It added statute knowledge in two of the four ThaiLLM 8B models, about half a point over Qwen3-8B, and in none of the larger pairs.
Every model ran with the defaults shipped in its repository and the same prompt (\S\ref{sec:models}). A post-training tuned for a different prompt style may do better than it does here.

\subsection{A general index does not order the Thai legal frontier}
\label{subsec:general-index}

\begin{figure}[!htb]\centering
\input{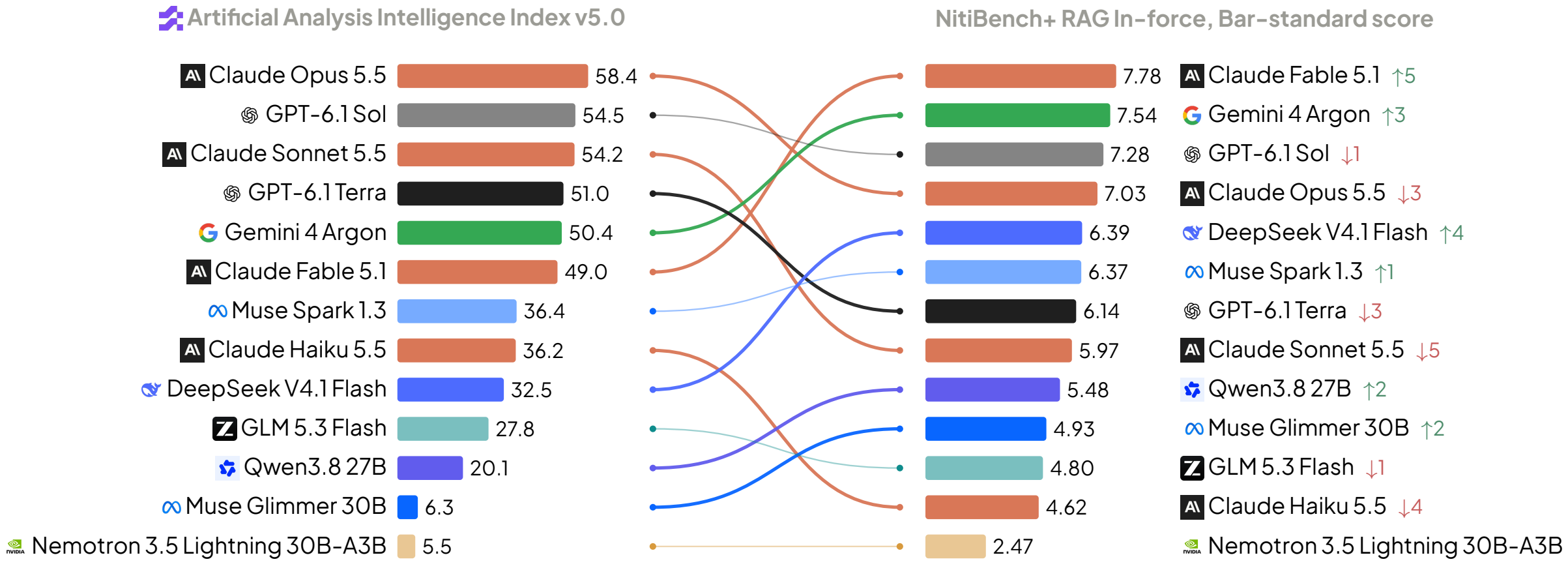}
\caption{The 13 leaderboard models that have an Artificial Analysis Intelligence Index v5.0, ranked by the index (left) and by the Bar-standard score in RAG In-force (right). One line per model. The arrows give the places gained or lost.}
\label{fig:general-index}\end{figure}

Model releases are judged by general indices built from coding, agentic and knowledge tests. We asked whether such an index predicts the standing on Thai legal work.
Thirteen of the 33 leaderboard models have an Artificial Analysis Intelligence Index v5.0~\citep{artificialanalysis2026index}, read through its API on 3 November 2026 at the headline configuration of each model (the four models added after 3 November are not in it). Figure~\ref{fig:general-index} ranks the 13 by the index and by their RAG In-force score.
Over the 13 models the two rankings agree on the tiers, a Spearman correlation of 0.67. Among the eight best models by the index they do not, 0.19. Claude Fable 5.1 is sixth by the index and first on \benchmark{}. Claude Sonnet 5.5 is third and eighth. DeepSeek V4.1 Flash is ninth and fifth.
Of the tests the index reports, Humanity's Last Exam (HLE) orders the eight closest to \benchmark{}, 0.69, and Terminal-Bench 4.0 farthest, 0.14.
Two caveats apply. The index scores the Claude models with extended thinking and the leaderboard runs them without it. Eight models cannot establish a correlation. The finding is the narrow one. A general index tells the tier of a model. The choice among the frontier models for Thai legal work needs an in-domain measurement.

\section{Discussion}
\label{sec:discussion}

\paragraph{What the setup must hold.}
A benchmark built from dated court decisions and scored against the current law measures two things at once, how well a model reasons and how far the law has moved since the decision.
The second part is large. Half of the 156{,}650 decisions we audited cite law that has since changed, and in the Supreme Court two in three do (\S\ref{sec:population-drift}).
It is also easy to miss. The current text rarely flips the verdict. It weakens the reasoning behind a verdict that was right either way (\S\ref{subsec:temporal_cost}). On the items where the Bar-standard score drops by 0.45, Citation F1 moves by 0.04 (\S\ref{subsec:analysis}).
A benchmark that checks the section number or the outcome alone does not see this loss. Its score mixes legal reasoning with statutory drift, in a share that changes every time the legislature sits.
A fair setup records the date of each decision, names the version of each gold citation, stores the text in force next to the current text, and releases the full versioned corpus. Without the corpus nobody can repeat the experiments or audit the gold answers. Without the dates a benchmark goes stale and nobody notices, as NitiBench shows two years after release (\S\ref{sec:population-drift}).
Every result of Part II is measured on this setup. Of the 53 legal benchmarks we audited, none meets it (\S\ref{sec:population-drift}).

\paragraph{For system builders.}
Searching the current law is the wrong default for a dispute rooted in a past event.
Missing repealed law costs three times more than reading amended law, 0.79 against 0.25 points. Repealed provisions must stay in the index.
The date filter belongs in the retrieval layer rather than in the prompt. A search over the law in force recovers 0.11 points on amended law and 0.47 on repealed law for the models that search, and nothing for the models that do not (\S\ref{subsec:fix}).
The strongest models gain the most from the text in force, up to 0.45 points on amended law and 1.35 on repealed law, because they build their answer on the text they are given. As models improve, the penalty for the wrong version grows.

\paragraph{Where the models stand.}
The best proprietary models score 7.0 to 7.8 on the leaderboard with the search tool, and Claude Fable 5.1 scores 7.4 from memory alone. On the Bar Council scale this is the good tier.
These models know the Thai statutes, they search, and they stay within half a point of their own ceiling.
They cite more sections than the court did, a precision of 0.39 to 0.61 against a recall of 0.69 to 0.88 (\S\ref{subsec:leaderboard}).
The best open-weight model, DeepSeek V4.1 Flash, scores 6.39, 1.4 points below Claude Fable 5.1, at 0.02 dollars per answer (\S\ref{subsec:cost}).
The 13 open-weight models of the benchmark protocol stop at 4.94 with the search tool and 5.75 with the court's sections in the prompt, the acceptable tier.
Two things stand between them and a good answer. The first is finding the law. The in-force index returns the court's sections for 48 percent of the searches (\S\ref{subsec:fix}). The second is applying it. With the right sections in hand no model passes 6, and about six answers in ten reach the court's verdict (Table~\ref{tab:verdict-split}).

\paragraph{The tool is the dividing line.}
The 33 models split into those that use the search tool and those that do not.
The 22 that use it gain 0.24 to 2.60 points. The six that do not lose points.
Four of the six never call the tool. Two call it and cannot use what it returns.
A Thai legal system built on a model that does not search can only know what the model remembers. The Parametric column of Table~\ref{tab:tool-use} shows how little that is for every open-weight model but the two DeepSeek models, Gemma 4 and Muse Glimmer.
Tool use comes first for Thai legal QA, before Thai data and before legal data.

\paragraph{A general index does not pick the model.}
Launch posts quote general indices. On the 13 leaderboard models that have one, the Artificial Analysis Intelligence Index sorts the tiers and not the frontier (\S\ref{subsec:general-index}).
A buyer who picks a model for Thai legal work from a general index gets the tier right and the order within it no better than chance. The in-domain measurement is the one to use.

\paragraph{What post-training on Thai data adds.}
Training an open-weight model further on Thai or legal text is the main route Thai groups take to a Thai legal model.
On \benchmark{} it added little law, about half a point in two of the four ThaiLLM 8B models over Qwen3-8B, which shares their base, and nothing in the larger pairs. In RAG In-force it raised one of the larger bases and lowered three, and in one pair the model stopped using the tool (\S\ref{subsec:adapt}).
Qwen3-8B and the four ThaiLLM 8B models land within 0.9 points of each other, and none of the five searches.
The one legal post-training that gains, GLM 5.3 FlashX Legal, gains in the two grounded settings and not from memory (\S\ref{subsec:adapt}).
The gains on this benchmark come from a stronger base and from searching the law in force.
The open problem is a post-training that keeps the tool use of the base and teaches the model to apply the statutes to the facts. The benchmark measures both.

\paragraph{Domains.}
Over the 27 leaderboard rows the mean score runs from 4.49 in Natural Resources \& Licensing to 3.76 in Public Personnel. The five lowest domains are Public Personnel, Administrative Contracts, Tax \& Revenue, Narcotics and Criminal (\S\ref{subsec:platform-domains}).
Two of the five are domains of the Supreme Administrative Court. Its cases cite acts, decrees and regulations outside the codes, and each of these has its own history of amendment and repeal.
A model trained on the codes alone is not a model of Thai law.
The per-domain scores are the numbers a user should read, and Appendix~\ref{app:matrix} gives them for every model.

\section{Conclusion}
\label{sec:conclusion}

\benchmark{} measures Thai legal question answering the way a Thai court and a Thai bar examiner would. The questions come from decided cases, the grade covers the verdict and the reasoning, and the sections the court relied on are read in the version in force on the day of the decision.

Part I shows that the version of the text is part of the task.
Half of 156{,}650 Thai decisions cite law that has since changed.
When a model reads the current text of a changed law, its Bar-standard score falls. This holds for every model, and verdict agreement and Citation F1 barely move.
A legal benchmark must therefore record the date of each question, name the version of the law in force on that date, and grade against that text. A deployed system must index the law by date and keep repealed provisions.
The text in force is the fair setting, and it is the setting of the leaderboard.

Part II measures today's models on that setting.
The best proprietary model is graded good.
The best open-weight model of the benchmark protocol is graded acceptable, and no open-weight model reaches the good tier.
With no text in front of them, the open-weight models know almost no Thai law. The two DeepSeek Flash models, Gemma 4 and Muse Glimmer are the exceptions.
A search tool over the law in force closes part of the distance for the models that use it, and none for the models that do not.
The court's sections in the prompt close more. What remains is the application of the law to the facts.
The administrative domains are the hardest. Training on Thai or legal data added about half a point of law knowledge in two of the four ThaiLLM 8B models, against Alibaba's post-training of the same base, and none in the three larger pairs we can test.
The leaderboard takes new models as they are released, on the same setup. Its test split is held out, and a validation split of the same shape is public.

\section*{Limitations}

\paragraph{Scope.}
\benchmark{} covers Thai law and follows Thai legal practice. To repeat the method in another jurisdiction, one needs the statute text as it stood on each date and a corpus of dated decisions.
The benchmark samples from three high courts. It leaves out military courts, lower courts, and advisory rulings from government agencies.
The 17 domains hold between 56 and 146 items each. The macro-average weights them equally.

\paragraph{Judges.}
An LLM judge scores the reasoning.
Its agreement with human bar examination graders was checked in earlier work~\citep{akarajaradwong2026barexam}. A judge model cannot fully replace a human expert, least of all on a fine or borderline point of law.

\paragraph{Models.}
The benchmark protocol holds 13 open-weight models of 8B to 120B parameters, served locally.
The leaderboard adds 13 proprietary models and seven further open-weight models on the 500-item split, with one sample per item for every row outside the 13, Qwen3-8B and Nemotron 3 Nano Omni. A single sample carries the sampling noise shown in Table~\ref{tab:sample-sd}.
Every model runs with the defaults shipped in its repository and the same prompt. A prompt or an agent loop tuned to one model can move its score.
The test split is held out, so a model appears on the leaderboard when we run it and not when its maker does.

\paragraph{Magnitudes.}
The version effects are modest.
Reading the current text of amended law costs about 0.25 points of the Bar-standard score on its 10-point scale, and missing repealed law costs about 0.8 points.
Seeing any legal text at all matters more. It adds 0.8 to 2.4 points across models.

\section*{Ethics Statement}

All rulings and decisions used are publicly available government documents.
Under Section 7(2) to (5) of Thailand's Copyright Act B.E.\ 2537 (1994), government documents are explicitly excluded from copyright protection and reside in the public domain.
\benchmark{} is intended for research evaluation and is not a substitute for professional legal advice.

\bibliography{references}

\appendix
\section{Dataset}
\label{app:dataset}

\subsection{Released files and dataset example}
\label{sec:examples}
The validation split is released as two JSON Lines files (\S\ref{sec:release}). Every record of both files carries the canary string in a \texttt{canary} field. In the text fields of the validation split the names of private individuals and of officials named in person are replaced by their initials in the style of Thai law reports, with the same initial for the same person within a case, and citizen ID numbers, personal phone numbers, bank account numbers, vehicle plates and house numbers are masked. The example that follows is a case of the validation split, the item of Figure~\ref{fig:hero-case}.

\paragraph{validation\_split.jsonl, one record per case.}
\texttt{id}, \texttt{domain} (one of the 17 domains), \texttt{court} and \texttt{source} (the deciding court, as a name and as a key), \texttt{source\_id} (the decision number), \texttt{decision\_date} (ISO-8601), \texttt{year\_be}, \texttt{decade}, \texttt{title}, \texttt{fact\_pattern} (the facts of the case, Thai), \texttt{drift\_band}, \texttt{n\_sub\_questions}, \texttt{n\_cited\_laws}, \texttt{asks\_shared\_question} (the cases that ask the same question, used to cluster confidence intervals and to keep the two splits apart), \texttt{corpus\_date} (the date of the current text) and \texttt{sub\_questions}.

\paragraph{Sub-question.}
{\sloppy \texttt{question}, \texttt{question\_type} (\texttt{ruling} for a question the court itself answered, or \texttt{governing\_provisions}, \texttt{legal\_consequences}, \texttt{provision\_interaction} and \texttt{lower\_court\_departure} for the four added kinds of \S\ref{sec:case-pipeline}), \texttt{depends\_on\_previous} (\textsc{true} when this sub-question's verdict is conditioned on the prior one), \texttt{verdict} and \texttt{reasoning} (the gold answer, copied from the decision), and \texttt{citations}.\par}

\paragraph{Citation.}
\texttt{law\_code} and \texttt{law\_name\_th} (the law), \texttt{label} (the section as cited), \texttt{is\_loadbearing} (\textsc{true} when the court relies on the section for the verdict), \texttt{version\_uri} with \texttt{valid\_from} and \texttt{valid\_to} (the version in force on the decision date), \texttt{current\_uri} (the version in force on the corpus date) and \texttt{drift} (\texttt{unchanged}, \texttt{amended} or \texttt{repealed}). Only citations the court relies on are scored. Background mentions are kept with \texttt{is\_loadbearing} set to \textsc{false}. A citation of an instrument the statute corpus does not hold, a notification or an order for example, carries the instrument's name in \texttt{law} and no identifier, and is left out of Citation F1.

\paragraph{gold\_sections.jsonl, one record per section version.}
\texttt{version\_uri} (of the form \texttt{th/law/\{law\_code\}/}\allowbreak\texttt{section-\{label\}/v/\{valid\_from\}}), \texttt{law\_code}, \texttt{law\_name\_th}, \texttt{label}, \texttt{component\_type}, \texttt{valid\_from}, \texttt{valid\_to}, \texttt{is\_law\_active} and \texttt{text}. The file holds both versions of every gold section of the split, 3{,}253 section versions, so that the Gold-cited In-force setting can be run from the release by joining on \texttt{version\_uri}.

\paragraph{The rest of the release.}
The rubric, the judge prompt and the citation-matching rules as documents, the composition of the split, a license file and checksums.

\begin{quote}\small
\textbf{Item} \texttt{labor::deka::18883-2557}. \emph{Domain}: labor. \emph{Source}: Supreme Court Decision 18883/2557. \emph{Decision date}: 2014. \emph{Drift band}: Amended.

\textbf{Fact pattern} (Thai, translated): A voluntarily insured person under the Social Security Act B.E.~2533 died of causes unrelated to work. She had no spouse and no children, her parents had died before her, and she had named no beneficiary. Her full sister claimed the death grant and the old-age lump sum. The Social Security Office refused both, and the appeal committee upheld the refusals. The sister sued, arguing that the contributions were savings that passed to her as a statutory heir. The Central Labour Court upheld the refusal of the death grant. It held that the old-age lump sum was part of the estate and passed to the statutory heirs. Both parties appealed.

\textbf{Sub-question 2} (contrast with the lower court): The Central Labour Court held that the right of a deceased insured person to the old-age lump sum passes to the statutory heirs under the Civil and Commercial Code, and that the full sister may therefore claim it. Was that correct, and why?

\textbf{Gold verdict}: No. The sister is not among the heirs that section 77 quater, paragraph two, entitles to the lump sum. Even when none of those heirs exists, money of the Social Security Fund cannot be paid to other heirs against sections 24 and 77 quater.

\textbf{Gold citations}: Social Security Act B.E.~2533, sections 77 quater, 77(2), and 24 paragraph one, all relied on. Section 77 bis is kept as background.

\textbf{Section 77 quater}. \emph{In force on the decision date}: \texttt{version\_uri} \texttt{.../v/2002-10-09}, valid 2002-10-09 to 2015-10-19. \emph{Current}: \texttt{current\_uri} \texttt{.../v/2023-10-25}, \texttt{drift} \texttt{amended}. Both texts are in \texttt{gold\_sections.jsonl}. The current text adds a paragraph that gives the lump sum to full siblings first when no heir of paragraph two exists.
\end{quote}

\subsection{Drift bands}
\label{app:drift-band-oracle}
Each gold citation is resolved to two versions of its section in the dated section corpus. One is the version in force on the decision date. The other is the current version, the one in force on 1 September 2026 in a law that is still active. A citation is \emph{repealed} when its law is no longer active or the section no longer exists. It is \emph{amended} when the two versions have different content hashes, and \emph{unchanged} otherwise. An item takes the most severe state among the citations its verdicts rely on, with repealed above amended above unchanged. Background citations do not set the band. The Repealed band also holds the few citations to a law that was already repealed on the decision date, because a court applies the law in force when the facts occurred (Penal Code \S2--3 set this rule for criminal matters). Such citations are rare. They are not drift, since the text did not change after the decision, but for a retriever over current law the text is missing all the same.

\section{Judging}
\label{app:judging}

\subsection{Bar Council eight-tier rubric}
\label{app:rubric}
Table~\ref{tab:rubric} in the body summarizes the grading regulation reproduced inline in the judge prompt.
The regulation was issued by the Office of Legal Education Training of the Thai Bar Council in 2014 and remains the standardized grading instrument for Thai legal essay answers.
We reproduce here an English translation of the scoring clause.

The grading considers the final answer and supporting reasons together, on a per-question basis:
\begin{enumerate}[leftmargin=*,itemsep=1pt]
    \item[(1)] Final answer differs from gold, reasoning unusable: $0$.
    \item[(2)] Final answer differs, reasoning somewhat acceptable: $1$ to $2$.
    \item[(3)] Final answer differs, reasoning good: $2$ to $4$.
    \item[(4)] Final answer matches gold, reasoning unusable: $0$ to $1$.
    \item[(5)] Final answer matches gold, reasoning somewhat acceptable: $2$ to $5$.
    \item[(6)] Final answer matches gold, reasoning acceptable: $5$ to $6$.
    \item[(7)] Final answer matches gold, reasoning good: $7$ to $8$.
    \item[(8)] Final answer matches gold, reasoning very good: $9$ to $10$.
\end{enumerate}
If a question contains multiple issues, the grader fixes the score for each issue first and then applies the criteria to each.
For items with several sub-questions, we apply the rubric per sub-question and average within an item.

\subsection{Judge prompt}
\label{app:judge-prompt}
The prompt is identical across judge models and across all items. Only the question text, decision date, gold answer, and model answer change per item. The judge is called once per answer at temperature $0$, with no tools, no retrieval, and no few-shot examples. It receives four inputs, the question, the decision date, the gold answer and the model answer, and returns for each sub-question a tier of the regulation, a score within that tier's range and a one-paragraph justification, as strict JSON. The system prompt and the message builder are published verbatim in the release, file \texttt{grading/judge\_prompt.md} (\S\ref{sec:release}).

\paragraph{Inputs the judge does not receive.}
The judge receives no information about the model that produced the answer, no information about the setting under which the answer was produced, no scores from any other judge, and no access to the FourCorners graph.
The citation scores are computed independently of the judge.

\subsection{Metric implementation}
\label{sec:metrics-details}
Citation scores are computed in two stages. Resolution maps each predicted citation to a section identifier in the corpus. Matching compares the resolved set with the gold set of the sub-question (\S\ref{sec:metrics}).

\begin{enumerate}[leftmargin=*,itemsep=1pt]
    \item \textbf{Resolution.} In every setting except Parametric the model copies a version identifier from the text it was given or retrieved. The scorer looks the identifier up in the dated section corpus and takes that version's section. In Parametric the model names a law and a section, or gives the section identifier. A law name is resolved against the law catalog of the corpus with \texttt{rapidfuzz.WRatio} (threshold 85). The section label is normalized (Thai numerals to Arabic digits, sub-paragraph markers removed). The section is then resolved to its version in force on the decision date.
    \item \textbf{Matching.} Citation F1 counts a match when the section identifiers agree. A predicted section that the sub-question's source cites only as background is removed from the prediction set first, so it is neither a hit nor a false positive.
    \item \textbf{Aggregation.} Precision, recall, and F1 are computed per sub-question, averaged within an item, averaged within a domain, and then macro-averaged over the seventeen domains.
\end{enumerate}

Edge cases. A version identifier that the corpus does not hold still resolves to its section. The 44 sub-questions whose gold has no resolvable citation are scored on the rubric only. An answer that does not parse scores zero on every citation metric.

\section{Serving and Compute}
\label{app:models}
Table~\ref{tab:model-configs} lists every evaluated model with its checkpoint, size, context window, reasoning mode, and tool-call parser.

\begin{table}[t]
\centering
\small
\setlength{\tabcolsep}{4pt}
\resizebox{\textwidth}{!}{%
\begin{tabular}{l l r r r l l}
\toprule
\textbf{Model} & \textbf{Checkpoint (Hugging Face)} & \textbf{Params} & \textbf{TP} & \textbf{Window} & \textbf{Reasoning} & \textbf{Tool parser} \\
\midrule
\multicolumn{7}{l}{\emph{General}} \\
Gemma 4 31B & \texttt{google/gemma-4-31B-it} & 31B & 2 & 262k & on (\texttt{gemma4}) & \texttt{gemma4} \\
Gemma 4 26B-A4B & \texttt{google/gemma-4-26B-A4B-it} & 26B / 4B & 2 & 262k & on (\texttt{gemma4}) & \texttt{gemma4} \\
Nemotron 3.5 Lightning 30B-A3B & \texttt{nvidia/NVIDIA-Nemotron-3.5-Lightning-30B-A3B-BF16} & 32B / 3B & 2 & 262k & on (\texttt{nemotron\_v3}) & \texttt{qwen3\_xml} \\
Muse Glimmer 30B & \texttt{meta-models/Muse-Glimmer-30B} & 30B & 2 & 131k & on (\texttt{muse\_glimmer}) & \texttt{muse\_glimmer} \\
\midrule
\multicolumn{7}{l}{\emph{Nemotron 3 bases and their adaptations}} \\
Nemotron 3 Nano 30B-A3B & \texttt{nvidia/NVIDIA-Nemotron-3-Nano-30B-A3B-BF16} & 32B / 3B & 2 & 262k & on (\texttt{nemotron\_v3}) & \texttt{qwen3\_coder} \\
\hspace{0.9em}$\hookrightarrow$\,SEA-LION v4.8 30B-A3B & \texttt{aisingapore/Nemotron-SEA-LION-v4.8-30B-A3B} & 33B / 3B & 2 & 262k & on (\texttt{nemotron\_v3}) & \texttt{qwen3\_coder} \\
OpenThai 2.0 Legal 30B-A3B & \texttt{iapp/openthai2.0-legal-thaillm-nemotron-3-nano-30b-a3b} & 32B / 3B & 2 & 262k & on (\texttt{nemotron\_v3}) & \texttt{qwen3\_coder} \\
Nemotron 3 Super 120B-A12B & \texttt{nvidia/NVIDIA-Nemotron-3-Super-120B-A12B-BF16} & 124B / 12B & 4 & 262k & on (\texttt{nemotron\_v3}) & \texttt{qwen3\_coder} \\
\hspace{0.9em}$\hookrightarrow$\,SEA-LION v4.8 120B-A12B & \texttt{aisingapore/Nemotron-SEA-LION-v4.8-120B-A12B} & 124B / 12B & 4 & 262k & on (\texttt{nemotron\_v3}) & \texttt{qwen3\_coder} \\
\midrule
\multicolumn{7}{l}{\emph{ThaiLLM 8B derivatives}} \\
Typhoon-S 8B & \texttt{typhoon-ai/typhoon-s-thaillm-8b-instruct-research-preview} & 8B & 1 & 32k & none & \texttt{hermes} \\
Pathumma 8B Think & \texttt{nectec/pathumma-thaillm-8b-think-3.0.0} & 8B & 1 & 40k & on (\texttt{qwen3}) & \texttt{hermes} \\
THaLLE 0.2 8B & \texttt{KBTG-Labs/THaLLE-0.2-ThaiLLM-8B-fa} & 8B & 1 & 40k & on (\texttt{qwen3}) & \texttt{hermes} \\
OpenThaiGPT 8B & \texttt{openthaigpt/openthaigpt-thaillm-8b-instruct-v0.7.2-research-preview} & 8B & 1 & 40k & on (\texttt{qwen3}) & \texttt{hermes} \\
\bottomrule
\end{tabular}
}
\caption{Serving configuration. All models are the bf16 release weights on vLLM 0.29.0 with each checkpoint's chat template and \texttt{generation\_config.json}. The harness sets no sampling parameter and allows 32{,}768 output tokens per call, reasoning included. Params are total parameters and, for mixture-of-experts models, active parameters per token. TP is the tensor-parallel degree. Window is the context length. Reasoning is the vLLM reasoning parser. Indented rows are post-trainings of the row above.}
\label{tab:model-configs}
\end{table}

\FloatBarrier

Every model runs locally with vLLM on one node of four A100 80GB GPUs. Table~\ref{tab:cost} reports, over the five settings and three samples, the tokens each model reads and writes, the share of what it writes that is reasoning, and the number of searches per item in the two RAG settings. The GPU hours column gives the wall-clock time of each model's runs on the node times the GPUs it occupied.

The two best models are also economical in the RAG In-force setting.
Gemma 4 31B reaches 4.55 with 7.8 thousand output tokens and 3.3 searches per item.
Muse Glimmer reaches 4.94 with 9.8 thousand output tokens, but its 15.6 searches per item bring 341 thousand input tokens of tool results, more than any other model.
Nemotron 3 Nano writes 30.7 thousand output tokens per item for a score of 1.66.
Output length does not predict the score.

\begin{table}[!ht]
\centering
\small
\setlength{\tabcolsep}{3.5pt}
\begin{tabular}{l r r r r r r}
\toprule
& \multicolumn{2}{c}{\textbf{Tokens (M)}} & \textbf{Reasoning} & \multicolumn{2}{c}{\textbf{Searches per item}} & \textbf{GPU} \\
\cmidrule(lr){2-3} \cmidrule(lr){4-4} \cmidrule(lr){5-6} \cmidrule(lr){7-7}
\textbf{Model} & \textbf{In} & \textbf{Out} & \textbf{share} & \textbf{RAG Cur.} & \textbf{RAG InF.} & \textbf{hours} \\
\midrule
\multicolumn{7}{l}{\textit{General}} \\
Gemma 4 31B & 416 & 126.4 & 84\% & 3.2 & 3.3 & 315 \\
Gemma 4 26B-A4B & 193 & 202.8 & 91\% & 1.3 & 1.2 & 200 \\
Nemotron 3.5 Lightning 30B-A3B & 1563 & 424.2 & 92\% & 5.7 & 5.1 & 380 \\
Muse Glimmer 30B & 3599 & 165.7 & 84\% & 16.4 & 15.6 & 455 \\
\midrule
\multicolumn{7}{l}{\textit{Nemotron 3 bases and their adaptations}} \\
Nemotron 3 Nano 30B-A3B & 729 & 389.1 & 93\% & 3.2 & 2.9 & 235 \\
SEA-LION v4.8 30B-A3B & 1251 & 330.5 & 92\% & 5.2 & 4.6 & 175 \\
OpenThai 2.0 Legal 30B-A3B & 101 & 195.0 & 93\% & 0.0 & 0.0 & 42 \\
Nemotron 3 Super 120B-A12B & 1853 & 183.7 & 83\% & 5.8 & 5.9 & 440 \\
SEA-LION v4.8 120B-A12B & 2117 & 156.5 & 86\% & 6.7 & 6.1 & 375 \\
\midrule
\multicolumn{7}{l}{\textit{ThaiLLM 8B derivatives}} \\
Typhoon-S 8B & 106 & 69.0 & 0\% & 0.2 & 0.3 & 60 \\
Pathumma 8B Think & 264 & 69.6 & 74\% & 2.9 & 2.8 & 40 \\
THaLLE 0.2 8B & 92 & 53.2 & 57\% & 0.0 & 0.0 & 25 \\
OpenThaiGPT 8B & 100 & 62.4 & 34\% & 0.0 & 0.0 & 36 \\
\bottomrule
\end{tabular}
\caption{Compute per model over the five settings and three samples. Tokens in include tool results. Reasoning share is the part of the written characters that is reasoning. Searches per item is the mean number of retrieval calls per item in the two RAG settings. GPU hours is the wall-clock time of the model's runs on the node times the GPUs it occupied.}
\label{tab:cost}
\end{table}

\FloatBarrier

\section{Leaderboard Tables}
\label{app:leaderboard}
Tables~\ref{tab:leaderboard} and \ref{tab:leaderboard-cost} give the numbers behind Figures~\ref{fig:leaderboard} and \ref{fig:cost}.

\begin{table}[t]
\centering
\small
\resizebox{\textwidth}{!}{%
\begin{tabular}{rllcccccc}
\toprule
 & & & \multicolumn{3}{c}{Bar-standard score} & \multicolumn{3}{c}{Cited sections (RAG In-force)} \\
\cmidrule(lr){4-6}\cmidrule(lr){7-9}
\# & Model & Access & RAG InF. & Parametric & GC InF. & F1 & Precision & Recall \\
\midrule
1 & Claude Fable 5.1 & Proprietary & \textbf{7.78} & \textbf{7.40} & \textbf{8.04} & 0.515 & 0.394 & 0.880 \\
2 & Gemini 4 Carbon & Proprietary & 7.68 & 6.64 & 7.62 & 0.516 & 0.406 & 0.868 \\
3 & Claude Fable 5.2 & Proprietary & 7.55 & 7.19 & 7.89 & 0.595 & 0.494 & 0.872 \\
4 & Gemini 4 Argon & Proprietary & 7.54 & 6.17 & 7.63 & 0.629 & 0.549 & 0.857 \\
5 & Gemini 4 Hydrogen & Proprietary & 7.53 & 6.63 & 7.58 & 0.480 & 0.361 & 0.878 \\
6 & GPT-6.1 Sol & Proprietary & 7.28 & 6.90 & 7.69 & \textbf{0.657} & 0.614 & 0.799 \\
7 & Claude Opus 5.5 & Proprietary & 7.03 & 6.25 & 7.56 & 0.585 & 0.500 & 0.809 \\
8 & DeepSeek V4.1 Flash & Open & 6.39 & 3.79 & 6.38 & 0.477 & 0.372 & 0.800 \\
9 & Muse Spark 1.3 & Proprietary & 6.37 & 5.28 & 7.09 & 0.576 & 0.530 & 0.720 \\
10 & GPT-6.1 Terra & Proprietary & 6.14 & 5.36 & 6.21 & 0.624 & 0.576 & 0.769 \\
11 & Muse Ember & Proprietary & 6.03 & 3.97 & 6.76 & 0.528 & 0.455 & 0.733 \\
12 & Claude Sonnet 5.5 & Proprietary & 5.97 & 4.79 & 6.67 & 0.583 & 0.531 & 0.747 \\
13 & DeepSeek V4 Flash & Open & 5.71 & 3.92 & 6.19 & 0.543 & 0.476 & 0.742 \\
14 & Qwen3.8 27B & Open & 5.48 & 3.18 & 6.04 & 0.480 & 0.402 & 0.704 \\
15 & GLM 5.3 FlashX Legal & Proprietary & 5.43 & 2.91 & 6.48 & 0.459 & 0.366 & 0.734 \\
16 & OpenThai 2.0 27B & Open & 5.10 & 2.96 & 5.85 & 0.509 & 0.459 & 0.665 \\
17 & Muse Glimmer 30B & Open & 4.93 & 3.35 & 5.83 & 0.541 & 0.536 & 0.620 \\
18 & GLM 5.3 Flash & Open & 4.80 & 3.08 & 5.89 & 0.414 & 0.355 & 0.577 \\
19 & Claude Haiku 5.5 & Proprietary & 4.62 & 3.49 & 5.36 & 0.513 & 0.454 & 0.692 \\
20 & Gemma 4 31B & Open & 4.57 & 3.66 & 5.16 & 0.503 & 0.642 & 0.460 \\
21 & Gemma 4 26B-A4B & Open & 3.68 & 3.44 & 4.84 & 0.254 & 0.358 & 0.217 \\
22 & Nemotron 3 Super 120B-A12B & Open & 3.23 & 2.05 & 4.29 & 0.424 & 0.498 & 0.421 \\
23 & SEA-LION v4.8 120B-A12B & Open & 2.88 & 2.04 & 4.29 & 0.351 & 0.394 & 0.367 \\
24 & Nemotron 3.5 Lightning 30B-A3B & Open & 2.47 & 1.73 & 3.49 & 0.342 & 0.420 & 0.330 \\
25 & SEA-LION v4.8 30B-A3B & Open & 2.45 & 1.66 & 3.53 & 0.335 & 0.401 & 0.330 \\
26 & Nemotron 3 Nano 30B-A3B & Open & 1.62 & 1.15 & 2.71 & 0.251 & 0.292 & 0.256 \\
27 & Typhoon-S 8B & Open & 1.54 & 1.74 & 2.77 & 0.034 & 0.045 & 0.030 \\
28 & Pathumma 8B Think & Open & 1.46 & 1.71 & 2.70 & 0.155 & 0.210 & 0.137 \\
29 & Nemotron 3 Nano Omni 30B-A3B & Open & 1.45 & 1.23 & 2.78 & 0.194 & 0.239 & 0.188 \\
30 & THaLLE 0.2 8B & Open & 1.10 & 1.18 & 2.62 & 0.003 & 0.004 & 0.003 \\
31 & Qwen3-8B & Proprietary & 0.96 & 1.17 & 2.57 & 0.000 & 0.000 & 0.000 \\
32 & OpenThaiGPT 8B & Open & 0.87 & 0.95 & 1.90 & 0.000 & 0.000 & 0.000 \\
33 & OpenThai 2.0 Legal 30B-A3B & Open & 0.87 & 1.18 & 1.92 & 0.005 & 0.006 & 0.004 \\
\bottomrule
\end{tabular}}
\caption{The \benchmark{} leaderboard as of 13 November 2026. Leaderboard protocol: the 500-item held-out test split, three settings, avg@3 for the 13 open-weight models of the benchmark protocol, Qwen3-8B and Nemotron 3 Nano Omni, and one sample for every other row (\S\ref{sec:protocols}). Bar-standard score 0--10 in each setting. Cited sections F1, precision and recall in RAG In-force. Best per column in bold.}
\label{tab:leaderboard}
\end{table}

\begin{table}[t]
\centering
\small
\resizebox{0.9\textwidth}{!}{%
\begin{tabular}{rlccccc}
\toprule
\# & Model & Bar-standard & Cost (USD) & Output tokens (K) & Searches & Hallucinated citations \\
\midrule
1 & Claude Fable 5.1 & 7.78 & 1.155 & 8.3 & 8.6 & 0.0\% \\
2 & Gemini 4 Carbon & 7.68 & 0.773 & 7.7 & 19.1 & 0.0\% \\
3 & Claude Fable 5.2 & 7.55 & 1.736 & 12.8 & 9.8 & 0.0\% \\
4 & Gemini 4 Argon & 7.54 & 0.338 & 5.2 & 7.6 & 0.0\% \\
5 & Gemini 4 Hydrogen & 7.53 & 0.893 & 42.6 & 45.5 & 0.0\% \\
6 & GPT-6.1 Sol & 7.28 & 0.134 & 4.1 & 6.3 & 0.0\% \\
7 & Claude Opus 5.5 & 7.03 & 0.411 & 5.3 & 5.3 & 0.0\% \\
8 & DeepSeek V4.1 Flash & 6.39 & 0.024 & 25.6 & 17.9 & 0.0\% \\
9 & Muse Spark 1.3 & 6.37 & 0.149 & 9.2 & 9.0 & 0.0\% \\
10 & GPT-6.1 Terra & 6.14 & 0.086 & 3.4 & 9.2 & 0.0\% \\
11 & Muse Ember & 6.03 & 0.043 & 24.2 & 26.8 & 0.0\% \\
12 & Claude Sonnet 5.5 & 5.97 & 0.204 & 5.1 & 5.7 & 0.0\% \\
13 & DeepSeek V4 Flash & 5.71 & 0.039 & 45.4 & 19.0 & 0.0\% \\
14 & Qwen3.8 27B & 5.48 & 0.235 & 52.3 & 17.6 & 0.1\% \\
15 & GLM 5.3 FlashX Legal & 5.43 & 0.171 & 44.3 & 16.1 & 0.0\% \\
16 & OpenThai 2.0 27B & 5.10 & 0.252 & 43.0 & 16.6 & 0.0\% \\
17 & Muse Glimmer 30B & 4.93 & 0.120 & 10.1 & 15.8 & 0.0\% \\
18 & GLM 5.3 Flash & 4.80 & 0.044 & 35.3 & 10.3 & 1.1\% \\
19 & Claude Haiku 5.5 & 4.62 & 0.024 & 12.3 & 7.4 & 0.0\% \\
20 & Gemma 4 31B & 4.57 & 0.011 & 7.9 & 3.4 & 0.0\% \\
21 & Gemma 4 26B-A4B & 3.68 & 0.005 & 11.8 & 1.2 & 0.3\% \\
22 & Nemotron 3 Super 120B-A12B & 3.23 & 0.096 & 12.9 & 6.5 & 0.0\% \\
23 & SEA-LION v4.8 120B-A12B & 2.88 & 0.084 & 8.3 & 6.2 & 1.0\% \\
24 & Nemotron 3.5 Lightning 30B-A3B & 2.47 & 0.015 & 22.6 & 5.3 & 0.7\% \\
25 & SEA-LION v4.8 30B-A3B & 2.45 & 0.010 & 15.3 & 4.7 & 0.1\% \\
26 & Nemotron 3 Nano 30B-A3B & 1.62 & 0.011 & 30.1 & 2.9 & 0.5\% \\
27 & Typhoon-S 8B & 1.54 & 0.001 & 2.6 & 0.4 & 6.8\% \\
28 & Pathumma 8B Think & 1.46 & 0.001 & 2.7 & 2.8 & 2.0\% \\
29 & Nemotron 3 Nano Omni 30B-A3B & 1.45 & 0.009 & 20.5 & 2.9 & 6.7\% \\
30 & THaLLE 0.2 8B & 1.10 & 0.000 & 2.8 & 0.0 & 11.3\% \\
31 & Qwen3-8B & 0.96 & 0.000 & 2.9 & 0.0 & 19.2\% \\
32 & OpenThaiGPT 8B & 0.87 & 0.001 & 3.5 & -- & 23.8\% \\
33 & OpenThai 2.0 Legal 30B-A3B & 0.87 & 0.010 & 11.5 & 0.0 & 19.4\% \\
\bottomrule
\end{tabular}}
\caption{Cost and behavior of each leaderboard row in RAG In-force (leaderboard protocol, 500 items). Cost is US dollars per item (basis in \S\ref{subsec:cost}). Output tokens per item in thousands. Searches are retrieval calls per item. Hallucinated citations is the share of cited sections that do not exist in the statute corpus (\S\ref{subsec:halluc}).}
\label{tab:leaderboard-cost}
\end{table}

\FloatBarrier

\section{Additional Results}
\label{app:results}

\subsection{Score by model and domain}
\label{app:matrix}
Table~\ref{tab:domains} gives the mean Bar-standard score per domain over the 13 open-weight models of the benchmark protocol in three settings, and Table~\ref{tab:matrix-rag} the score of every model in every domain in RAG In-force.

\begin{table}[t]
\centering
\small
\setlength{\tabcolsep}{4pt}
\begin{tabular}{@{}l r ccc l c c@{}}
\toprule
\headrow & & \multicolumn{3}{c}{\textbf{Mean Bar-standard}} & \multicolumn{2}{c}{\textbf{Best (RAG InF.)}} & \\
\headrow \textbf{Domain} & \textbf{Items} & Param. & GC InF. & RAG InF. & Model & Score & \textbf{Lowest} \\
\midrule
Family \& Succession & 146 & 2.02 & 3.99 & 2.72 & Muse Glimmer 30B & 5.09 & 0.90 \\
Intellectual Property & 109 & 2.31 & 3.66 & 2.70 & Muse Glimmer 30B & 5.30 & 1.04 \\
Public Law & 98 & 2.26 & 3.73 & 2.69 & Muse Glimmer 30B & 5.49 & 0.85 \\
Natural Resources \& Licensing & 84 & 2.07 & 3.82 & 2.66 & Muse Glimmer 30B & 5.39 & 0.93 \\
Traffic \& Transport & 63 & 1.88 & 3.57 & 2.62 & Muse Glimmer 30B & 5.45 & 0.88 \\
Labor & 116 & 1.97 & 3.56 & 2.56 & Muse Glimmer 30B & 5.10 & 0.89 \\
Civil \& Commercial & 144 & 1.99 & 3.76 & 2.49 & Gemma 4 31B & 4.92 & 0.82 \\
Criminal & 146 & 2.02 & 3.51 & 2.47 & Muse Glimmer 30B & 4.98 & 0.81 \\
Tax \& Revenue & 113 & 1.94 & 3.55 & 2.45 & Muse Glimmer 30B & 4.98 & 0.79 \\
Securities \& Finance & 56 & 2.01 & 3.56 & 2.44 & Muse Glimmer 30B & 4.88 & 0.73 \\
Civil Procedure & 87 & 1.95 & 3.55 & 2.43 & Muse Glimmer 30B & 5.00 & 0.81 \\
Administrative Acts & 82 & 2.03 & 3.50 & 2.40 & Muse Glimmer 30B & 4.94 & 0.82 \\
Land \& Property & 117 & 1.84 & 3.38 & 2.36 & Gemma 4 31B & 4.46 & 0.74 \\
State Liability & 80 & 2.08 & 3.39 & 2.31 & Muse Glimmer 30B & 4.73 & 0.76 \\
Narcotics & 74 & 1.80 & 3.13 & 2.30 & Muse Glimmer 30B & 4.70 & 0.83 \\
Administrative Contracts & 74 & 2.00 & 3.30 & 2.21 & Muse Glimmer 30B & 4.46 & 0.78 \\
Public Personnel & 81 & 1.81 & 2.97 & 1.99 & Muse Glimmer 30B & 4.22 & 0.53 \\
\midrule
\textbf{All domains} & 1{,}670 & 2.00 & 3.52 & 2.46 & & & \\
\bottomrule
\end{tabular}
\caption{Bar-standard score by domain, mean over the 13 models, avg@3, in the Parametric, Gold-cited In-force and RAG In-force settings. Domains are ordered from the easiest to the hardest in RAG In-force. Best is the highest RAG In-force score of any model in the domain and the model that reached it. Lowest is the lowest.}
\label{tab:domains}
\end{table}

\begin{table}[htbp]
\centering
\scriptsize
\setlength{\tabcolsep}{2.5pt}
\resizebox{\textwidth}{!}{\begin{tabular}{@{}l cccccccccccccccccc@{}}
\toprule
\headrow \textbf{Model} & \textbf{All} & \rotatebox{90}{Family \& Succession} & \rotatebox{90}{Intellectual Property} & \rotatebox{90}{Public Law} & \rotatebox{90}{Natural Resources \& Licensing} & \rotatebox{90}{Traffic \& Transport} & \rotatebox{90}{Labor} & \rotatebox{90}{Civil \& Commercial} & \rotatebox{90}{Criminal} & \rotatebox{90}{Tax \& Revenue} & \rotatebox{90}{Securities \& Finance} & \rotatebox{90}{Civil Procedure} & \rotatebox{90}{Administrative Acts} & \rotatebox{90}{Land \& Property} & \rotatebox{90}{State Liability} & \rotatebox{90}{Narcotics} & \rotatebox{90}{Administrative Contracts} & \rotatebox{90}{Public Personnel} \\
\midrule
Muse Glimmer 30B & \cellcolor{ciGreen!47}4.94 & \cellcolor{ciGreen!50}5.09 & \cellcolor{ciGreen!55}5.30 & \cellcolor{ciGreen!60}5.49 & \cellcolor{ciGreen!58}5.39 & \cellcolor{ciGreen!59}5.45 & \cellcolor{ciGreen!51}5.10 & \cellcolor{ciGreen!45}4.87 & \cellcolor{ciGreen!48}4.98 & \cellcolor{ciGreen!48}4.98 & \cellcolor{ciGreen!45}4.88 & \cellcolor{ciGreen!48}5.00 & \cellcolor{ciGreen!47}4.94 & \cellcolor{ciGreen!35}4.45 & \cellcolor{ciGreen!42}4.73 & \cellcolor{ciGreen!41}4.70 & \cellcolor{ciGreen!35}4.46 & \cellcolor{ciGreen!29}4.22 \\
Gemma 4 31B & \cellcolor{ciGreen!37}4.55 & \cellcolor{ciGreen!49}5.05 & \cellcolor{ciGreen!44}4.83 & \cellcolor{ciGreen!44}4.85 & \cellcolor{ciGreen!42}4.75 & \cellcolor{ciGreen!38}4.59 & \cellcolor{ciGreen!42}4.76 & \cellcolor{ciGreen!46}4.92 & \cellcolor{ciGreen!42}4.73 & \cellcolor{ciGreen!38}4.59 & \cellcolor{ciGreen!40}4.64 & \cellcolor{ciGreen!36}4.50 & \cellcolor{ciGreen!33}4.37 & \cellcolor{ciGreen!35}4.46 & \cellcolor{ciGreen!29}4.20 & \cellcolor{ciGreen!28}4.16 & \cellcolor{ciGreen!29}4.23 & \cellcolor{ciGreen!19}3.79 \\
Gemma 4 26B-A4B & \cellcolor{ciGreen!17}3.71 & \cellcolor{ciGreen!23}3.98 & \cellcolor{ciGreen!25}4.04 & \cellcolor{ciGreen!26}4.08 & \cellcolor{ciGreen!19}3.78 & \cellcolor{ciGreen!19}3.77 & \cellcolor{ciGreen!20}3.85 & \cellcolor{ciGreen!22}3.94 & \cellcolor{ciGreen!22}3.92 & \cellcolor{ciGreen!18}3.76 & \cellcolor{ciGreen!18}3.75 & \cellcolor{ciGreen!18}3.74 & \cellcolor{ciGreen!16}3.68 & \cellcolor{ciGreen!14}3.58 & \cellcolor{ciGreen!11}3.47 & \cellcolor{ciGreen!8}3.33 & \cellcolor{ciGreen!12}3.51 & \cellcolor{ciOrange!1}2.97 \\
Nemotron 3 Super & \cellcolor{ciGreen!7}3.31 & \cellcolor{ciGreen!17}3.70 & \cellcolor{ciGreen!11}3.45 & \cellcolor{ciGreen!12}3.52 & \cellcolor{ciGreen!12}3.52 & \cellcolor{ciGreen!19}3.79 & \cellcolor{ciGreen!11}3.48 & \cellcolor{ciGreen!9}3.37 & \cellcolor{ciGreen!9}3.39 & \cellcolor{ciGreen!6}3.25 & \cellcolor{ciGreen!8}3.32 & \cellcolor{ciGreen!7}3.31 & \cellcolor{ciGreen!3}3.13 & \cellcolor{ciGreen!5}3.20 & \cellcolor{ciOrange!0}3.00 & \cellcolor{ciGreen!5}3.21 & \cellcolor{ciOrange!2}2.93 & \cellcolor{ciOrange!9}2.63 \\
SEA-LION v4.8 120B & \cellcolor{ciOrange!2}2.94 & \cellcolor{ciGreen!4}3.19 & \cellcolor{ciGreen!2}3.10 & \cellcolor{ciGreen!10}3.43 & \cellcolor{ciGreen!11}3.47 & \cellcolor{ciGreen!8}3.33 & \cellcolor{ciOrange!0}3.01 & \cellcolor{ciOrange!2}2.93 & \cellcolor{ciOrange!4}2.84 & \cellcolor{ciOrange!1}2.98 & \cellcolor{ciOrange!0}2.99 & \cellcolor{ciOrange!4}2.86 & \cellcolor{ciOrange!5}2.80 & \cellcolor{ciOrange!5}2.79 & \cellcolor{ciOrange!7}2.71 & \cellcolor{ciOrange!7}2.73 & \cellcolor{ciOrange!11}2.56 & \cellcolor{ciOrange!19}2.21 \\
SEA-LION v4.8 30B & \cellcolor{ciOrange!12}2.50 & \cellcolor{ciOrange!4}2.84 & \cellcolor{ciOrange!5}2.82 & \cellcolor{ciOrange!9}2.64 & \cellcolor{ciOrange!5}2.81 & \cellcolor{ciOrange!9}2.66 & \cellcolor{ciOrange!10}2.62 & \cellcolor{ciOrange!12}2.52 & \cellcolor{ciOrange!16}2.35 & \cellcolor{ciOrange!13}2.46 & \cellcolor{ciOrange!12}2.51 & \cellcolor{ciOrange!17}2.31 & \cellcolor{ciOrange!11}2.54 & \cellcolor{ciOrange!11}2.55 & \cellcolor{ciOrange!12}2.53 & \cellcolor{ciOrange!18}2.28 & \cellcolor{ciOrange!23}2.08 & \cellcolor{ciOrange!25}1.96 \\
Nemotron 3.5 Lightning & \cellcolor{ciOrange!13}2.47 & \cellcolor{ciOrange!5}2.81 & \cellcolor{ciOrange!8}2.70 & \cellcolor{ciOrange!11}2.57 & \cellcolor{ciOrange!7}2.74 & \cellcolor{ciOrange!7}2.70 & \cellcolor{ciOrange!10}2.61 & \cellcolor{ciOrange!13}2.46 & \cellcolor{ciOrange!12}2.51 & \cellcolor{ciOrange!11}2.56 & \cellcolor{ciOrange!13}2.46 & \cellcolor{ciOrange!16}2.36 & \cellcolor{ciOrange!14}2.45 & \cellcolor{ciOrange!14}2.45 & \cellcolor{ciOrange!21}2.13 & \cellcolor{ciOrange!16}2.34 & \cellcolor{ciOrange!22}2.09 & \cellcolor{ciOrange!24}2.02 \\
Nemotron 3 Nano & \cellcolor{ciOrange!33}1.66 & \cellcolor{ciOrange!24}2.02 & \cellcolor{ciOrange!25}1.97 & \cellcolor{ciOrange!32}1.69 & \cellcolor{ciOrange!29}1.81 & \cellcolor{ciOrange!31}1.74 & \cellcolor{ciOrange!31}1.74 & \cellcolor{ciOrange!33}1.65 & \cellcolor{ciOrange!33}1.64 & \cellcolor{ciOrange!32}1.68 & \cellcolor{ciOrange!35}1.58 & \cellcolor{ciOrange!36}1.51 & \cellcolor{ciOrange!34}1.63 & \cellcolor{ciOrange!32}1.70 & \cellcolor{ciOrange!36}1.54 & \cellcolor{ciOrange!33}1.64 & \cellcolor{ciOrange!42}1.28 & \cellcolor{ciOrange!40}1.34 \\
Typhoon-S 8B & \cellcolor{ciOrange!35}1.56 & \cellcolor{ciOrange!33}1.64 & \cellcolor{ciOrange!26}1.93 & \cellcolor{ciOrange!26}1.92 & \cellcolor{ciOrange!35}1.56 & \cellcolor{ciOrange!34}1.59 & \cellcolor{ciOrange!35}1.57 & \cellcolor{ciOrange!38}1.45 & \cellcolor{ciOrange!37}1.50 & \cellcolor{ciOrange!37}1.49 & \cellcolor{ciOrange!37}1.49 & \cellcolor{ciOrange!35}1.56 & \cellcolor{ciOrange!38}1.44 & \cellcolor{ciOrange!38}1.45 & \cellcolor{ciOrange!32}1.69 & \cellcolor{ciOrange!41}1.30 & \cellcolor{ciOrange!34}1.59 & \cellcolor{ciOrange!41}1.31 \\
Pathumma 8B Think & \cellcolor{ciOrange!37}1.47 & \cellcolor{ciOrange!30}1.76 & \cellcolor{ciOrange!36}1.51 & \cellcolor{ciOrange!35}1.55 & \cellcolor{ciOrange!32}1.69 & \cellcolor{ciOrange!35}1.56 & \cellcolor{ciOrange!33}1.66 & \cellcolor{ciOrange!39}1.41 & \cellcolor{ciOrange!36}1.52 & \cellcolor{ciOrange!40}1.34 & \cellcolor{ciOrange!41}1.33 & \cellcolor{ciOrange!36}1.53 & \cellcolor{ciOrange!40}1.35 & \cellcolor{ciOrange!38}1.46 & \cellcolor{ciOrange!42}1.28 & \cellcolor{ciOrange!37}1.50 & \cellcolor{ciOrange!42}1.28 & \cellcolor{ciOrange!42}1.28 \\
THaLLE 0.2 8B & \cellcolor{ciOrange!46}1.11 & \cellcolor{ciOrange!42}1.26 & \cellcolor{ciOrange!40}1.34 & \cellcolor{ciOrange!40}1.34 & \cellcolor{ciOrange!45}1.17 & \cellcolor{ciOrange!47}1.05 & \cellcolor{ciOrange!47}1.09 & \cellcolor{ciOrange!46}1.10 & \cellcolor{ciOrange!47}1.06 & \cellcolor{ciOrange!47}1.08 & \cellcolor{ciOrange!47}1.08 & \cellcolor{ciOrange!47}1.06 & \cellcolor{ciOrange!45}1.15 & \cellcolor{ciOrange!49}0.97 & \cellcolor{ciOrange!45}1.13 & \cellcolor{ciOrange!49}1.00 & \cellcolor{ciOrange!47}1.07 & \cellcolor{ciOrange!50}0.94 \\
OpenThaiGPT 8B & \cellcolor{ciOrange!52}0.88 & \cellcolor{ciOrange!51}0.90 & \cellcolor{ciOrange!47}1.06 & \cellcolor{ciOrange!48}1.02 & \cellcolor{ciOrange!50}0.93 & \cellcolor{ciOrange!52}0.88 & \cellcolor{ciOrange!51}0.89 & \cellcolor{ciOrange!53}0.82 & \cellcolor{ciOrange!53}0.81 & \cellcolor{ciOrange!52}0.86 & \cellcolor{ciOrange!50}0.96 & \cellcolor{ciOrange!53}0.81 & \cellcolor{ciOrange!50}0.94 & \cellcolor{ciOrange!55}0.74 & \cellcolor{ciOrange!51}0.90 & \cellcolor{ciOrange!53}0.83 & \cellcolor{ciOrange!53}0.83 & \cellcolor{ciOrange!55}0.72 \\
OpenThai 2.0 Legal & \cellcolor{ciOrange!52}0.88 & \cellcolor{ciOrange!47}1.06 & \cellcolor{ciOrange!48}1.04 & \cellcolor{ciOrange!52}0.85 & \cellcolor{ciOrange!49}0.98 & \cellcolor{ciOrange!50}0.93 & \cellcolor{ciOrange!50}0.93 & \cellcolor{ciOrange!50}0.94 & \cellcolor{ciOrange!51}0.90 & \cellcolor{ciOrange!54}0.79 & \cellcolor{ciOrange!55}0.73 & \cellcolor{ciOrange!47}1.07 & \cellcolor{ciOrange!53}0.82 & \cellcolor{ciOrange!52}0.84 & \cellcolor{ciOrange!54}0.76 & \cellcolor{ciOrange!50}0.95 & \cellcolor{ciOrange!54}0.78 & \cellcolor{ciOrange!60}0.53 \\
\midrule
\textbf{Mean} & \cellcolor{ciOrange!13}2.46 & \cellcolor{ciOrange!7}2.72 & \cellcolor{ciOrange!8}2.70 & \cellcolor{ciOrange!8}2.69 & \cellcolor{ciOrange!8}2.66 & \cellcolor{ciOrange!9}2.62 & \cellcolor{ciOrange!11}2.56 & \cellcolor{ciOrange!13}2.49 & \cellcolor{ciOrange!13}2.47 & \cellcolor{ciOrange!14}2.45 & \cellcolor{ciOrange!14}2.44 & \cellcolor{ciOrange!14}2.43 & \cellcolor{ciOrange!15}2.40 & \cellcolor{ciOrange!16}2.36 & \cellcolor{ciOrange!17}2.31 & \cellcolor{ciOrange!17}2.30 & \cellcolor{ciOrange!19}2.21 & \cellcolor{ciOrange!25}1.99 \\
\bottomrule
\end{tabular}}
\caption{Bar-standard score by model and domain, RAG In-force, avg@3. Columns follow the order of Table~\ref{tab:domains}, rows the order of Table~\ref{tab:main-results}. Cells run from orange (lowest) through white to green (highest). The last row is the mean over models.}
\label{tab:matrix-rag}
\end{table}

\FloatBarrier

Table~\ref{tab:leaderboard-domains} gives the RAG In-force score of every leaderboard row in every domain.

\begin{table}[p]
\centering
\scriptsize
\resizebox{\textwidth}{!}{%
\begin{tabular}{lccccccccccccccccc}
\toprule
Model & \rotatebox{90}{Land \& Property} & \rotatebox{90}{Intellectual Property} & \rotatebox{90}{Family \& Succession} & \rotatebox{90}{Tax \& Revenue} & \rotatebox{90}{Civil \& Commercial} & \rotatebox{90}{Criminal} & \rotatebox{90}{Labor} & \rotatebox{90}{Securities \& Finance} & \rotatebox{90}{Narcotics} & \rotatebox{90}{Civil Procedure} & \rotatebox{90}{Natural Resources \& Licensing} & \rotatebox{90}{Traffic \& Transport} & \rotatebox{90}{Public Law} & \rotatebox{90}{Administrative Acts} & \rotatebox{90}{Administrative Contracts} & \rotatebox{90}{Public Personnel} & \rotatebox{90}{State Liability} \\
\midrule
Claude Fable 5.1 & 7.83 & 8.08 & 8.09 & 7.28 & 7.81 & 7.41 & 8.06 & 7.83 & 7.59 & 7.75 & 8.02 & 8.13 & 8.17 & 8.10 & 7.16 & 7.21 & 7.80 \\
Gemini 4 Carbon & 7.69 & 7.72 & 7.32 & 7.26 & 7.72 & 7.45 & 8.54 & 8.25 & 7.60 & 8.31 & 7.72 & 7.64 & 7.57 & 7.68 & 7.08 & 7.23 & 7.77 \\
Claude Fable 5.2 & 7.37 & 7.78 & 7.68 & 6.93 & 7.87 & 7.38 & 8.21 & 7.73 & 7.24 & 8.25 & 7.80 & 7.45 & 7.93 & 7.21 & 6.86 & 7.00 & 7.69 \\
Gemini 4 Argon & 7.23 & 7.57 & 7.62 & 6.76 & 7.68 & 7.51 & 8.28 & 7.65 & 6.39 & 7.82 & 8.04 & 7.62 & 8.16 & 7.67 & 7.00 & 7.45 & 7.71 \\
Gemini 4 Hydrogen & 7.45 & 7.31 & 7.52 & 6.75 & 7.93 & 7.23 & 8.25 & 7.66 & 7.08 & 8.16 & 7.72 & 7.60 & 7.66 & 7.91 & 6.94 & 7.17 & 7.70 \\
GPT-6.1 Sol & 7.01 & 7.68 & 6.96 & 6.38 & 7.55 & 7.31 & 7.75 & 7.70 & 6.50 & 7.52 & 7.77 & 7.39 & 7.84 & 7.84 & 6.40 & 6.98 & 7.19 \\
Claude Opus 5.5 & 6.75 & 7.49 & 6.95 & 6.12 & 7.30 & 6.74 & 7.85 & 7.13 & 6.74 & 7.20 & 7.59 & 7.23 & 7.25 & 7.01 & 6.28 & 6.76 & 7.16 \\
DeepSeek V4.1 Flash & 5.97 & 6.37 & 6.18 & 5.93 & 6.64 & 5.67 & 6.67 & 6.49 & 6.58 & 6.62 & 6.84 & 6.85 & 6.87 & 6.76 & 5.58 & 6.23 & 6.36 \\
Muse Spark 1.3 & 5.97 & 6.82 & 6.35 & 5.85 & 6.83 & 6.31 & 6.58 & 6.79 & 6.00 & 5.72 & 6.77 & 6.92 & 6.85 & 6.12 & 5.95 & 6.07 & 6.32 \\
GPT-6.1 Terra & 6.02 & 6.52 & 5.93 & 5.57 & 6.34 & 5.97 & 6.23 & 6.40 & 5.45 & 5.74 & 6.50 & 6.61 & 6.46 & 6.42 & 5.96 & 6.05 & 6.19 \\
Muse Ember & 5.40 & 5.96 & 5.94 & 5.19 & 6.43 & 5.67 & 7.07 & 6.27 & 5.53 & 5.67 & 6.98 & 5.93 & 6.67 & 6.41 & 5.83 & 5.74 & 5.90 \\
Claude Sonnet 5.5 & 5.36 & 6.25 & 6.05 & 5.49 & 6.37 & 5.38 & 6.35 & 6.54 & 5.67 & 5.87 & 6.47 & 6.09 & 6.39 & 6.23 & 5.58 & 5.44 & 6.03 \\
DeepSeek V4 Flash & 5.76 & 6.02 & 5.66 & 5.17 & 6.11 & 5.73 & 6.12 & 5.78 & 4.96 & 5.41 & 6.32 & 5.71 & 6.25 & 6.30 & 5.29 & 4.80 & 5.66 \\
Qwen3.8 27B & 5.68 & 5.06 & 5.78 & 5.20 & 5.88 & 4.95 & 5.75 & 5.26 & 4.28 & 5.18 & 5.89 & 6.02 & 6.15 & 6.11 & 5.42 & 4.99 & 5.65 \\
GLM 5.3 FlashX Legal & 5.58 & 5.90 & 5.20 & 5.28 & 6.25 & 4.42 & 5.70 & 5.92 & 5.09 & 4.98 & 6.07 & 4.74 & 6.05 & 5.71 & 5.01 & 5.24 & 5.14 \\
OpenThai 2.0 27B & 5.27 & 5.12 & 5.41 & 4.83 & 4.54 & 4.15 & 5.18 & 5.39 & 5.00 & 5.35 & 5.93 & 5.68 & 5.88 & 4.84 & 4.69 & 4.47 & 4.99 \\
Muse Glimmer 30B & 4.77 & 5.36 & 4.98 & 4.45 & 4.53 & 4.70 & 5.04 & 5.14 & 4.69 & 5.02 & 5.42 & 5.53 & 5.45 & 5.35 & 4.58 & 4.36 & 4.49 \\
GLM 5.3 Flash & 4.62 & 5.97 & 5.34 & 4.88 & 5.21 & 3.92 & 4.67 & 4.60 & 3.89 & 4.99 & 5.45 & 4.75 & 5.87 & 4.95 & 4.33 & 4.11 & 4.07 \\
Claude Haiku 5.5 & 4.64 & 4.51 & 4.49 & 4.17 & 4.80 & 4.42 & 5.05 & 4.62 & 4.41 & 4.59 & 4.78 & 5.26 & 5.25 & 4.68 & 4.47 & 4.05 & 4.41 \\
Gemma 4 31B & 4.56 & 4.64 & 4.88 & 4.30 & 4.81 & 4.86 & 4.71 & 4.89 & 4.09 & 4.65 & 5.05 & 4.91 & 4.62 & 4.22 & 4.54 & 3.84 & 4.11 \\
Gemma 4 26B-A4B & 3.59 & 3.58 & 3.77 & 3.60 & 3.76 & 4.06 & 4.24 & 3.83 & 3.16 & 3.75 & 3.88 & 4.05 & 3.52 & 3.60 & 3.81 & 2.95 & 3.36 \\
Nemotron 3 Super 120B-A12B & 3.35 & 3.23 & 3.80 & 3.12 & 3.31 & 2.99 & 3.15 & 3.32 & 3.00 & 3.21 & 3.33 & 3.76 & 3.33 & 3.28 & 3.07 & 2.87 & 2.84 \\
SEA-LION v4.8 120B-A12B & 2.89 & 2.98 & 3.18 & 2.86 & 2.71 & 2.63 & 2.74 & 2.90 & 2.74 & 2.59 & 3.48 & 3.53 & 2.94 & 3.00 & 2.74 & 2.30 & 2.77 \\
Nemotron 3.5 Lightning 30B-A3B & 2.61 & 2.50 & 2.94 & 2.31 & 2.44 & 2.45 & 2.48 & 2.61 & 2.31 & 2.29 & 2.62 & 2.69 & 2.50 & 2.56 & 2.26 & 2.17 & 2.32 \\
SEA-LION v4.8 30B-A3B & 2.77 & 2.63 & 2.76 & 2.20 & 2.27 & 2.28 & 2.19 & 2.55 & 2.31 & 2.41 & 2.56 & 2.68 & 2.56 & 2.55 & 2.02 & 2.29 & 2.55 \\
Nemotron 3 Nano 30B-A3B & 1.64 & 2.06 & 2.02 & 1.39 & 1.62 & 1.63 & 1.43 & 1.78 & 1.59 & 1.46 & 1.82 & 1.60 & 1.71 & 1.63 & 1.33 & 1.38 & 1.49 \\
Typhoon-S 8B & 1.53 & 1.82 & 1.66 & 1.35 & 1.44 & 1.60 & 1.62 & 1.54 & 1.36 & 1.38 & 1.62 & 1.55 & 1.76 & 1.50 & 1.56 & 1.23 & 1.62 \\
Pathumma 8B Think & 1.59 & 1.38 & 1.87 & 1.30 & 1.18 & 1.46 & 1.70 & 1.33 & 1.63 & 1.51 & 1.63 & 1.66 & 1.49 & 1.40 & 1.37 & 1.15 & 1.23 \\
Nemotron 3 Nano Omni 30B-A3B & 1.56 & 1.65 & 1.80 & 1.44 & 1.23 & 1.27 & 1.47 & 1.32 & 1.13 & 1.25 & 1.95 & 1.56 & 1.47 & 1.58 & 1.16 & 1.23 & 1.49 \\
THaLLE 0.2 8B & 0.81 & 1.25 & 1.12 & 1.10 & 1.09 & 1.12 & 1.14 & 1.03 & 1.07 & 1.09 & 1.18 & 1.01 & 1.18 & 1.37 & 1.14 & 0.85 & 1.19 \\
Qwen3-8B & 0.77 & 1.14 & 1.00 & 0.92 & 0.97 & 0.95 & 0.87 & 0.94 & 1.03 & 0.64 & 1.16 & 0.80 & 1.19 & 1.03 & 1.02 & 0.71 & 1.17 \\
OpenThaiGPT 8B & 0.73 & 0.87 & 0.88 & 0.76 & 0.72 & 0.86 & 0.93 & 0.96 & 1.06 & 0.71 & 0.98 & 0.85 & 0.88 & 1.07 & 0.96 & 0.70 & 0.91 \\
OpenThai 2.0 Legal 30B-A3B & 0.77 & 0.88 & 0.95 & 0.79 & 0.87 & 0.97 & 0.98 & 0.77 & 1.04 & 1.15 & 0.92 & 0.93 & 0.85 & 0.76 & 0.84 & 0.55 & 0.81 \\
\bottomrule
\end{tabular}}
\caption{Bar-standard score in RAG In-force by leaderboard row and domain (leaderboard protocol, 500 items). Domain order follows Table~\ref{tab:domain-inventory}.}
\label{tab:leaderboard-domains}
\end{table}

\FloatBarrier

\subsection{Best of three samples}
\label{app:bestk}
Table~\ref{tab:best3} takes the best of the three samples of each item and then aggregates as for avg@3. It is a capability ceiling. No claim in the report rests on it.

\begin{table}[t]
\centering
\small
\setlength{\tabcolsep}{3.5pt}
\resizebox{\textwidth}{!}{%
\begin{tabular}{l ccccc c ccccc}
\toprule
& \multicolumn{5}{c}{\textbf{Citation F1, best@3}} && \multicolumn{5}{c}{\textbf{Bar-standard, best@3}} \\
\cmidrule(lr){2-6} \cmidrule(lr){8-12}
\textbf{Model} & \textbf{Param.} & \textbf{GC Cur.} & \textbf{GC InF.} & \textbf{RAG Cur.} & \textbf{RAG InF.} && \textbf{Param.} & \textbf{GC Cur.} & \textbf{GC InF.} & \textbf{RAG Cur.} & \textbf{RAG InF.} \\
\midrule
\multicolumn{12}{l}{\textit{General}} \\
Gemma 4 31B & \textbf{0.373} & 0.715 & 0.861 & \underline{0.506} & \underline{0.604} && \textbf{4.38} & \underline{5.46} & \underline{6.05} & \underline{4.93} & \underline{5.33} \\
Gemma 4 26B-A4B & \underline{0.341} & 0.712 & 0.850 & 0.352 & 0.414 && \underline{4.11} & 5.07 & 5.53 & 4.37 & 4.52 \\
Nemotron 3.5 Lightning 30B-A3B & 0.054 & 0.717 & 0.842 & 0.377 & 0.466 && 2.28 & 3.81 & 4.21 & 2.95 & 3.18 \\
Muse Glimmer 30B & 0.271 & \textbf{0.773} & \textbf{0.914} & \textbf{0.521} & \textbf{0.632} && 4.00 & \textbf{5.94} & \textbf{6.58} & \textbf{5.16} & \textbf{5.78} \\
\midrule
\multicolumn{12}{l}{\textit{Nemotron 3 bases and their adaptations}} \\
Nemotron 3 Nano 30B-A3B & 0.036 & 0.719 & 0.842 & 0.306 & 0.379 && 1.70 & 3.09 & 3.39 & 2.09 & 2.26 \\
\hspace{0.9em}$\hookrightarrow$\,SEA-LION v4.8 30B-A3B & 0.071 & 0.716 & 0.839 & 0.383 & 0.467 && 2.23 & 3.96 & 4.33 & 2.95 & 3.29 \\
OpenThai 2.0 Legal 30B-A3B & 0.143 & 0.627 & 0.705 & 0.011 & 0.013 && 1.85 & 2.76 & 2.90 & 1.49 & 1.50 \\
Nemotron 3 Super 120B-A12B & 0.160 & \underline{0.740} & \underline{0.874} & 0.453 & 0.550 && 2.62 & 4.57 & 5.10 & 3.70 & 4.12 \\
\hspace{0.9em}$\hookrightarrow$\,SEA-LION v4.8 120B-A12B & 0.153 & 0.733 & 0.873 & 0.439 & 0.530 && 2.65 & 4.61 & 5.16 & 3.61 & 4.00 \\
\midrule
\multicolumn{12}{l}{\textit{ThaiLLM 8B derivatives}} \\
Typhoon-S 8B & 0.163 & 0.688 & 0.796 & 0.034 & 0.092 && 2.26 & 3.36 & 3.58 & 2.18 & 2.22 \\
Pathumma 8B Think & 0.235 & 0.686 & 0.787 & 0.228 & 0.285 && 2.34 & 3.43 & 3.57 & 2.17 & 2.26 \\
THaLLE 0.2 8B & 0.043 & 0.695 & 0.817 & 0.004 & 0.006 && 1.76 & 3.07 & 3.40 & 1.64 & 1.65 \\
OpenThaiGPT 8B & 0.041 & 0.695 & 0.815 & 0.000 & 0.000 && 1.51 & 2.55 & 2.74 & 1.43 & 1.42 \\
\bottomrule
\end{tabular}
}
\caption{best@3, the best of the three samples per item. Best per column in bold, second best underlined. Column abbreviations as in Table~\ref{tab:main-results}.}
\label{tab:best3}
\end{table}

\FloatBarrier

\subsection{Variation across samples}
\label{app:sample-sd}
Table~\ref{tab:sample-sd} scores each of the three samples alone and reports the standard deviation of the setting-level score.

\begin{table}[t]
\centering
\small
\setlength{\tabcolsep}{3.5pt}
\resizebox{\textwidth}{!}{%
\begin{tabular}{l ccccc c ccccc}
\toprule
& \multicolumn{5}{c}{\textbf{Citation F1, SD over samples}} && \multicolumn{5}{c}{\textbf{Bar-standard, SD over samples}} \\
\cmidrule(lr){2-6} \cmidrule(lr){8-12}
\textbf{Model} & \textbf{Param.} & \textbf{GC Cur.} & \textbf{GC InF.} & \textbf{RAG Cur.} & \textbf{RAG InF.} && \textbf{Param.} & \textbf{GC Cur.} & \textbf{GC InF.} & \textbf{RAG Cur.} & \textbf{RAG InF.} \\
\midrule
\multicolumn{12}{l}{\textit{General}} \\
Gemma 4 31B & 0.003 & 0.005 & 0.004 & 0.002 & 0.007 && 0.00 & 0.02 & 0.02 & 0.01 & 0.02 \\
Gemma 4 26B-A4B & 0.005 & 0.001 & 0.004 & 0.002 & 0.004 && 0.04 & 0.02 & 0.03 & 0.02 & 0.02 \\
Nemotron 3.5 Lightning 30B-A3B & 0.002 & 0.003 & 0.004 & 0.001 & 0.002 && 0.00 & 0.02 & 0.01 & 0.01 & 0.03 \\
Muse Glimmer 30B & 0.003 & 0.003 & 0.002 & 0.002 & 0.002 && 0.01 & 0.01 & 0.02 & 0.02 & 0.04 \\
\midrule
\multicolumn{12}{l}{\textit{Nemotron 3 bases and their adaptations}} \\
Nemotron 3 Nano 30B-A3B & 0.001 & 0.004 & 0.002 & 0.006 & 0.005 && 0.01 & 0.02 & 0.03 & 0.02 & 0.00 \\
\hspace{0.9em}$\hookrightarrow$\,SEA-LION v4.8 30B-A3B & 0.001 & 0.001 & 0.003 & 0.002 & 0.001 && 0.02 & 0.03 & 0.02 & 0.00 & 0.02 \\
OpenThai 2.0 Legal 30B-A3B & 0.002 & 0.005 & 0.004 & 0.001 & 0.000 && 0.01 & 0.01 & 0.02 & 0.01 & 0.01 \\
Nemotron 3 Super 120B-A12B & 0.001 & 0.003 & 0.002 & 0.001 & 0.004 && 0.02 & 0.03 & 0.03 & 0.01 & 0.03 \\
\hspace{0.9em}$\hookrightarrow$\,SEA-LION v4.8 120B-A12B & 0.003 & 0.006 & 0.002 & 0.010 & 0.008 && 0.01 & 0.03 & 0.05 & 0.07 & 0.02 \\
\midrule
\multicolumn{12}{l}{\textit{ThaiLLM 8B derivatives}} \\
Typhoon-S 8B & 0.003 & 0.001 & 0.002 & 0.001 & 0.002 && 0.00 & 0.01 & 0.02 & 0.01 & 0.09 \\
Pathumma 8B Think & 0.002 & 0.010 & 0.003 & 0.001 & 0.006 && 0.03 & 0.06 & 0.03 & 0.01 & 0.02 \\
THaLLE 0.2 8B & 0.001 & 0.004 & 0.007 & 0.000 & 0.001 && 0.01 & 0.03 & 0.03 & 0.01 & 0.01 \\
OpenThaiGPT 8B & 0.000 & 0.001 & 0.001 & 0.000 & 0.000 && 0.02 & 0.01 & 0.01 & 0.01 & 0.01 \\
\bottomrule
\end{tabular}
}
\caption{Standard deviation of the setting-level score over the three samples, each scored alone. Column abbreviations as in Table~\ref{tab:main-results}.}
\label{tab:sample-sd}
\end{table}

\FloatBarrier

\subsection{Per-model contrasts, precision and recall}
\label{app:per-model}
Table~\ref{tab:contrast-rubric-full} gives the per-model contrasts of Table~\ref{tab:contrast-rubric} with their 95\% intervals and the No-drift control of each model. Table~\ref{tab:prf} reports the section-level precision and recall behind the Citation F1 of Table~\ref{tab:main-results}.

\begin{table}[t]
\centering
\small
\setlength{\tabcolsep}{3pt}
\resizebox{\textwidth}{!}{%
\begin{tabular}{l ccc c ccc}
\toprule
& \multicolumn{3}{c}{\textbf{Gold-cited, In-force minus Current}} && \multicolumn{3}{c}{\textbf{RAG, In-force minus Current}} \\
\cmidrule(lr){2-4} \cmidrule(lr){6-8}
\textbf{Model} & \textbf{No drift} & \textbf{Amended} & \textbf{Repealed} && \textbf{No drift} & \textbf{Amended} & \textbf{Repealed} \\
\midrule
\multicolumn{8}{l}{\textit{General}} \\
Gemma 4 31B & $+0.14$ {\scriptsize[$+0.04$, $+0.24$]} & $+0.45$ {\scriptsize[$+0.35$, $+0.56$]} & $+0.98$ {\scriptsize[$+0.83$, $+1.12$]} && $-0.02$ {\scriptsize[$-0.09$, $+0.05$]} & $+0.29$ {\scriptsize[$+0.21$, $+0.38$]} & $+0.87$ {\scriptsize[$+0.74$, $+0.99$]} \\
Gemma 4 26B-A4B & $+0.08$ {\scriptsize[$+0.01$, $+0.14$]} & $+0.31$ {\scriptsize[$+0.22$, $+0.39$]} & $+0.97$ {\scriptsize[$+0.83$, $+1.11$]} && $+0.01$ {\scriptsize[$-0.07$, $+0.08$]} & $+0.03$ {\scriptsize[$-0.05$, $+0.10$]} & $+0.35$ {\scriptsize[$+0.26$, $+0.44$]} \\
Nemotron 3.5 Lightning 30B-A3B & $+0.02$ {\scriptsize[$-0.05$, $+0.09$]} & $+0.31$ {\scriptsize[$+0.23$, $+0.39$]} & $+0.86$ {\scriptsize[$+0.72$, $+1.00$]} && $-0.06$ {\scriptsize[$-0.13$, $+0.01$]} & $+0.14$ {\scriptsize[$+0.08$, $+0.20$]} & $+0.60$ {\scriptsize[$+0.50$, $+0.70$]} \\
Muse Glimmer 30B & $+0.02$ {\scriptsize[$-0.05$, $+0.09$]} & $+0.45$ {\scriptsize[$+0.35$, $+0.55$]} & $+1.35$ {\scriptsize[$+1.19$, $+1.51$]} && $-0.01$ {\scriptsize[$-0.10$, $+0.07$]} & $+0.32$ {\scriptsize[$+0.23$, $+0.40$]} & $+1.51$ {\scriptsize[$+1.35$, $+1.67$]} \\
\midrule
\multicolumn{8}{l}{\textit{Nemotron 3 bases and their adaptations}} \\
Nemotron 3 Nano 30B-A3B & $+0.08$ {\scriptsize[$+0.01$, $+0.14$]} & $+0.22$ {\scriptsize[$+0.16$, $+0.29$]} & $+0.58$ {\scriptsize[$+0.47$, $+0.68$]} && $-0.01$ {\scriptsize[$-0.06$, $+0.04$]} & $+0.10$ {\scriptsize[$+0.05$, $+0.15$]} & $+0.38$ {\scriptsize[$+0.31$, $+0.46$]} \\
\hspace{0.9em}$\hookrightarrow$\,SEA-LION v4.8 30B-A3B & $-0.08$ {\scriptsize[$-0.15$, $+0.00$]} & $+0.29$ {\scriptsize[$+0.21$, $+0.38$]} & $+0.88$ {\scriptsize[$+0.74$, $+1.01$]} && $-0.00$ {\scriptsize[$-0.07$, $+0.07$]} & $+0.18$ {\scriptsize[$+0.12$, $+0.24$]} & $+0.63$ {\scriptsize[$+0.52$, $+0.73$]} \\
OpenThai 2.0 Legal 30B-A3B & $-0.03$ {\scriptsize[$-0.12$, $+0.05$]} & $+0.12$ {\scriptsize[$+0.04$, $+0.19$]} & $+0.27$ {\scriptsize[$+0.17$, $+0.37$]} && $-0.00$ {\scriptsize[$-0.06$, $+0.05$]} & $+0.03$ {\scriptsize[$-0.02$, $+0.07$]} & $-0.00$ {\scriptsize[$-0.06$, $+0.06$]} \\
Nemotron 3 Super 120B-A12B & $+0.03$ {\scriptsize[$-0.05$, $+0.10$]} & $+0.39$ {\scriptsize[$+0.30$, $+0.49$]} & $+1.10$ {\scriptsize[$+0.95$, $+1.25$]} && $+0.02$ {\scriptsize[$-0.05$, $+0.10$]} & $+0.29$ {\scriptsize[$+0.21$, $+0.37$]} & $+0.92$ {\scriptsize[$+0.80$, $+1.05$]} \\
\hspace{0.9em}$\hookrightarrow$\,SEA-LION v4.8 120B-A12B & $-0.05$ {\scriptsize[$-0.14$, $+0.04$]} & $+0.42$ {\scriptsize[$+0.32$, $+0.52$]} & $+1.13$ {\scriptsize[$+0.98$, $+1.29$]} && $-0.12$ {\scriptsize[$-0.22$, $-0.01$]} & $+0.14$ {\scriptsize[$+0.05$, $+0.23$]} & $+0.70$ {\scriptsize[$+0.57$, $+0.83$]} \\
\midrule
\multicolumn{8}{l}{\textit{ThaiLLM 8B derivatives}} \\
Typhoon-S 8B & $+0.02$ {\scriptsize[$-0.06$, $+0.09$]} & $+0.09$ {\scriptsize[$+0.02$, $+0.16$]} & $+0.48$ {\scriptsize[$+0.37$, $+0.59$]} && $-0.05$ {\scriptsize[$-0.11$, $+0.01$]} & $-0.05$ {\scriptsize[$-0.10$, $+0.00$]} & $-0.05$ {\scriptsize[$-0.11$, $+0.01$]} \\
Pathumma 8B Think & $-0.11$ {\scriptsize[$-0.19$, $-0.03$]} & $-0.01$ {\scriptsize[$-0.09$, $+0.07$]} & $+0.45$ {\scriptsize[$+0.34$, $+0.56$]} && $+0.04$ {\scriptsize[$-0.04$, $+0.11$]} & $+0.01$ {\scriptsize[$-0.05$, $+0.07$]} & $+0.22$ {\scriptsize[$+0.15$, $+0.30$]} \\
THaLLE 0.2 8B & $+0.03$ {\scriptsize[$-0.04$, $+0.09$]} & $+0.19$ {\scriptsize[$+0.12$, $+0.26$]} & $+0.70$ {\scriptsize[$+0.59$, $+0.81$]} && $-0.00$ {\scriptsize[$-0.05$, $+0.04$]} & $+0.01$ {\scriptsize[$-0.03$, $+0.05$]} & $-0.00$ {\scriptsize[$-0.05$, $+0.05$]} \\
OpenThaiGPT 8B & $-0.06$ {\scriptsize[$-0.14$, $+0.01$]} & $+0.09$ {\scriptsize[$+0.03$, $+0.15$]} & $+0.48$ {\scriptsize[$+0.38$, $+0.58$]} && $-0.00$ {\scriptsize[$-0.05$, $+0.05$]} & $+0.00$ {\scriptsize[$-0.04$, $+0.04$]} & $-0.01$ {\scriptsize[$-0.06$, $+0.05$]} \\
\midrule
All models & $+0.01$ {\scriptsize[$-0.02$, $+0.03$]} & $+0.25$ {\scriptsize[$+0.21$, $+0.30$]} & $+0.79$ {\scriptsize[$+0.70$, $+0.88$]} && $-0.02$ {\scriptsize[$-0.04$, $+0.00$]} & $+0.11$ {\scriptsize[$+0.09$, $+0.14$]} & $+0.47$ {\scriptsize[$+0.42$, $+0.52$]} \\
\quad minus No drift & -- & $+0.25$ {\scriptsize[$+0.20$, $+0.30$]} & $+0.78$ {\scriptsize[$+0.69$, $+0.87$]} && -- & $+0.13$ {\scriptsize[$+0.10$, $+0.16$]} & $+0.49$ {\scriptsize[$+0.43$, $+0.54$]} \\
\bottomrule
\end{tabular}
}
\caption{Change in Bar-standard score from the text in force over the current text, by drift band, avg@3, with 95\% intervals from a bootstrap over item clusters. All models pools the items. Minus No drift subtracts the No-drift change.}
\label{tab:contrast-rubric-full}
\end{table}

\begin{table}[t]
\centering
\small
\setlength{\tabcolsep}{3.5pt}
\resizebox{\textwidth}{!}{%
\begin{tabular}{l ccccc}
\toprule
\textbf{Model} & \textbf{Param.} & \textbf{GC Cur.} & \textbf{GC InF.} & \textbf{RAG Cur.} & \textbf{RAG InF.} \\
\midrule
\multicolumn{6}{l}{\textit{Citation precision}} \\
\multicolumn{6}{l}{\textit{General}} \\
Gemma 4 31B & 0.411 & 0.761 & 0.881 & 0.551 & 0.647 \\
Gemma 4 26B-A4B & 0.375 & 0.791 & 0.904 & 0.319 & 0.363 \\
Nemotron 3.5 Lightning 30B-A3B & 0.035 & 0.763 & 0.871 & 0.351 & 0.430 \\
Muse Glimmer 30B & 0.206 & 0.791 & 0.898 & 0.458 & 0.548 \\
\multicolumn{6}{l}{\textit{Nemotron 3 bases and their adaptations}} \\
Nemotron 3 Nano 30B-A3B & 0.023 & 0.746 & 0.851 & 0.232 & 0.302 \\
\hspace{0.9em}$\hookrightarrow$\,SEA-LION v4.8 30B-A3B & 0.046 & 0.764 & 0.869 & 0.338 & 0.419 \\
OpenThai 2.0 Legal 30B-A3B & 0.090 & 0.510 & 0.546 & 0.005 & 0.006 \\
Nemotron 3 Super 120B-A12B & 0.108 & 0.786 & 0.895 & 0.421 & 0.522 \\
\hspace{0.9em}$\hookrightarrow$\,SEA-LION v4.8 120B-A12B & 0.099 & 0.775 & 0.886 & 0.339 & 0.403 \\
\multicolumn{6}{l}{\textit{ThaiLLM 8B derivatives}} \\
Typhoon-S 8B & 0.132 & 0.758 & 0.836 & 0.017 & 0.050 \\
Pathumma 8B Think & 0.189 & 0.754 & 0.852 & 0.169 & 0.219 \\
THaLLE 0.2 8B & 0.029 & 0.755 & 0.862 & 0.002 & 0.003 \\
OpenThaiGPT 8B & 0.026 & 0.677 & 0.779 & 0.000 & 0.000 \\
\midrule
\multicolumn{6}{l}{\textit{Citation recall}} \\
\multicolumn{6}{l}{\textit{General}} \\
Gemma 4 31B & 0.268 & 0.601 & 0.739 & 0.396 & 0.470 \\
Gemma 4 26B-A4B & 0.239 & 0.621 & 0.750 & 0.196 & 0.226 \\
Nemotron 3.5 Lightning 30B-A3B & 0.027 & 0.629 & 0.751 & 0.280 & 0.345 \\
Muse Glimmer 30B & 0.219 & 0.748 & 0.902 & 0.517 & 0.630 \\
\multicolumn{6}{l}{\textit{Nemotron 3 bases and their adaptations}} \\
Nemotron 3 Nano 30B-A3B & 0.019 & 0.633 & 0.753 & 0.213 & 0.266 \\
\hspace{0.9em}$\hookrightarrow$\,SEA-LION v4.8 30B-A3B & 0.038 & 0.611 & 0.724 & 0.282 & 0.342 \\
OpenThai 2.0 Legal 30B-A3B & 0.071 & 0.408 & 0.461 & 0.004 & 0.004 \\
Nemotron 3 Super 120B-A12B & 0.108 & 0.672 & 0.804 & 0.356 & 0.441 \\
\hspace{0.9em}$\hookrightarrow$\,SEA-LION v4.8 120B-A12B & 0.100 & 0.652 & 0.785 & 0.321 & 0.379 \\
\multicolumn{6}{l}{\textit{ThaiLLM 8B derivatives}} \\
Typhoon-S 8B & 0.085 & 0.555 & 0.643 & 0.011 & 0.032 \\
Pathumma 8B Think & 0.126 & 0.536 & 0.608 & 0.110 & 0.144 \\
THaLLE 0.2 8B & 0.020 & 0.565 & 0.664 & 0.001 & 0.002 \\
OpenThaiGPT 8B & 0.019 & 0.545 & 0.647 & 0.000 & 0.000 \\
\bottomrule
\end{tabular}
}
\caption{Citation precision, recall, and section-only F1 by model and setting, avg@3. Column abbreviations as in Table~\ref{tab:main-results}.}
\label{tab:prf}
\end{table}

\FloatBarrier

\paragraph{Rule change labels.} For each changed gold section of an Amended item, we manually annotated whether the amendment changed the rule the question turns on or only the wording. Table~\ref{tab:rule-change} in \S\ref{subsec:analysis} gives the loss by group.

\fcend
\end{document}